%% file: main.tex
\documentclass[10pt,twocolumn,letterpaper]{article}

\usepackage[pagenumbers]{cvpr} 

\input{preamble}

\definecolor{cvprblue}{rgb}{0.21,0.49,0.74}
\usepackage[pagebackref,breaklinks,colorlinks,allcolors=cvprblue]{hyperref}

\def\confName{CVPR}
\def\confYear{2027}
\title{ArtiFix: Localizing and Repairing Artifacts in AI-Generated Videos}  

\author{Henry Kam\textsuperscript{1}\quad
Jenna Kang\textsuperscript{1}\quad
Niall L. Williams\textsuperscript{2}\quad
Sai Qian Zhang\textsuperscript{1}\quad
Qi Sun\textsuperscript{1}\\
\textsuperscript{1}New York University\quad
\textsuperscript{2}Barnard College\\
{\tt\small \{henrykam, jennakang, sai.zhang, qisun\}@nyu.edu\quad nlw@barnard.edu}
}
\begin{document}
\input{figures/fig_teaser}
\input{sec/0_abstract}
\input{sec/1_intro}

\input{sec/2_related}
\input{sec/3_method}
\input{sec/4_experiments}
\input{sec/4b_study}

\input{sec/4c_results}

\input{sec/5_discussion}

\input{sec/5_conclusion}

\input{sec/llm}
{
    \small
    \bibliographystyle{ieeenat_fullname}
    \bibliography{main}
}

\input{sec/X_suppl}

\end{document}

%% file: preamble.tex
\makeatletter
\renewenvironment{abstract}{%
  \centerline{\large\bf Abstract}\par
  \vspace{0.5em}%
  \itshape
}{\par}
\makeatother

\usepackage{xspace}
\newcommand{\method}{ArtiFix\xspace}

\newif\ifdraft
\drafttrue

\usepackage{titletoc}

%% file: figures/fig_teaser.tex
\twocolumn[{%
\renewcommand\twocolumn[1][]{#1}%
\maketitle
\begin{center}
\captionsetup{type=figure}
\includegraphics[width=0.97\textwidth]{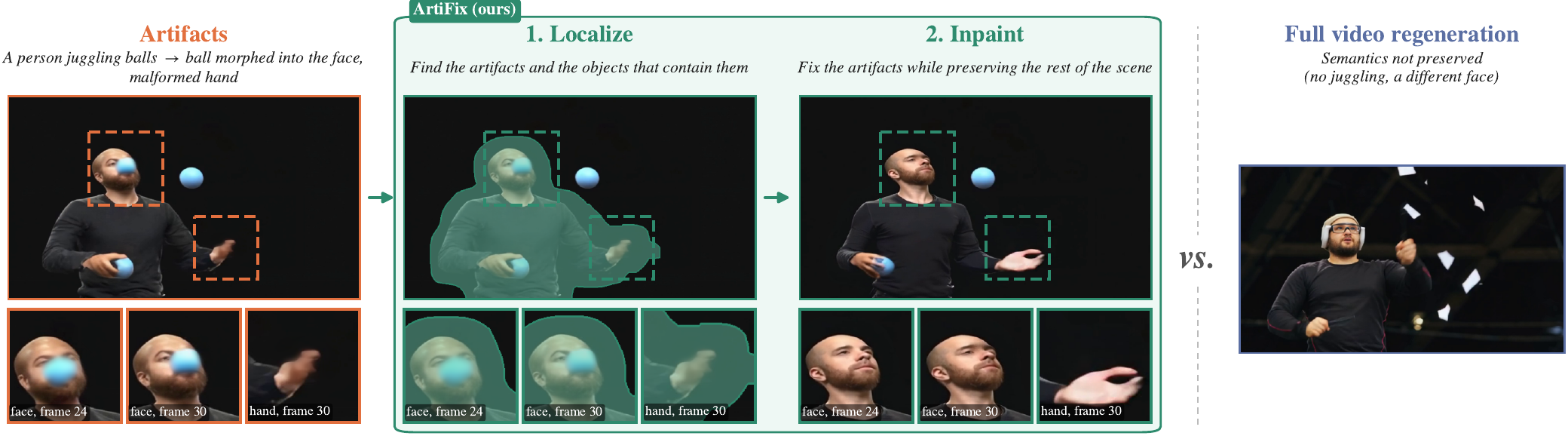}
\captionof{figure}{
We present \emph{ArtiFix},  an automatic pipeline that repairs local artifacts in AI-generated videos while keeping the rest of the video intact. In this generated clip of a person juggling, a ball morphs into the face while the hand is malformed (left). Our localizer finds the artifacts together with the objects that contain them (middle left), and our inpainter redraws only that region while every other pixel is copied directly from the input (middle right, insets below). Regenerating the whole video also removes the artifacts, but it changes the person and no longer shows them juggling (right). ArtiFix therefore turns artifact repair into a local edit that needs neither the original prompt nor the generator. Video generated by CogVideoX-5B.
}
\label{fig:teaser}
\end{center}%
}]

%% file: sec/0_abstract.tex
\begin{abstract}
AI-generated videos may appear realistic, but an occasional local artifact, such as an extra finger or two, can ruin the entire result.
Iteratively regenerating the entire video incurs the full generation cost and risks changing content that was already correct.
We introduce \method, an automatic pipeline that repairs such artifacts from the video alone and leaves every pixel outside the repair region unchanged.
A lightweight video localizer identifies the region containing the artifact, and a video inpainting model reconstructs the masked region, conditioned on the surrounding video and a reference cutout of the original object to preserve its identity.
Additionally, the computational cost of artifact inpainting scales with the size of the repair region rather than the full video.
We further refine this inpainter through preference tuning on its own samples, ranked by a Video Jev video realism model, improving the observer-rated repair quality without a detected loss in preservation.
Because current automated metrics do not reliably capture artifact removal, we evaluate repairs through a perceptual study.
Observers rank \method highest in both artifact removal and original video preservation.
\end{abstract}

%% file: sec/1_intro.tex
\section{Introduction}
\label{sec:intro}
AI-generated videos can appear realistic yet contain brief, localized artifacts.
For example, a deck of cards may momentarily collapse into a distorted sheet during shuffling. Rather than regenerate the video and risk altering its composition or appearance, creators need to repair the defect while preserving the correct content. 
Such artifacts are often spatially localized, as the median BrokenVideos~\cite{lin2025brokenvideos} artifact mask covers roughly 4.4\% of the frame and 99\%  of masks cover less than half of the frame.

Existing approaches do not directly address this task. Regeneration may correct artifacts but alter otherwise valid content, while generation-time correction methods require access to the original generation process~\cite{jang2026self,polowczyk2026diamond}. 
Video inpainting better preserves surrounding pixels but requires a repair mask and does not automatically determine what to redraw~\cite{zhou2023propainter,litman2026editctrl}. We introduce \method, an automatic pipeline that takes only the input video, identifies the region needing repair, and reconstructs it while leaving every pixel outside the region unchanged.

A key challenge is that the region \emph{needed for repair} may extend beyond the \emph{visibly defective pixels}. For example, a displaced object may require repairing both its incorrect position and its original location (see the supplementary material). We construct object-level repair labels using a 72B vision-language model and SAM3~\cite{carion2026sam3}, then distill them into a 45M-parameter video localizer. This model predicts complete repair regions over 100 times faster than the teacher pipeline.

Another challenge is preserving the affected object's appearance without reproducing its defect. Conditioning on corrupted pixels can propagate the artifact, while masking them removes useful appearance cues. We address this by combining masked video conditioning with a single-frame object reference in VACE-14B~\cite{jiang2025vace}. We further reduce computation by restricting denoising to the repair region, a surrounding context band, and sparse global tokens, rather than processing the entire video densely.

Finally, improving repair quality requires distinguishing artifact removal from unwanted content changes.
Our evaluation shows that whole-video scores are unreliable judges of repair quality across methods with varying preservation behavior. However, candidates for the same local repair also share the same unedited surroundings.
In this restricted setting, we find that Video Jev~\cite{retina_videojev}, a video realism model, provides a useful signal for ranking repair candidates.
We use these rankings to construct preference pairs and fine-tune the inpainter with Diffusion-DPO~\cite{wallace2024diffusion}, improving repair quality without additional inference-time sampling.

\input{figures/fig_pipeline}

We evaluate on 103 held-out BrokenVideos clips and a separate 580-clip example pool. In a perceptual study with eleven observers and 1,100 trials, \method significantly outperforms evaluated open inpainting and regeneration methods in both artifact removal and content preservation. 
Compared with Gemini Omni~\cite{googledeepmind_geminiomni}, it achieves higher average artifact-removal ratings and significantly better preservation. Our localizer covers 97\% of annotated artifact pixels and correctly identifies 88\% of clean real videos as requiring no repair. Regional denoising further reduces runtime from 125\,s to 77\,s.
In summary, we make the following main contributions.
\begin{itemize}
    \item \emph{Automatic repair localization:}
    A compact video localizer that identifies complete repair regions beyond visibly defective pixels.
    \item \emph{Appearance-preserving regional repair:} Combining masked video conditioning with object appearance references and efficient regional denoising.
    \item \emph{Same-context preference tuning:}
    A video realism model ranks candidates of the same local
    repair and constructs preference pairs for tuning.
    Our perceptual study, with 1,100 rated trials from eleven observers, assesses both artifact removal
    and preservation against inpainting and regeneration baselines.
\end{itemize}

%% file: figures/fig_pipeline.tex
\begin{figure*}[t]
\centering
\includegraphics[width=0.92\textwidth]{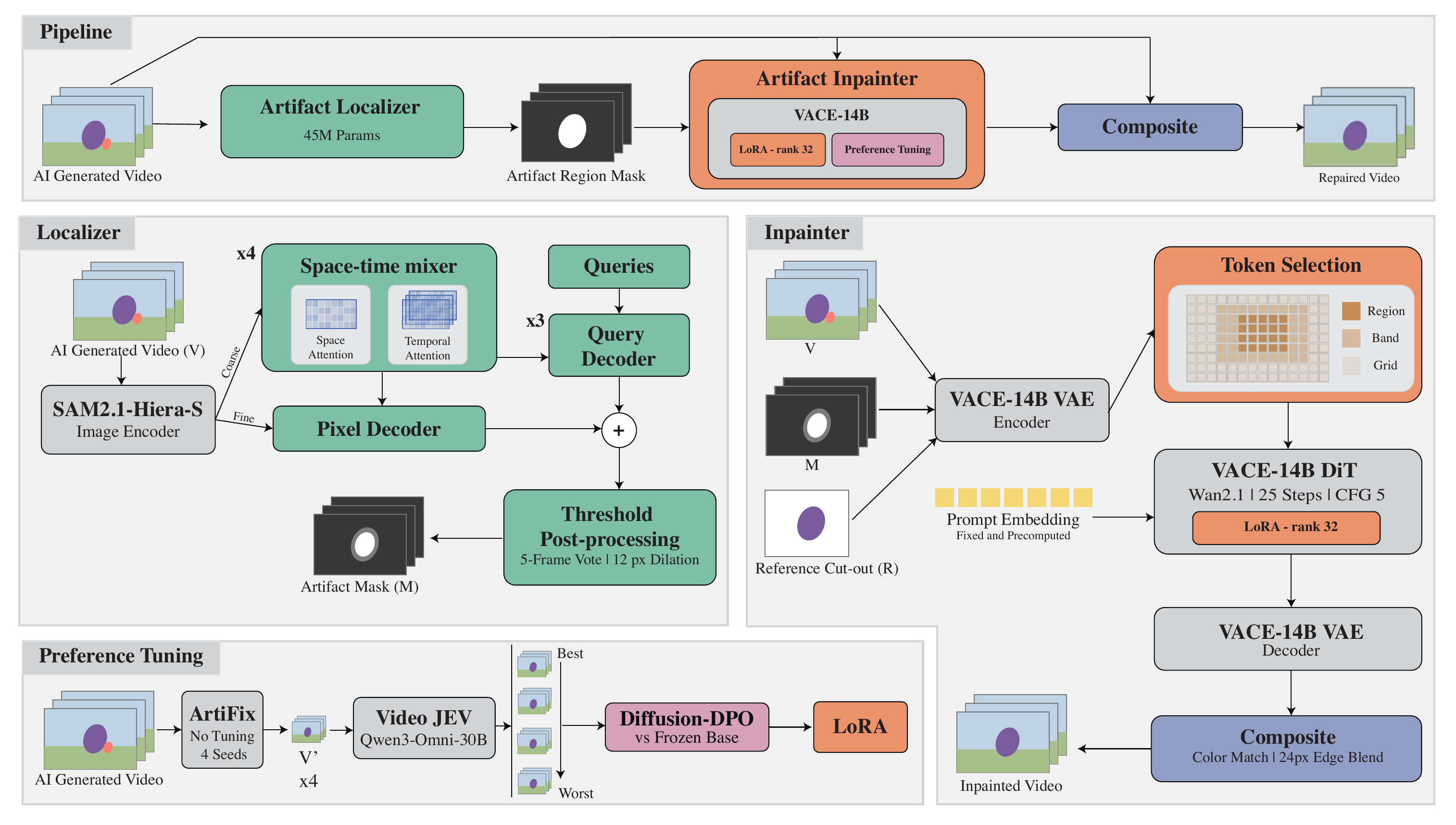}
\caption{\textbf{Overview of \method.} The localizer predicts the region $M$ to redraw, the inpainter redraws it from the video with the region hidden and a reference cut-out, and the composite keeps every pixel outside $M$. Preference tuning (training only) trains the inpainter's LoRA on pairs of its own samples ranked by Video Jev.}
\label{fig:pipeline}
\end{figure*}

%% file: sec/2_related.tex
\section{Related Work}
\label{sec:related}

\paragraph{Generative video artifacts.}
Artifacts that violate anatomy, physics or object permanence remain common in strong video generators~\cite{bansal2025videophy,kang2024far}.
They break the perceptual straightness of natural videos~\cite{henaff2019perceptual,interno2026ai}, and motion artifacts draw more gaze than texture artifacts~\cite{klein2025perceptual}.
Datasets mark artifacts with pixel masks~\cite{lin2025brokenvideos}, per-frame boxes~\cite{kang2026geneva} or timed traces~\cite{fu2025learning}, but multimodal models struggle with small or temporal artifacts~\cite{tang2026artifact}.

\paragraph{Artifact localization.}
Promptable segmentation~\cite{ravi2025sam2,carion2026sam3} and reasoning segmentation~\cite{lai2024lisa,bai2024one,yan2024visa,lin2025glus,cuttano2025samwise} segment a specified target but do not decide which region a repair needs.
Image artifact detectors~\cite{zhang2023perceptual,wang2024detecting,kang2025legion} miss temporal artifacts, and video detectors classify clips~\cite{chen2026demamba} or output boxes~\cite{li2026skyra} rather than masks.
We train a video localizer that identifies the regions of the videos to redraw.

\paragraph{Video inpainting and editing.}
Video inpainting propagates content from neighboring frames~\cite{zhou2023propainter}, which can copy a corrupted source, or generates it from context and text~\cite{bian2025videopainter,zhang2024avid,jiang2025vace,litman2026editctrl}.
Language-driven inpainting~\cite{wu2024towards} and reference-guided insertion~\cite{tu2025videoanydoor} still need the user to say where to edit or what to add.
We build on VACE~\cite{jiang2025vace}, which combines masked inpainting with reference images.

\paragraph{Refining generated content.}
Image refiners detect and inpaint hands, faces and general artifacts~\cite{lu2024handrefiner,wang2025realishuman,xu2026genshield}, but their per-frame application is not temporally coherent.
Video refiners act inside the sampling of one generator~\cite{jang2026self,polowczyk2026diamond}, search over samples~\cite{liu2025video,ma2025inference,xue2025phyt2v} or regenerate regions with the original prompt and seed~\cite{lee2026self}, so they need access to the generator.
Per-frame image editing~\cite{mao2026lynnreal} and regeneration~\cite{meng2021sdedit} risk changing pixel content that was already correct.

\paragraph{Efficient localized inference.}
Sparse inference restricts image edits to the edited region~\cite{li2022efficient,nitzan2024lazy}, and video editors cache tokens away from the edit~\cite{kahatapitiya2024object} or denoise masked tokens through trained adapters~\cite{litman2026editctrl}.
General accelerators~\cite{bolya2023token,xi2025sparse,liu2025timestep,liu2026region} are complementary.
We skip untouched tokens exactly through their known noisy states in flow matching, without training.

\paragraph{Preference optimization.}
Diffusion-DPO~\cite{wallace2024diffusion}, flow-matching objectives for video~\cite{liu2026improving,liu2026flow}, dense and regional feedback~\cite{wu2026densedpo,huang2026mind} and reward-ranked samples~\cite{dong2023raft} improve generation, and preference tuning of inpainters has focused on images~\cite{liu2024prefpaint,yuan2026follow}.
Learned video evaluators can rank candidates~\cite{zhang2026vq,liu2026improving}.
We use Video Jev~\cite{retina_videojev} to rank repairs of the same region, whose shared surroundings make the ranking more informative about the edit.

%% file: sec/3_method.tex
\section{Method}
\label{sec:method}

Given a generated video $V \in \mathbb{R}^{T\times H\times W\times 3}$, a localizer predicts a repair mask $M \in \{0,1\}^{T\times H\times W}$ (\cref{sec:loc}), an inpainter redraws the content inside $M$ (\cref{sec:inpaint}), and the output copies every pixel outside $M$ from $V$.
Preference tuning improves the inpainter (\cref{sec:pref}), and region-sparse denoising reduces its cost (\cref{sec:sparse}).

\subsection{Repair scope and design requirements}
\label{sec:artifacts}
We target localized failures for which most of the video should stay unchanged, and leave global failures out of scope.
The repair mask may need to include the affected object and the locations of its displacement rather than only the visibly defective pixels (see the supplementary material).
Undercoverage leaves part of the defect unfixed, whereas overcoverage causes correct content to be re-generated.
We therefore design our pipeline to predict the region needed for a consistent repair, retain appearance information without conditioning on all defective frames, reduce denoising cost for smaller regions, and operate without the original generation inputs or human annotations.

\subsection{Localizer}
\label{sec:loc}
A vision-language model (VLM) combined with a segmentation model can outline the broken object, but it takes about 140\,s per 5-second clip on two NVIDIA H200 GPUs (see the supplementary material), mostly to generate text for a question that is the same for every clip.
We therefore train a small video segmentation model to answer this question directly, and we use the VLM pipeline only to create its training labels.

The localizer predicts two maps from shared features (see the supplementary material).
The \emph{region map} marks the pixels that the inpainter has to redraw for a consistent repair, including the whole affected object and the places it moves through (requirement 1).
It defines the repair mask $M$ and is the only map used at test time.
The \emph{artifact map} marks only the pixels that visibly look broken, as annotated in BrokenVideos.
The region label includes these annotated pixels, so the predicted region also covers the broken pixels, and the artifact map serves only as auxiliary supervision for the shared layers.

\paragraph{Training targets.}
The artifact map is trained on the BrokenVideos pixel masks~\cite{lin2025brokenvideos}.
No dataset marks the region to redraw, so a teacher similar to reasoning segmentation~\cite{lai2024lisa,bai2024one} creates these labels.
Qwen2.5-VL-72B~\cite{bai2025qwen25vl} views the clip with the annotation outlined and lists the objects to redraw, and SAM3~\cite{carion2026sam3} tracks them.
The ground-truth label is the union of these tracks, the annotation and the artifact map of an earlier localizer, under a soft budget of 15\% of the pixels (see the supplementary material).
The annotation is used only to build these labels, and neither our localizer nor the annotation-free VLM baseline receives it at test time.

\paragraph{Architecture.}
The localizer has 45M parameters: a frozen 34M SAM2.1-Hiera-S image encoder~\cite{ravi2025sam2} and 11M trainable parameters.
Some artifacts, such as morphing or flicker, are only visible in motion, so the coarsest encoder features pass through four space-time attention layers~\cite{bertasius2021space}.

A pixel decoder then adds back the finer features and gives an embedding $e(i) \in \mathbb{R}^{C}$, $C=256$, for every pixel $i$ at stride 4.
Reasoning segmentation models~\cite{lai2024lisa,bai2024one} use a language model to turn a text question into a query vector, from which a mask decoder predicts the mask.
Our question is the same for every clip, so we replace the language model with $K=4$ query vectors per map, which are learned jointly with the localizer as in Mask2Former~\cite{cheng2022masked}.
A three-layer transformer decoder adapts these queries to the clip, since in each layer they attend to each other and to the space-time features of all frames.
Each map $m \in \{\mathrm{art}, \mathrm{reg}\}$ then compares its updated queries $q_{m,k}$ with every pixel embedding,

\begin{equation}
s_m(i) = \log \frac{1}{K} \sum_{k=1}^{K} \exp\!\Big(\frac{q_{m,k}^{\top} e(i)}{\sqrt{C}}\Big),
\label{eq:locprob}
\end{equation}
and the probability is $p_m(i) = \sigma\big(s_m(i)\big)$, where $\sigma$ is the logistic function.
The log-sum-exp acts as a soft maximum over the queries, so a pixel receives a high probability when any single query matches it strongly.
Different queries can thus cover different parts of the region, and the map approximates their union.

At test time, the localizer processes a whole clip in one pass in 1.2\,s on one NVIDIA H200 GPU.

\paragraph{Loss.}
Each map is trained with a pixel-weighted binary cross entropy (BCE) and a Dice loss~\cite{milletari2016v}, with a reduced weight on unlabeled pixels because the annotators mark only the most visibly broken ones.
To keep the localizer from marking a region on every video, a quarter of the training samples are real videos without artifacts from DeeptraceReward~\cite{fu2025learning}, on which both maps are trained to be empty (see the supplementary material).

\paragraph{Inference.}
At test time, the region map is thresholded, smoothed over time, cleaned of small components and enlarged by 12 pixels, because a region that is too small does more harm than one that is too large (see the supplementary material).

\subsection{Region inpainting}
\label{sec:inpaint}
If the inpainter sees the broken pixels, it tends to draw the same defect again.
Removing the region can also remove appearance cues needed to preserve the object (\cref{tab:ladder}).
We therefore hide the region from the model and show it the object in a separate reference image (requirement 2).

We use VACE-14B~\cite{jiang2025vace}, a video diffusion transformer based on Wan2.1~\cite{wan2025wan} that takes a conditioning video, a mask, and reference images (\cref{fig:pipeline}), on the 49-frame window at $832\times480$.
The region is set to zero in the conditioning video, and a reference cut-out of the region on a white background supplies the appearance of the object (requirement 2).
All clips use the generic prompt ``the video with all artifacts corrected, physically plausible'', since clip-specific descriptions from the 72B model do not improve the repairs (see the supplementary material).
Inside $M$, the output is blended toward the original over 24 pixels from the border and color-matched to the original region, which gives $\tilde V$, and the repair is
\begin{equation}
\hat V = (1 - M) \odot V + M \odot \tilde V,
\label{eq:composite}
\end{equation}
so every pixel outside $M$ equals the input.

\subsection{Preference tuning}
\label{sec:pref}
The inpainter starts from random noise, so the quality of a repair depends on the seed.
One sample may remove the artifact, while another draws the same defect in a new form.
However, picking the best of several samples at test time would multiply the cost.
Instead, we tune the inpainter with Diffusion-DPO~\cite{wallace2024diffusion}, a version of direct preference optimization (DPO) for diffusion models, which needs only a ranking of the samples (\cref{fig:pipeline}).

\paragraph{Ranking samples.}
Similarity to the original, such as CLIP similarity, cannot tell whether the artifact is gone, and adding the localizer's estimate of the remaining artifact makes the tuned repairs worse (see the supplementary material).
We therefore rank the samples with Video Jev~\cite{retina_videojev}, a video model that answers yes-or-no questions about a video with probabilities.
Video Jev is asked whether the video is AI-generated and whether it contains visual artifacts, and the ranking score of a sample $x$ is
\begin{equation}
s(x) = p_{\mathrm{real}}(x)\, p_{\mathrm{no\text{-}artifact}}(x).
\end{equation}
Video Jev rates how real the whole video looks and is therefore not reliable for comparing methods, since a method that regenerates the whole video can look real without repairing the original (\cref{sec:measuring}).
The candidates of one clip, however, are repairs of the same region that copy the same pixels outside it.
They differ only inside the region, so a difference in their Video Jev scores mainly reflects how well the region was redrawn.

\paragraph{Preference pairs.}
The untuned pipeline repairs 400 BrokenVideos training clips with up to four seeds each, on the regions of an earlier version of our localizer (see the supplementary material).
Each pair joins the sample with the highest Jev score $s$, which is preferred, and the sample with the lowest.
We drop near ties, with a gap in $s$ below $0.05$, since they carry little preference signal.
Jev rates realism rather than preservation, so a preferred sample could look more real only because it rewrote more of the region.
We therefore also drop pairs whose preferred sample is less similar to the original than the rejected one, by more than $0.02$ in CLIP similarity~\cite{radford2021learning} averaged over frames, so that the tuning cannot learn to rewrite the region.

\paragraph{Training.}
The tuned model $v_\theta$ adds a rank-32 low-rank adapter (LoRA)~\cite{hu2021lora} with 184M parameters to the untuned inpainter $v_{\mathrm{ref}}$.
With the denoising error restricted to the region $M_z$ on the latent grid, with $C=16$ channels,
\begin{equation}
e(v, x) = \frac{1}{C\,|M_z|} \sum_{i \in M_z} \big\| (\epsilon - z_0^x)_i - v(z_t^x, t)_i \big\|^2 ,
\end{equation}
the Diffusion-DPO loss for a preferred sample $w$ and a rejected sample $l$ is
\begin{equation}
\mathcal{L}_{\mathrm{DPO}} = -\log \sigma\big(-\beta\,[\delta_w - \delta_l]\big),
\end{equation}
where $\delta_x = e(v_\theta, x) - e(v_{\mathrm{ref}}, x)$, so the loss favors a larger improvement over $v_{\mathrm{ref}}$ on the preferred sample than on the rejected one.

\subsection{Restricting compute to the region}
\label{sec:sparse}
A dense denoising pass costs the same whether the region covers 5\% or 60\% of the frame, although every pixel outside $M$ is copied from the input (\cref{eq:composite}).
In the flow-matching formulation of Wan2.1, the latent at noise level $\sigma_t$ is $z_t = (1-\sigma_t)\,z_0 + \sigma_t\,\epsilon$, where $z_0$ is the latent of the original video and $\epsilon$ the initial noise, and the model predicts the velocity $\epsilon - z_0$.
We mark every latent token that contains a pixel of $M$, together with the ring of tokens around it, one token in each spatial direction, so that the border of the region is redrawn with it instead of meeting frozen tokens.
These tokens are \emph{active}.
Every token that is not active receives the exact velocity $\epsilon - z_0$, which keeps it on the noised original without a model call (requirement 3).
The transformer runs on the active and reference tokens and on a sparse context of nearby and grid tokens (\cref{fig:pipeline}), and dense inference is used when more than 85\% of tokens are kept (see the supplementary material).
Because attention sees only these tokens, the result differs from a dense pass, and we measure its effect on quality (\cref{tab:ladder}).

%% file: sec/4_experiments.tex
\section{Experiments}
\label{sec:exp}

\subsection{Setup}

\paragraph{Data.}
The test set has 103 BrokenVideos~\cite{lin2025brokenvideos} test clips from 12 generators, excluded from training.
All methods process the same 49-frame window, centered between the first and the last annotated artifact frame.
The annotations select the window but are not provided to our localizer, which processes the full clip before its prediction is restricted to the window.
Besides the test set, we evaluate on an example pool of 580 clips that played no part in training or tuning.
It contains 520 clips generated by Wan2.2-A14B, Wan2.1-1.3B~\cite{wan2025wan}, CogVideoX-5B and CogVideoX-2B~\cite{yang2025cogvideox} from 130 artifact-prone prompts, and 60 BrokenVideos clips outside the test set.
The generated clips carry no annotations, so the pool serves for automatic measures at scale and as a source of examples, while the 103 test clips remain the main benchmark.
Repair regions in the pool are predicted without annotations, and the localizer proposes a region for 529 of the 580 clips and returns the remaining 51 unchanged.

\paragraph{Baselines.}
We evaluate each localizer in two ways.
We first compare its masks with the dataset annotation, and then we repair every clip from its masks with the same inpainter, our pipeline without preference tuning, so that differences between the repairs come only from the masks.
Localizers are the dataset annotation and the teacher pipeline with Qwen2.5-VL-72B~\cite{bai2025qwen25vl} and SAM3 run without the annotation, and the supplementary material compares further localizers.
Repair methods receive our region.
Region inpainting baselines are ProPainter~\cite{zhou2023propainter}, per-frame SDXL inpainting~\cite{podell2024sdxl} and EditCtrl~\cite{litman2026editctrl} on VACE-14B, none of which uses a reference image.
Regeneration baselines redraw the whole window with VACE-1.3B and VACE-14B~\cite{jiang2025vace} and with Wan2.2-A14B at strength $0.85$, all with our prompt.
Gemini Omni 1.1 Flash~\cite{googledeepmind_geminiomni} is a closed commercial reference that re-renders the clip from a repair instruction.
Its content filters refused some clips, mostly with photorealistic children, so it covers 63 test clips and 117 pool clips, on which its automatic scores are computed.

\paragraph{Measures.}
Preservation against the original window uses LPIPS~\cite{zhang2018unreasonable}, FID~\cite{heusel2017gans}, and the similarity of CLIP~\cite{radford2021learning} and DINOv2~\cite{oquab2023dinov2} embeddings.
Qwen2.5-VL-72B rates the artifact severity $s$ from 0 to 9, and we report $\Delta = s(\hat V) - s(V)$, where negative values mean lower rated severity, not necessarily a successful repair.
Open models run on one NVIDIA H200 GPU with a single fixed seed.

\subsection{Measuring artifact repair}
\label{sec:measuring}

\input{tables/t0_measures}

Successful repair requires both artifact removal and preservation of the video.
Neither alone is sufficient, since returning the input unchanged maximizes preservation by construction while removing no artifact, and a different clean video does not preserve the scene.
Preservation metrics compare with an original that contains the artifact, so they cannot show removal (\cref{tab:measures}).
Whole-video scores have the opposite limitation, as the VLM judge rates regeneration as cleaner ($\Delta=-0.72$ for VACE-14B and $-0.87$ for Wan2.2) despite large changes (DINOv2 $0.71$ and $0.69$), and Video Jev~\cite{retina_videojev} also favors regeneration.
These scores may reflect cleaner-looking outputs, but they do not establish a repair of the original.
The judge rates our repaired video slightly worse in quality than the input on average ($\Delta=+0.13$, cleaner on 49 of 103 clips) and tracks observer ratings of removal only weakly ($\rho=0.43$, \cref{sec:auto_vs_obs}), consistent with multimodal models missing small and temporal artifacts~\cite{tang2026artifact}.
Observer ratings are therefore the primary evidence for removal (\cref{sec:study}).

%% file: tables/t0_measures.tex
\begin{table}[t]
\centering\footnotesize
\setlength{\tabcolsep}{2.2pt}
\sbox0{\begin{tabular}{lcccccc}
\toprule
 & \multicolumn{4}{c}{Automatic} & \multicolumn{2}{c}{Observers} \\
\cmidrule(lr){2-5}\cmidrule(lr){6-7}
 & LPIPS & DINOv2 & Judge & Jev & Fixed & Pres. \\
 & $\downarrow$ & $\uparrow$ & $\Delta\downarrow$ & $\uparrow$ & $\uparrow$ & $\uparrow$ \\
\midrule
Input unchanged & 0 & 1 & 0 & 0.21 & -- & -- \\
\method & \textbf{0.119} & \textbf{0.906} & +0.13 & 0.22 & \textbf{3.63} & \textbf{4.36} \\
\quad w/o tuning & 0.122 & \textbf{0.906} & +0.30 & 0.20 & 3.32 & 4.24 \\
EditCtrl & 0.210 & 0.741 & +1.93 & 0.11 & 1.36 & 1.71 \\
Omni 1.1 Flash & 0.158 & \textbf{0.906} & -0.59 & 0.33 & 2.95 & 3.40 \\
VACE-14B regen. & 0.371 & 0.711 & -0.72 & \textbf{0.36} & 2.06 & 1.52 \\
Wan2.2 regen. & 0.391 & 0.694 & \textbf{-0.87} & 0.32 & 2.37 & 1.48 \\
\bottomrule
\end{tabular}}
\ifdim\wd0>\linewidth\resizebox{\linewidth}{!}{\usebox0}\else\usebox0\fi
\caption{What the measures see on the 103 test clips. Returning the input scores perfectly on preservation while fixing nothing, and the VLM judge and Video Jev rate regeneration as the cleanest, whereas observers (SUREAL MOS, ten study clips) rate it lowest.}
\label{tab:measures}
\end{table}

%% file: sec/4b_study.tex
\subsection{Perceptual experiment}
\label{sec:study}

We ran a within-subjects experiment on generated videos with artifacts.
For each video, observers rated how well a method localizes the artifact, how well it removes it, and how well it preserves the rest of the video.
Every observer rated every combination of clip and condition, so differences between conditions are measured within observers and within clips.
We test the following hypotheses.
\begin{description}
\item[H1] Local repair with \method preserves the original video better than regenerating it, and removes the artifact at least as well as the strongest regenerator.
\item[H2] With the same inpainter repairing every region, our localizer, which needs no annotation, marks the artifact as well as the human dataset annotation and better than the annotation-free 72B pipeline it learns from.
\item[H3] Preference tuning (\cref{sec:pref}) improves the repairs without costing preservation.
\end{description}

\paragraph{Stimuli.}
We selected 10 clips at random from the videos that have a Gemini Omni 1.1 Flash repair, so that the strongest commercial baseline appears on every clip.
We did not inspect the clips before drawing them.
Seven of them are BrokenVideos test clips with dataset annotations, and the other three are generated clips from the example pool, which have no annotation.
Each trial shows the original on the left, the edited region as a red overlay on the original in the center, and the result on the right.
Observers were never told which localization or inpainting method they saw.
The localization overlay shows a whole-frame region for the regenerators. 

\paragraph{Conditions.}
Each clip appears under ten conditions, seven that compare repair methods and three that compare localizers, which gives 100 trials per observer.
Of the seven repair methods, four edit only the region of our localizer, namely \method, the same pipeline without preference tuning, the same pipeline without preference tuning and without the reference cut-out, and EditCtrl~\cite{litman2026editctrl}.
The other three regenerate the whole video, namely Gemini Omni 1.1 Flash~\cite{googledeepmind_geminiomni}, Wan2.2-A14B and VACE-14B~\cite{jiang2025vace}.
The three localization conditions repair the regions from the dataset annotation, the Qwen2.5-VL-72B + SAM3 pipeline, and our localizer with \method without preference tuning (see the supplementary material for details).

\paragraph{Measurements.}
Observers rated, on a 5-point scale, how well the red region localizes the artifacts, how well the artifacts are fixed, and how well the original video is preserved (see the supplementary material for the scale anchors).
We ask for fixing and preservation separately because a repair can remove the artifact while changing the rest of the video, or leave the video unchanged without removing the artifact.

\paragraph{Participants.}
A total of eleven observers (five female, four male, two undisclosed; aged 20 to 31 years, median 23) completed all 100 trials.
Observers were required to have normal or corrected-to-normal vision, did not know the purpose of the study, and took part as unpaid volunteers.
The study protocol was approved by the university's institutional review board.

\paragraph{Apparatus and procedure.}
Observers ran the experiment on a 2.5K display, read instruction pages (see the supplementary material), and completed two practice trials and the 100 main trials in a random order, with a break after every 20 trials.

%% file: sec/4c_results.tex
\section{Results}
\label{sec:res_loc}
\subsection{Localization}
\label{sec:study_results}

\input{figures/fig_mos_rows_localization_col}

Throughout the results, observer scores are SUREAL estimates~\cite{netflix_sureal} averaged over clips, and differences are tested with Holm-corrected Wilcoxon signed-rank tests over clips (details in the supplementary material).

We first compare the regions of our localizer with the widened dataset annotation and with the regions of the annotation-free 72B pipeline, holding the inpainter fixed (\cref{fig:mosloc}).
On the seven annotated clips, our regions receive a mean localization rating of $3.63$ and the annotation $3.94$, a difference that is not significant ($p=.38$).
Over all ten clips, our regions have a mean localization rating of $3.67$ against $3.33$ for the 72B pipeline and are rated higher on 9 of the 10 clips ($p=.020$).
We detect no significant differences in fixing or preservation among the repairs from the three regions (all $p>.3$).

With 45M parameters, our localizer is the smallest of the localizers we compared, about 1,600 times smaller than the 72B pipeline it learns from and about half the size of the next smallest, DiffDoctor (85M).
It predicts a region in 1.2\,s per 5-second clip on one H200 GPU, more than 100 times faster than the 72B pipeline, which takes about 140\,s on two H200 GPUs.
Coverage, region size and the comparison with every localizer are in the supplementary material.

\subsection{Repair quality}
\label{sec:main}
\input{figures/fig_frames}

\paragraph{Artifact removal.}
\Cref{fig:frames} shows repairs frame by frame, and the supplementary material plots each method by its two ratings and gives the ratings of every clip.
The method affects the fixing rating (Friedman, $\chi^2(6)=40.6$, $p<.001$).
\method has the highest mean rating ($\mu=3.63$).
It is rated significantly above the regenerations with Wan2.2 ($\mu=2.37$) and VACE-14B ($\mu=2.06$), above EditCtrl ($\mu=1.36$), and above the pipeline without the reference cut-out ($\mu=1.84$), each on all 10 clips ($p=.012$).
Against Gemini Omni 1.1 Flash, \method has a higher mean rating ($3.63$ against $2.95$) and is rated higher on 8 of the 10 clips.
With ten clips, this difference does not reach significance after correction ($p=.064$).

\paragraph{Preservation.}
The method also affects the preservation rating (Friedman, $\chi^2(6)=43.5$, $p<.001$).
\method has the highest mean preservation rating ($\mu=4.36$), significantly above every other method except the same pipeline without tuning.
Against Gemini Omni 1.1 Flash, it is rated higher on 9 of the 10 clips ($4.36$ against $3.40$, $p=.012$).
Regeneration with open models changes the video ($\mu=1.48$ for Wan2.2 and $1.52$ for VACE-14B).
Without the reference cut-out, the untuned pipeline is rated far lower than with it ($1.70$ against $4.24$, $p=.010$), which supports the role of the cut-out in keeping the appearance of the object.

\subsection{Preference tuning}
\label{sec:res_pref}

Compared with the same pipeline without preference tuning, observers rate \method higher at fixing on 8 of the 10 clips ($3.63$ against $3.32$, $p=.039$).
Preservation shows no detected decrease, with a higher mean ($4.36$ against $4.24$) and higher ratings on 8 of the 10 clips, but the difference is not significant ($p=.084$).
On the 529 clips of the example pool with a predicted region, which played no part in training or tuning, the judge $\Delta$ drops from $+0.49$ to $+0.33$ ($p<.001$), and it drops on the test clips as well (see the supplementary material).
Picking the best of four untuned samples by Jev score reaches a lower judge $\Delta$ ($0.00$, against $+0.41$ for a random sample) but needs four repairs per clip.

\subsection{Efficiency and components}

\input{tables/t4_ladder}

On one H200, region-sparse denoising lowers the inpainting time on our regions from 125\,s to 77\,s per clip, with 60\% of the tokens processed (\cref{tab:ladder}), and the share falls to 42\% for regions under 15\% of the frame (see the supplementary material).
With preference tuning, a repair takes 83\,s plus 1.2\,s for localization, at a peak GPU memory of about 42\,GB.
\Cref{tab:ladder} adds the components one at a time.
Hiding the region without a reference loses the object, and observers rate both fixing and preservation low ($1.84$ and $1.70$).
Adding the reference cut-out, region-sparse denoising and edge blending restores preservation (DINOv2 $0.906$, observers $4.24$) at 60\% of the dense compute, and preference tuning further raises the fixing rating to $3.63$. Dense denoising gives the same preservation at $125$\,s instead of $77$\,s per clip (see the supplementary material for every component).

%% file: figures/fig_mos_rows_localization_col.tex
\begin{figure}[t]
\centering
\includegraphics[width=\linewidth]{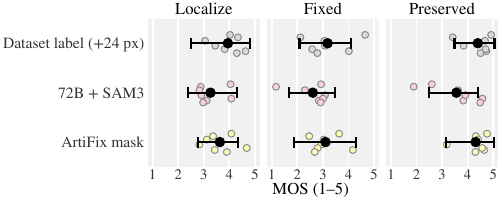}
\caption{Ratings of the three localizations on the 7 annotated clips, each repaired by \method without preference tuning. The comparison of our localizer with the 72B pipeline on all 10 clips is reported in the text. Dots are clips, and black marks the mean with its 95\% confidence interval (SUREAL MOS, eleven observers).}
\label{fig:mosloc}
\end{figure}

%% file: figures/fig_frames.tex
\begin{figure*}[t]
\centering
\includegraphics[width=\textwidth]{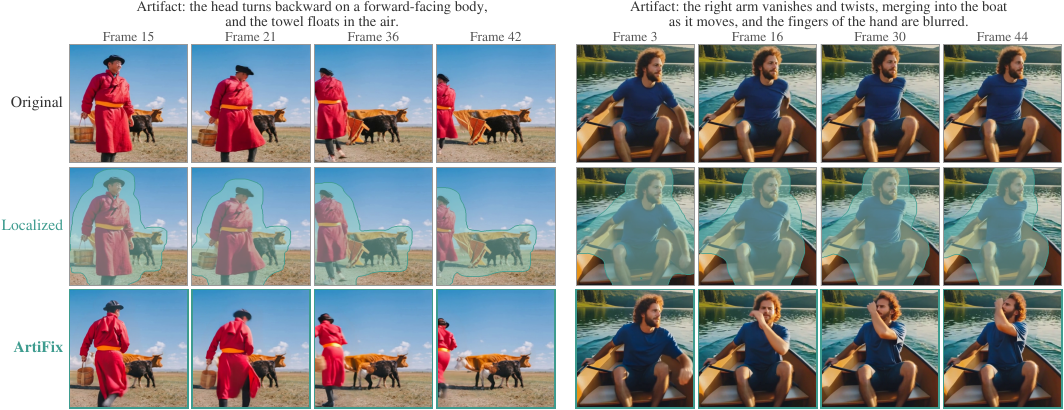}
\caption{Frame-by-frame repairs on two zoomed-in BrokenVideos test clips. Top, the generated video. Middle, the region of our localizer. Bottom, the \method repair, which keeps every pixel outside the region.}
\label{fig:frames}
\end{figure*}

%% file: tables/t4_ladder.tex

\begin{table}[t]
\centering\footnotesize
\setlength{\tabcolsep}{2.6pt}
\resizebox{\linewidth}{!}{%
\begin{tabular}{lrcccc}
\toprule
 & s/clip & DiT & DINOv2$\uparrow$ & Fixed$\uparrow$ & Preserved$\uparrow$ \\
\midrule
Region hidden, no reference & 115 & 100\% & 0.591 & 1.84 & 1.70 \\
+ reference, sparse, edge blend & \textbf{77} & \textbf{60\%} & \textbf{0.906} & 3.32 & 4.24 \\
+ preference tuning (\textbf{\method}) & 83 & \textbf{60\%} & \textbf{0.906} & \textbf{3.63} & \textbf{4.36} \\
\midrule
\emph{dense instead of sparse} & 125 & 100\% & \textbf{0.907} & -- & -- \\
\bottomrule
\end{tabular}%
}
\caption{Components of \method on test clips. Each row adds components to the row above. DiT is the share of tokens computed, DINOv2 measures preservation, and Fixed and Preserved are observer MOS on the ten study clips. Full ladder in the supplement.}
\label{tab:ladder}
\end{table}

%% file: sec/5_discussion.tex
\section{Discussion and limitations}
\label{sec:discussion}

\paragraph{Local repair as an alternative to regeneration.}
The study supports local repair whenever the generated video should be kept.
\method was rated above every evaluated open method for both removal and preservation, supporting H1 against the open regenerators.
Against the commercial Omni, it preserved the video significantly better, and its removal rating was higher on average and more consistent across clips, although not significantly with ten clips.
Even at strength $0.85$, regeneration lets the scene drift (DINOv2 $0.69$ to $0.71$), whereas \method copies every pixel outside the region.
The commercial model is a strong reference rather than a ground-truth repair, and an open pipeline whose removal ratings trend above it while preserving more makes local repair a practical alternative to regeneration.

\paragraph{Regions and preference tuning.}
The localization results support predicting the region that a repair needs rather than reproducing the visibly broken pixels. 
Comparing samples of the same repair and rejecting pairs whose preferred sample has a lower CLIP similarity to the input may keep the tuning from rewriting the region, although we do not isolate these effects (see the supplementary material).

\paragraph{Evaluation.}
\label{sec:auto_vs_obs}
The preservation metrics track the observers' preservation ratings (Spearman $\rho=0.58$ to $0.83$), but none of the evaluated measures tracks removal well enough to replace observers (VLM judge $\rho=0.43$, see the supplementary material).
We therefore use automatic scores to rank samples of the same repair rather than to compare methods.

\paragraph{Limitations.}
A generated video has no paired clean counterpart and a defect can admit several valid repairs, so training relies on proxies and the fixing rating measures how convincingly an artifact is removed rather than agreement with a target.
The pipeline repairs one 49-frame window at $832\times480$, fast complex motion remains hard for the inpainter, and large regions reduce the savings of sparse denoising.
The reference frame can contain part of the defect, and the localizer proposes a region on 12\% of real videos without artifacts.
The study used ten randomly drawn clips accepted by the commercial model and eleven observers, a size bounded by the 100 trials each observer completed, and the region overlay revealed whether a method edited a region or the whole frame.

%% file: sec/5_conclusion.tex
\section{Conclusion}
\label{sec:conclusion}

We presented \method, a pipeline that repairs a generated video from the video alone by redrawing only the region that contains an artifact.
It predicts the region a repair needs, hides it from the inpainter while a reference cut-out keeps the appearance of the object, denoises only the tokens near it, and is tuned on preferences between its own samples.
Observers rated it significantly above every evaluated open inpainting and regeneration method for both removal and preservation.
Against the commercial Gemini Omni 1.1 Flash, it preserved the video significantly better and received a higher mean removal rating, which was not significant with ten clips.
Repair during generation and across longer videos are natural next steps, for which the small localizer and region-sparse denoising are a starting point.

%% file: sec/llm.tex
\section*{LLM Usage Statement}
We used LLMs to revise and improve the manuscript writing and develop code for part of the algorithm.

%% file: sec/X_suppl.tex
\clearpage
\setcounter{page}{1}
\maketitlesupplementary

\startcontents[supp]
\printcontents[supp]{}{1}[2]{\section*{Contents}}


\section{Repairs judged without the region overlay}
\label{sec:supp_pairwise}
The perceptual study of the main paper shows the edited region so that observers can judge it, and to remove any confounding effect of this overlay on the comparison of repairs, we ran a blind follow-up study.
The follow-up is a two-alternative forced-choice (2AFC) experiment on 30 clips that the main study did not use, drawn at random from our example pool of generated videos with artifacts among those that Omni accepted, with six new observers (two female, three male, one undisclosed; aged 19 to 28 years, median 23, one undisclosed), none of whom took part in the main study.
In each trial, observers see the original next to two repairs of the same clip, labeled only A and B and shown without any overlay, and choose which repair removes the artifact better (\emph{fixed}) and which keeps the rest of the video unchanged (\emph{preserved}).
\input{figures/supp_block_pairapp}
Every observer compared \method, Omni and Wan2.2 with each other on every clip and each repair with the unrepaired original on a third of the clips, in an order chosen by ASAP active sampling~\cite{mikhailiuk2021active}.
The six observers completed 720 such trials and gave 1{,}440 judgments.

\paragraph{Statistical analysis.}
We scale the choices with pwcmp~\cite{perezortiz2017practicalguidesoftwareanalysing} (Thurstone Case V) in just-objectionable-difference (JOD) units, where a difference of 1\,JOD means that 75\% of observers prefer the better condition.
The scale is fixed at 0 for the unrepaired original on \emph{fixed} and for \method on \emph{preserved}, since the original cannot lose on preservation.
Confidence intervals (95\%) come from 2{,}000 bootstrap resamples of the observers, with the scale refitted each time.
Two-tailed $p$-values use the bootstrap covariance of each difference and are Holm-adjusted over the three repair contrasts of each question, with the three contrasts against the original as their own family.

\paragraph{Results.}

\begin{figure}[t]
\centering
\includegraphics[width=\linewidth]{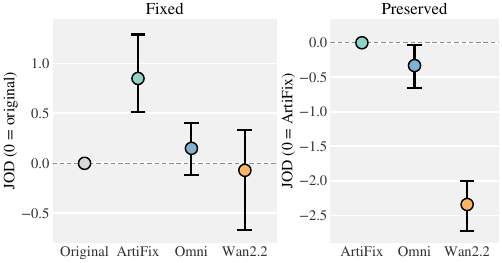}
\caption{\textbf{Without the overlay, observers still prefer \method on both questions} (six observers, 30 clips). JOD with 95\% confidence intervals over observers, where 0 is the unrepaired original for fixing and \method for preservation.}
\label{fig:supp_pair_jod}
\end{figure}

\Cref{fig:supp_pair_jod} shows that the blind comparison supports the main study.
\method is the only repair that observers prefer over leaving the artifact in place ($+0.85$\,JOD), and it is preferred over Omni on fixing by $0.70$\,JOD ($p=.002$), consistent with the higher fixing rating of \method in the main study.
Without the overlay, \method also stays ahead on preservation, by $0.33$\,JOD over Omni ($p=.045$) and by $2.34$\,JOD over Wan2.2.
Against the unrepaired original, observers chose \method as better fixed in 48 of 60 trials (80\%), on 8 of 10 clips, while Omni (27 of 60) and Wan2.2 (29 of 60) stayed at chance.
On fixing, \method is also preferred over Wan2.2 by $0.92$\,JOD ($p=.041$).

\section{Repair quality across methods and measures}
\label{sec:supp_results}
This section extends the comparison of repair methods in the main paper to every method we evaluated and to further measures of what a repair changes.

\paragraph{All inpainting methods.}
\Cref{tab:main} compares every inpainting method on the 103 test clips with the automatic measures, together with the observer ratings for the methods in the perceptual study.
On the 63 clips that Omni returned, \method keeps a DINOv2 similarity of $0.914$ against $0.906$ for Omni.

\input{tables/t1_main}

\input{figures/supp_block_examples}

\paragraph{Effect of preference tuning.}
On the 103 test clips, preference tuning lowers the judge $\Delta$ from $+0.30$ to $+0.13$ (unadjusted Wilcoxon over clips, $p=.01$), with 49 instead of 41 clips rated cleaner than the input, while LPIPS ($0.122$ to $0.119$) and DINOv2 ($0.906$) change little.

\paragraph{Without preference tuning.}
The pipeline without preference tuning is rated significantly above EditCtrl, Wan2.2 and VACE-14B on fixing and preservation (Holm-adjusted $p\le.012$), and above Gemini Omni 1.1 Flash on preservation ($4.24$ against $3.40$, $p=.014$), which the automatic measures confirm (LPIPS $0.122$ against $0.158$ for Omni, \cref{tab:main}).


\paragraph{Identity inside the region.}
Preservation over the whole frame says little about the redrawn object itself, so we also compare the object inside our region with the input on the frames where the dataset annotation marks no artifact and the object should stay the same.
On the 13 test clips with at least three such frames, the DINOv2 similarity of the crop around the region is $0.879$ for \method and $0.876$ without preference tuning, against $0.736$ for VACE-14B regeneration, $0.704$ for Wan2.2 and $0.672$ for EditCtrl.
Omni returned 8 of these clips, on which \method reaches $0.889$ against $0.747$ for Omni.
\method thus keeps the identity of the object where it was already correct, while regeneration and editing change it.



\paragraph{Seed variance.}
Every reported repair uses one fixed seed, so we repeat the main methods with two further seeds on the 103 test clips (\cref{tab:supp_seeds}), the first seed being the run reported in the main paper.
The spread over seeds is far smaller than the differences between methods, with DINOv2 varying by less than $0.01$ for every method against a gap of about $0.2$ between \method and regeneration.
The gain of preference tuning holds over seeds (judge $\Delta$ $+0.15\pm0.04$ against $+0.43\pm0.17$ without preference tuning), and the tuned inpainter also varies less between seeds.

\begin{table}[t]
\centering\footnotesize
\setlength{\tabcolsep}{3pt}
\begin{tabular}{lccc}
\toprule
 & LPIPS$\downarrow$ & DINOv2$\uparrow$ & Judge $\Delta\downarrow$ \\
\midrule
\method & \textbf{0.121}{\scriptsize$\pm$0.001} & \textbf{0.905}{\scriptsize$\pm$0.003} & \textbf{+0.15}{\scriptsize$\pm$0.04} \\
\method w/o tuning & 0.124{\scriptsize$\pm$0.002} & 0.900{\scriptsize$\pm$0.006} & +0.43{\scriptsize$\pm$0.17} \\
VACE-14B regeneration & 0.372{\scriptsize$\pm$0.002} & 0.704{\scriptsize$\pm$0.008} & -- \\
Wan2.2-A14B regeneration & 0.392{\scriptsize$\pm$0.001} & 0.685{\scriptsize$\pm$0.008} & -- \\
\bottomrule
\end{tabular}
\caption{\textbf{Differences between seeds are far smaller than differences between methods} (mean and standard deviation over three seeds, 103 test clips). The judge was run for \method and the pipeline without preference tuning only.}
\label{tab:supp_seeds}
\end{table}

\paragraph{Example pool.}
The example pool holds 580 videos with artifacts and no annotation, 520 generated clips and 60 BrokenVideos clips, on which we check that the results on the test clips carry over.
On the 529 pool clips where our localizer finds a region, \method lowers the judge $\Delta$ of the pipeline without preference tuning from $+0.49$ to $+0.33$ ($p<.001$) and raises DINOv2 similarity from $0.880$ to $0.884$ ($p<.001$).
Of the 117 pool clips where Omni ran, our localizer finds no artifact in 13 and returns them unchanged.
On the other 104, the repair of the pipeline without preference tuning is judged cleaner than the input on 56, and on 26 of those it is closer to the original in per-clip FID than all four whole-video methods.
Regeneration with open models is far less faithful on these clips (median per-clip FID $141$ for Wan2.2, against $33$ for our pipeline and $28$ for Omni).

\section{What the localizer gives the inpainter}
The localizer decides what the inpainter redraws, so we report how much of the artifact its regions cover, where they miss part of it, and how it behaves on videos without artifacts.

\paragraph{All localizers.}
For localization we report coverage, the share of annotated artifact pixels inside the region, together with region area, since coverage alone favors larger masks.
Region similarity J and boundary accuracy F~\cite{perazzi2016benchmark} compare against visible-defect annotations, whereas our regions intentionally include the whole object.
A clean clip counts as a false positive if the post-processed region is non-empty.
The further localizers are the teacher pipeline with Qwen2.5-VL-32B, a localizer head on SAM3, Skyra~\cite{li2026skyra}, DiffDoctor~\cite{wang2025diffdoctor}, and SAMWISE~\cite{cuttano2025samwise} fine-tuned on BrokenVideos.
Several of them miss the artifact entirely on many test clips, which rules them out as the first stage of a repair.
DiffDoctor predicts an empty region on 67 of the 103 test clips (65\%) and Skyra on 25 (24\%), and SAMWISE never returns an empty region but its median region covers only 9\% of the frame against 32\% for our localizer.
\Cref{tab:supp_localization} extends the localization results of the main paper to every localizer we evaluated, with region similarity J, boundary accuracy F, coverage, and frame-level detection on clean videos.
J and F reward tight masks around the visibly broken pixels, so they penalize regions that cover the whole object, and every object-level region scores F between $0.05$ and $0.07$.
SAMWISE, trained to reproduce the masks, scores F $0.34$ but covers only 72\% of the artifact and gives the worst repairs among the localizers with high coverage (judge $\Delta$ $+0.55$).
Image-level artifact detectors transfer poorly to video, as DiffDoctor marks almost nothing (coverage $0.03$) and the boxes of Skyra rarely cover the annotated artifact ($0.29$).
The SAM3 localizer head and SAMWISE propose a region on more than 99\% of the 300 clean videos, against 13\% for ours, while DiffDoctor and Skyra propose fewer (9 and 10\%) but also miss most artifacts.
Counting a frame as detected when the region is non-empty, our localizer reaches a frame-level F1 of $0.92$, against $0.50$ to $0.62$ for the other localizers.

\input{tables/t3_localization_full}

\paragraph{Localization in detail.}
Our localizer covers 97\% of the annotated artifact pixels, as much as the 72B pipeline, with regions of a similar size (31 to 33\% of the frame).
The dataset labels mark only the visibly broken pixels, not the objects that carry the artifact, while our regions cover the whole object, so J\&F against the labels is low ($0.13$) by design.
What matters is the repair the region leads to. With the same inpainter, our regions give a lower judge $\Delta$ than the dataset annotation ($+0.30$ against $+0.38$) and the 72B pipeline ($+0.48$), and observers rate the repairs from the three similarly on fixing and preservation (Sec.~5.1 of the main paper).
With the inpainter without preference tuning, its regions yield a lower judge $\Delta$ than those of the 72B pipeline on the 103 test clips ($+0.30$ against $+0.48$).


\paragraph{False positives on videos without artifacts.}
The localizer proposes a region on 38 of the 300 held-out real videos without artifacts (13\%).
A false region does not make such a video worse, because the inpainter redraws the region from the same video with a cut-out of the same objects as reference, and every pixel outside the region is copied from the input.
We measure this with the preservation measures, which track the observers' preservation ratings (Sec.~6 of the main paper).
Over the 38 videos, the repairs keep a DINOv2 similarity of $0.925$ to the input over the frame (median $0.961$) and $0.906$ inside the false region, as close as the repairs of real artifacts on the test clips ($0.906$) and far from regeneration (Wan2.2 $0.694$), and \cref{fig:supp_fp} shows eight of them.

\begin{figure*}[t]
\centering
\input{figures/grid_fp}
\caption{\textbf{A false region can leave a video without artifacts unchanged} (the eight of the 38 held-out real videos without artifacts that \method preserves best). The input with the false region outlined is shown next to the output of \method, at the frame where the region is largest, with the share of the frame covered by the false region on the input and the DINOv2 similarity between the output and the input on the output.}
\label{fig:supp_fp}
\end{figure*}

\section{Where the gains come from}
This section attributes the preservation, the cleanliness and the speed of \method to its components, and measures the choices behind each of them on the 103 test clips.

\paragraph{Components.}
\input{tables/s6_ladder_full}

Table~2 of the main paper combines the reference cut-out, region-sparse denoising and edge blending in one row, because observers rated only that combination.
\Cref{tab:supp_ladder} separates them with the automatic measures.
The reference cut-out accounts for nearly all of the gain in preservation (DINOv2 from $0.591$ to $0.907$), while region-sparse denoising and edge blending leave it unchanged ($0.905$ and $0.906$).
Dense denoising is the same local repair computed on every token rather than a regeneration, since it also edits only the region and keeps every pixel outside it.
It gives the same result as region-sparse denoising (DINOv2 $0.907$ against $0.905$, judge $+0.33$ for both) at $125$ instead of $77$\,s per clip, so it is a row of cost rather than of quality, and the observer ratings of the region-sparse pipeline apply to it.

\paragraph{Reference frame.}
The frame where the region is largest shows the most of the object, but it can also show the artifact at its worst.
We therefore compare this rule with two alternatives on the 103 test clips, the earliest frame of the window and the frame where the artifact probability of our localizer inside the region is lowest, with the same model, seed and settings (\cref{tab:supp_refframe}).
The preservation measures show how much of the object's appearance a cut-out keeps, and the judge $\Delta$ shows whether it carries the artifact into the repair.
Our rule is best on both, so the larger cut-out does not bring the artifact back.
The earliest frame often shows the object only in part or from another angle, and preservation drops (DINOv2 $0.893$, lower on 62 of 103 clips, $p=.031$).
The cleanest frame does not change the judge ($p=.68$) but lowers preservation (DINOv2 $0.889$, lower on 68 clips, $p=.008$), since a frame with little visible artifact often shows little of the object.

\begin{table}[t]
\centering\footnotesize
\setlength{\tabcolsep}{3pt}
\begin{tabular}{lcccc}
\toprule
Reference frame & Judge $\Delta\downarrow$ & DINOv2$\uparrow$ & LPIPS$\downarrow$ & FID$\downarrow$ \\
\midrule
Largest region, middle half (ours) & \textbf{+0.13} & \textbf{0.906} & \textbf{0.119} & \textbf{37.3} \\
Earliest frame & +0.29 & 0.893 & 0.128 & 41.2 \\
Lowest artifact probability & +0.25 & 0.889 & 0.127 & 40.7 \\
\bottomrule
\end{tabular}
\caption{\textbf{The frame with the largest region gives the best reference cut-out} (\method on the 103 test clips, same model, seed and settings).}
\label{tab:supp_refframe}
\end{table}

\paragraph{Ablations.}
\Cref{tab:ablations} varies one design choice at a time on the 103 test clips.
A physics description of the artifact written by the 72B model performs like the generic prompt (judge $\Delta$ $+0.21$ for both, DINOv2 $0.907$ against $0.904$, paired Wilcoxon over clips $p=.54$), so we keep the generic prompt, which needs no per-clip description and lets its embedding be cached.
A reward built from CLIP similarity and the residual artifact estimate of the localizer makes the judge rate the repairs worse ($+0.68$ against $+0.30$ without preference tuning), so the choice of Video Jev as the ranking signal matters.

\input{tables/t5_ablations}





\paragraph{Prompt embedding.}
The prompt is the same for every clip, so we compute its umT5 embedding once and store it.
The 5.7B-parameter text encoder is then never loaded, and a repair fits on one 48\,GB GPU (\cref{tab:supp_cache}, three test clips on one H200).

\begin{table}[t]
\centering\small
\resizebox{\linewidth}{!}{%
\begin{tabular}{lcc}
\toprule
 & Text encoder & Cached embedding \\
\midrule
Parameters loaded & 23.1B & 17.5B \\
GPU memory after loading & 43.5\,GiB & 33.0\,GiB \\
Peak GPU memory of a repair & 52.5\,GiB & 41.9\,GiB \\
\bottomrule
\end{tabular}}
\caption{\textbf{Caching the fixed prompt embedding saves about 10\,GiB of GPU memory} (three test clips, one H200). The 5.7B-parameter umT5 text encoder is not loaded, and the outputs differ from those with the text encoder by 0.8 to 1.0 gray levels on average, as much as two runs with the text encoder differ from each other.}
\label{tab:supp_cache}
\end{table}

\paragraph{Video Jev as a measure.}
The main comparison reports the Video Jev score of every method (Table~1 of the main paper).
Among samples of the same repair, the Jev score picks better repairs, since choosing the best of four seeds of the pipeline without preference tuning by the Jev score lowers the judge $\Delta$ from $+0.41$ to $0.00$ at no cost in preservation (\cref{tab:supp_jev}), at four times the cost of one repair.
Ranking samples of one repair with a video realism model makes preference tuning possible without human preference labels, which is a new and promising way to improve a repair model, and better questions to the model may strengthen this signal further.

\input{tables/s5_jev}

\input{figures/supp_block_jev}

\input{tables/t6_dpo_sweep}

\paragraph{Preference-tuning hyperparameters.}
\label{sec:supp_sweep}
To see how much the gain of preference tuning depends on its settings, we train the LoRA with ranks from 8 to 64 and learning rates from $4\times10^{-6}$ to $1\times10^{-4}$, on the same 178 pairs for 200 steps each (\cref{tab:dposweep}).
The learning rate matters most.
At $4\times10^{-6}$ the LoRA barely changes the model within 200 steps, and the repairs stay as clean as without preference tuning (judge $\Delta$ $+0.23$ to $+0.33$).
At $1\times10^{-4}$ the training becomes unstable as the rank grows, and at rank 64 it destroys the repairs (judge $\Delta$ $+3.84$, DINOv2 $0.58$).
At $2\times10^{-5}$ every rank from 8 to 64 makes the repairs cleaner than without preference tuning (judge $\Delta$ $+0.10$ to $+0.21$ against $+0.30$, 42 to 53 instead of 41 clips cleaner than the input), at unchanged preservation (LPIPS $0.118$ to $0.119$, DINOv2 $0.906$ to $0.909$).
The differences between these ranks are not significant (Wilcoxon over clips against rank 32, $p=.42$ to $.66$), so the gain does not depend on a finely tuned rank.
Further experiments show that smaller ranks keep the gain, and rank 16 at this learning rate even reaches a slightly lower judge $\Delta$ than our rank 32 ($+0.10$ against $+0.13$) with half the LoRA parameters, so the LoRA can be made smaller, and a further calibration of these hyperparameters can improve \method further.

\paragraph{Region-sparse denoising.}
Region-sparse denoising computes only the tokens that the repair needs and keeps all others on the trajectory of the original.
A latent token of Wan2.1 covers $16\times16$ pixels and four frames, and the first latent frame covers the first video frame alone.
The region is mapped to tokens by marking a token whenever any of its pixels is in $M$, and the active set is this token mask grown by one token in space.
The context band is two tokens around the active set in space and one latent frame before and after it in time, and the global grid is every fourth token in height and width on every latent frame.
The tokens of the reference cut-out are always kept and redrawn.
The predictions of the band and grid tokens are discarded, and these tokens, like all skipped video tokens, receive the exact velocity $\epsilon - z_0$.
Under classifier-free guidance the conditional and unconditional velocities of these tokens are equal, so guidance leaves them on the noised original as well.
When more than 85\% of the tokens would be kept, gathering and scattering them costs more than the skipped tokens save, and we use dense inference, in which the model predicts every token and only Eq.~(2) of the main paper preserves the pixels outside the region.
The band, the grid and the reference tokens add a fixed cost, so the savings are smaller than the share of the frame outside the region (\cref{tab:efficiency}), and the dense fallback was reached only for regions above half of the frame.

\input{tables/t2_efficiency}


\section{How the perceptual study was run}
\label{sec:supp_study}
This section describes how the perceptual study of the main paper was run and analyzed.

\subsection{Experiment application}
\begin{figure*}[t]
\centering
\newcommand{\appshot}[1]{\IfFileExists{figures/assets/#1}{\includegraphics[width=0.32\textwidth]{figures/assets/#1}}{\fbox{\parbox[c][0.18\textwidth][c]{0.3\textwidth}{\centering\scriptsize #1}}}}
\appshot{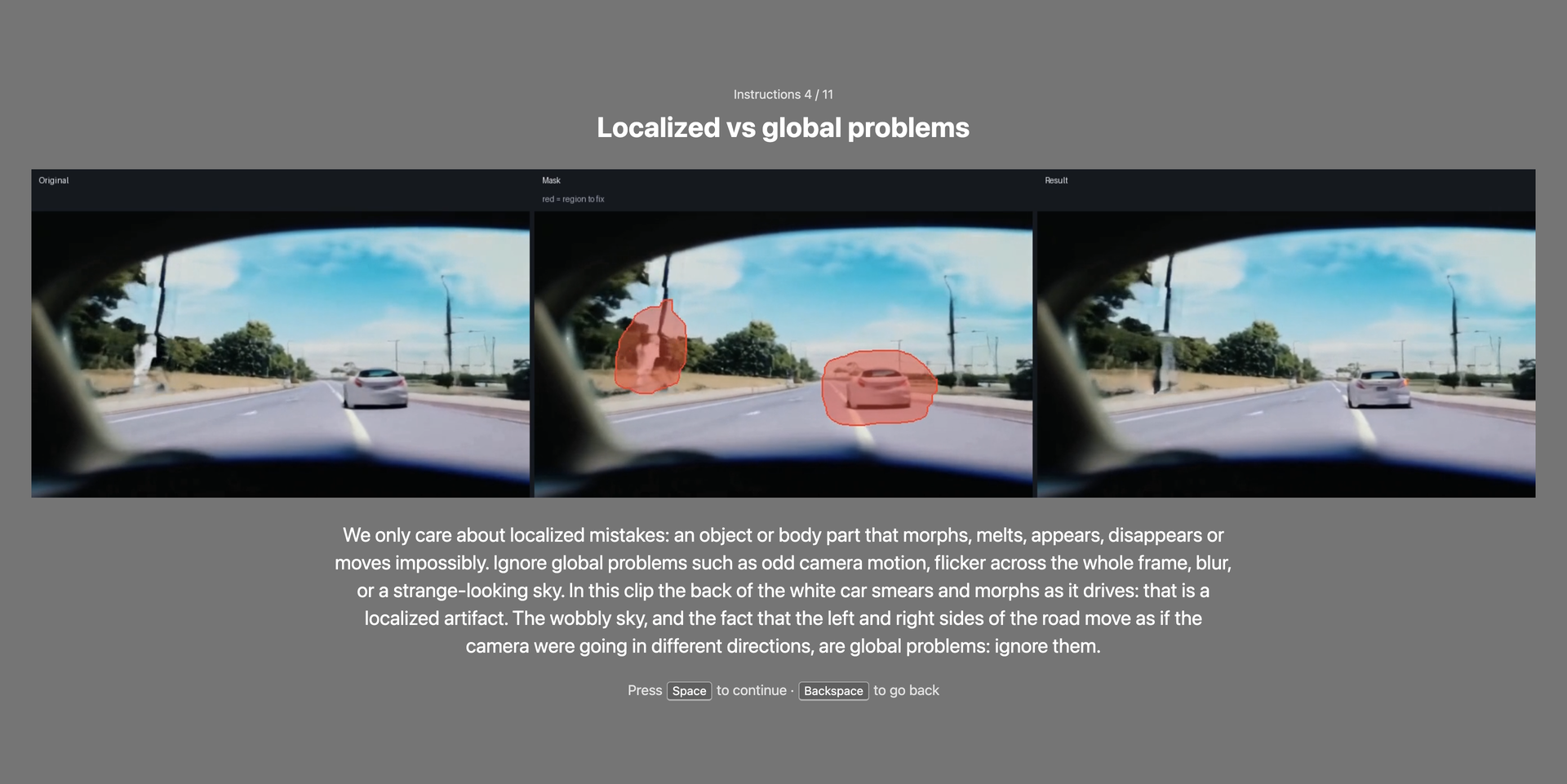}\hfill
\appshot{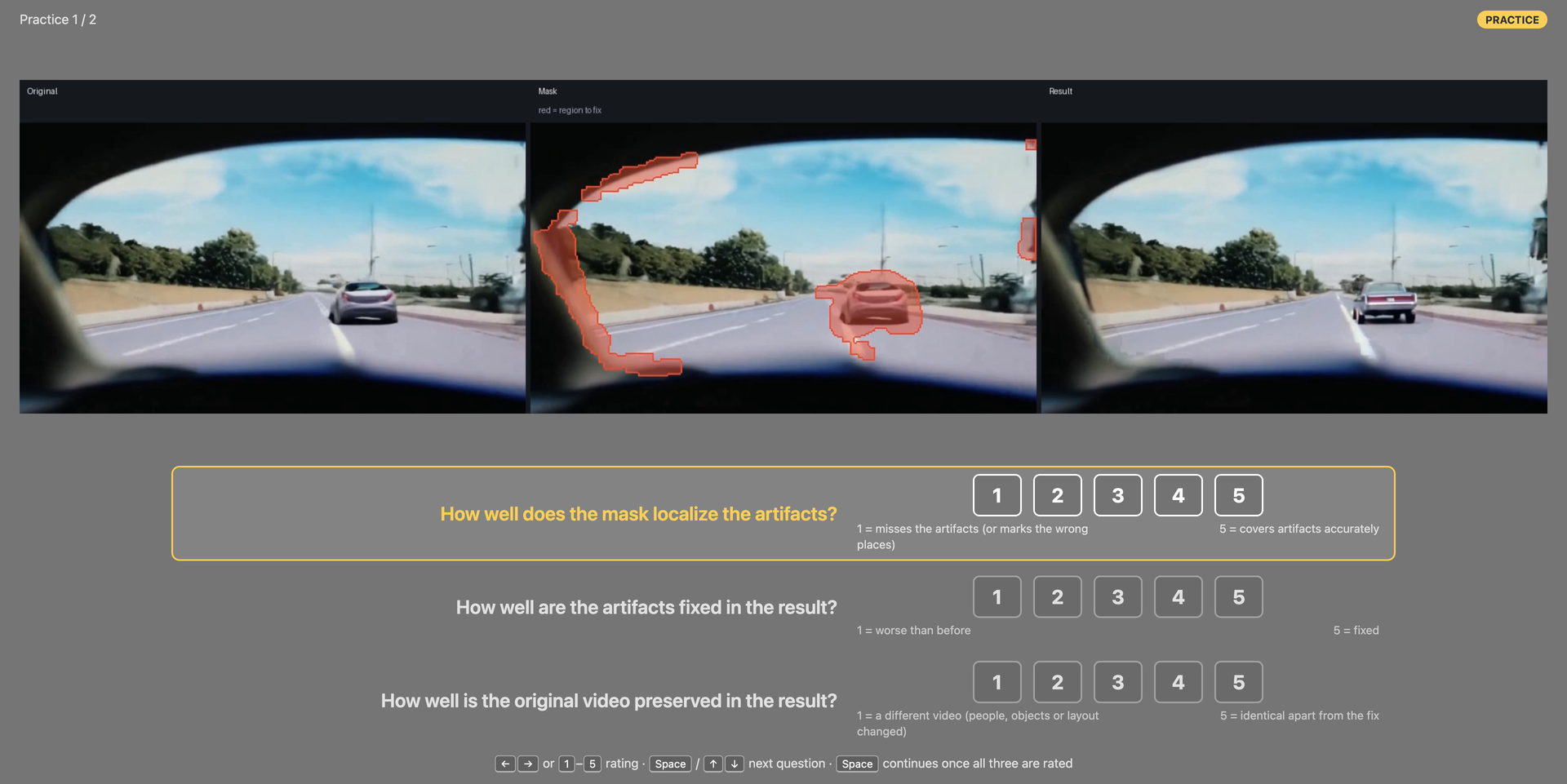}\hfill
\appshot{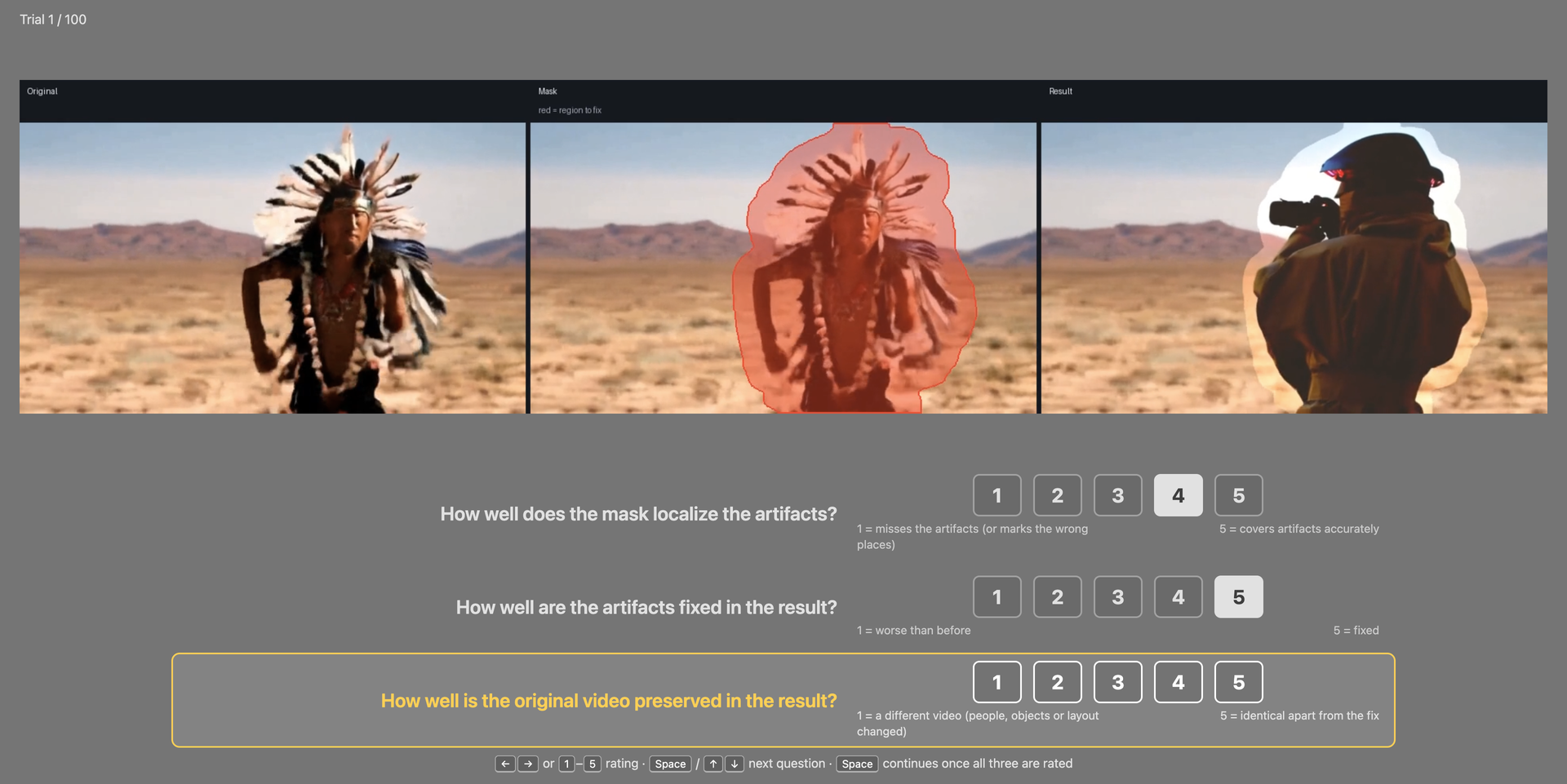}
\caption{\textbf{The perceptual study shows the edited region as a red overlay.} One instruction page, a practice trial, and a main trial with the original, the edited region in red and the result side by side, and the three statements rated with the arrow keys.}
\label{fig:supp_app}
\end{figure*}

Observers rated the videos in a full-screen desktop application (\cref{fig:supp_app}), which leads them through illustrated instructions with worked examples, two practice trials, and the 100 main trials in a random order per observer.

\paragraph{Instructions.}
The instructions teach observers to separate localized artifacts, which the study is about, from global problems, with two worked examples.
In the first (an Open-Sora~1.2 clip of a white car), the sky changes over the whole video, which is a global problem observers are told to ignore, while the car carries a localized artifact.
In the second (a CogVideoX-2B clip of a climber), the climber morphs into the wall, and the correct mask therefore includes the part of the wall that morphs with the climber.
A third example shows three repairs of one clip that preserve the video without fixing the artifact, fix the artifact without preserving the video, and do both.

\paragraph{Procedure.}
The experiment began with the consent screen and optional demographic questions.
Two practice trials followed and were not analyzed.
During a trial, observers moved between the three statements and set each rating with the arrow keys, and the space bar continued to the next trial once all three were rated and the videos had played at least once.


\subsection{Stimuli}
Each trial shows the three videos side by side at their native size of $640\times368$ pixels each.
Many artifacts become visible only in motion, so the three videos play in sync at the source frame rate (8 to 24\,fps) and loop three times.
The BrokenVideos masks leave out the parts of a morphing object around the broken pixels (\cref{fig:patterns}), so the dataset annotation is widened by 24 pixels before inpainting.
The two margins differ because the two masks differ. Our region map already covers the whole affected object, so its 12-pixel margin only pads the border of the region. The annotation marks only the visibly broken pixels, so it receives the 24-pixel margin that the teacher also adds around annotated pixels when it builds the region labels, which makes the annotation a stronger reference rather than a weaker one.
\input{figures/fig_patterns}
The localization trial with our localizer repeats the video of the condition without preference tuning, and the two trials are analyzed separately within their own comparisons.
The three clips from our example pool have no labels, so their dataset-annotation trial shows our mask and is excluded from the comparison with the annotation, which uses seven clips.
Observers rated how well the red region localizes the artifacts, from 1 (misses them or marks the wrong places) to 5 (covers every artifact and its remnants), how well the artifacts are fixed, from 1 (not fixed or worse) to 5 (completely fixed), and how well the original video is preserved, from 1 (a different video) to 5 (identical apart from the fix).

\paragraph{Region overlay.}
The center video shows the edited region in red, because the localization statement asks observers to judge that region, and because the preservation statement asks whether the video outside the intended edit is unchanged, which observers can judge only when they know where the edit is.
Without the overlay, observers would have to find the artifact within a few seconds of looping video before judging its repair.
The overlay also reveals that the regenerators edit the whole frame, which could bias the preservation ratings toward local methods, and the blind follow-up (\cref{sec:supp_pairwise}) removes the overlay and finds the same order.
The automatic preservation measures, which do not see the overlay, rank the methods in the same order (Table~1 of the main paper).

\subsection{Analysis}
\paragraph{Statistical analysis.}
SUREAL~\cite{netflix_sureal} is fit separately to the ratings of each statement.
Confidence intervals (95\%) come from a two-level bootstrap that resamples observers, refits SUREAL, and then resamples clips (1000 replicates).
A Friedman test across the seven repair-method conditions first tests whether the method affects each statement, and post-hoc pairwise comparisons use Wilcoxon signed-rank tests on the per-clip scores.
Significance is judged on the Holm-adjusted p-value at $\alpha=0.05$.

\paragraph{Comparison families.}
All p-values are Holm-adjusted within each statement and each family of comparisons.
The first family compares \method with each of the other six inpainting conditions.
The second compares the pipeline without preference tuning with the other five inpainting conditions without preference tuning, after its own Friedman test across all six.
The third compares the dataset annotation and the 72B pipeline with our localizer within the localization trials.

\paragraph{Study size and power.}
The study has eleven observers and ten clips, and every observer rated every clip under every condition, which gives 1,100 ratings in about 45 minutes per observer and lets the paired tests compare conditions within clips and within observers.
Studies with more observers and clips would add power and generality, and the blind follow-up (\cref{sec:supp_pairwise}) adds 30 further clips for the comparison with Omni.

\subsection{All conditions}
\Cref{tab:supp_mos} gives the ratings of all ten conditions on all three statements, including the localization conditions that the main paper shows only on the seven labeled clips.
The regenerators edit the whole frame, so their regions receive localization ratings of $1.02$ to $1.05$.

\input{tables/s3_study_mos}

\subsection{Automatic measures versus observers}
\label{sec:supp_auto}
Observer studies are costly, so we test how well the automatic measures reproduce the ratings, which decides how far they can stand in for observers.
For each of the 80 distinct trials without preference tuning we have the mean rating per statement.
For the same clip and method we have the artifact severity that the Qwen2.5-VL-72B judge assigns to the original and to the result, judged on a common view for all methods, and the preservation metrics LPIPS, FID per clip, CLIP and DINOv2 similarity.
\Cref{tab:supp_auto} reports rank correlations across trials and how often a measure orders two methods on the same clip the way the observers do.

The judge agrees with the observers on the clear failures and disagrees on the cases that decide the comparison.
Across the 80 trials its rated improvement correlates only moderately with the fixing rating (Spearman $\rho=0.43$) and weakly with preservation ($\rho=0.31$).
On pairs of methods applied to the same clip, it orders fixing as the observers do 68\% of the time and preservation 64\% of the time, where chance is 50\%.
Averaged per method it ranks EditCtrl and the pipeline without the reference last, like the observers.
It ranks Gemini Omni 1.1 Flash as cleaner than our pipeline without preference tuning ($-1.09$ against $-0.72$) and Wan2.2 close to it, while observers rate our pipeline higher on fixing ($3.32$ against $2.95$ for Omni and $2.37$ for Wan2.2).
The judge rates each video on its own and sees a clean-looking video.
Observers compare the result with the original, and they do not count a repair as fixed when the scene has become a different video or still contains the defect in a new form.

LPIPS, FID per clip, CLIP and DINOv2 similarity track the preservation rating well ($\rho=0.58$ to $0.83$, 80 to 88\% of same-clip pairs in the observers' order).
In this study they also correlate with the fixing rating ($\rho=0.28$ to $0.66$), because the methods that rewrite the video are also those that observers judge as not fixed.
This correlation does not make them measures of fixing, since they cannot see the artifact.

\input{tables/s4_auto_human}

\section{Implementation details}
\label{sec:supp_impl}
This section gives the settings needed to reproduce \method, in the order of the method section.


\paragraph{Localizer.}
We train the localizer for 16k steps on clips of 16 frames, two clips per batch, with AdamW (learning rate $2\times10^{-4}$, weight decay $0.05$), a linear warm-up over 300 steps and cosine decay.
Its hidden size is 256, with four space-time mixer layers, three query-decoder layers and four queries per output map.
The SAM2.1 encoder gives features at strides 4, 8 and 16 of the $512\times512$ input. The space-time mixer runs on the stride-16 tokens, the pixel decoder adds the stride-8 and stride-4 features, and the output maps are predicted at stride 4.
We hold out 150 BrokenVideos clips for validation and 300 real clips from DeeptraceReward to measure false positives.

\paragraph{Region labels.}
The region map learns to cover whole objects, so its labels come from the objects that the 72B teacher names for each artifact (\cref{sec:supp_prompts}).
In each frame, SAM3 keeps the instance of each named object that overlaps the annotation most, and the label is the union of these object masks over a window of $\pm w$ frames, the artifact annotation, and the artifact map of an earlier SAM3-based localizer head, each enlarged by $m$ pixels.
A large object, such as a table or the floor, can make the label cover most of the frame, and the localizer would then learn to redraw large objects that are correct.
We therefore try four increasingly restrictive settings, $(w, m) = (6, 24)$, $(3, 24)$, $(3, 12)$ and $(0, 12)$, with $w$ in frames and $m$ in pixels, and keep the first one whose label covers at most 15\% of the pixels of the clip.
The budget is soft, since a label that still exceeds it after the last setting is kept at that setting and no object is removed.
Because the annotation is part of the union in every setting, tightening only narrows its margin and never removes an annotated pixel, and a label becomes the whole frame only when one object alone covers more than 70\% of it.
The median label covers 27\% of the pixels of its clip, and the teacher labels 834 clips, 790 for training and 44 for validation.

\paragraph{Localizer loss.}
The labels $y_m$ are the share of each $4\times4$ cell that is labeled, and each map is trained with
\begin{equation}
\mathcal{L}_m =
\begin{cases}
\mathrm{BCE}_w(p_m, y_m) + \mathrm{Dice}(p_m, y_m) & \text{defect video,}\\
\mathrm{BCE}(p_m, \mathbf{0}) & \text{clean video,}
\end{cases}
\label{eq:locloss}
\end{equation}
and the total loss is $\mathcal{L}_{\mathrm{art}} + \mathcal{L}_{\mathrm{reg}}$.
Here $\mathrm{BCE}_w$ is the binary cross entropy with a weight $w$ for each pixel.
Frames without an annotation are excluded from the loss of the artifact map.
In annotated frames, labeled pixels get $w=1$ and the remaining pixels get $w=0.3$, which reduces the penalty for marking broken pixels that the annotators missed.
The region map is trained with $w=1$ on every frame of the clips that the teacher labeled, and it receives no loss on the other clips.
On the real videos without artifacts, BCE alone at $w=1$ supervises empty predictions for both maps, since the Dice loss is undefined for an empty target.

\paragraph{Localizer inference.}
At test time, we threshold the region map at $0.5$, smooth it by a majority vote over 5 frames, and remove components smaller than $0.3\%$ of the frame.
A frame left empty inside the span of the region takes the mask of the nearest non-empty frame, and the region is dilated by 12 pixels.
The localizer computes $s_m$ at stride 4, and $s_m$ is bilinearly upsampled to the frame size before the logistic function.
The inpainter repairs a 49-frame window, which on BrokenVideos clips is centered on the annotated frames, and a video without a region is left unchanged.

\paragraph{Inpainter.}
VACE-14B runs on a 49-frame window at $832\times480$ with 25 steps of the UniPC sampler at a flow shift of 3 and classifier-free guidance of 5, with the standard negative prompt of Wan2.1.
The reference cut-out is taken from the frame in the middle half of the window where the region is largest, cropped to the region with a 16-pixel margin.
The decoded output of VACE-14B differs slightly from the original in color and fine detail, even outside the region, so we correct the edge and the color of the region before the composite.
The edge correction measures the difference to the original on a 3-pixel ring inside the region border, spreads it inward with a Gaussian of 12 pixels, and adds it with a weight $\exp(-d/8)$ at a distance of $d$ pixels inside the ring.
The color of the region is then matched to the original region by matching the mean and standard deviation in Lab space over all frames, with strength $0.7$.

\paragraph{Preference tuning.}
Preference tuning ranks repairs with Video Jev and trains a LoRA of the inpainter on pairs of them.
Video Jev sees each sample at about 2 frames per second of the source video, at least 8 frames, resized to 448 pixels on the longer side, with the original and its repairs read at the same frame indices.
It answers two yes-or-no questions, ``Is this video AI-generated?'' and ``Does this video contain visual artifacts, such as objects or body parts that morph, melt, or appear from nowhere?'', and we use the probabilities of ``no''.
The candidates are 1568 samples of 400 training clips, repaired on the same 49-frame windows as at test time and on the regions of an earlier version of our localizer, which was trained identically except that no real videos without artifacts were included.
A pair is kept only when the Video Jev scores of its two samples differ by at least $0.05$ and the CLIP similarity of the preferred sample to the input is at most $0.02$ below that of the other sample, so that the tuning does not learn to rewrite the region.
Of the 178 pairs, 170 are used for training and 8 are held out for validation.
The LoRA has rank 32 and $\alpha=32$ on all 480 linear layers of the transformer, and is trained for 200 steps with $\beta=2500$ and two pairs per step, AdamW at a learning rate of $2\times10^{-5}$, a 50-step warm-up, gradient clipping at 1 and a flow shift of 3.
The two samples of a pair are noised with the same noise $\epsilon$ and the same noise level $t$, and the denoising error is restricted to the region, which focuses the optimization on the edit.
A latent cell covers $8\times8$ pixels and four frames, and it belongs to $M_z$ if any of its pixels is in the region, while the reference frame is excluded.
Video Jev is used only to build the pairs, not at test time.
Sampling the candidates takes about 35 GPU hours and training about 2.9 hours on one H200, a one-time cost, after which the tuned inpainter needs a single repair per clip.

\paragraph{Evaluation data and measures.}
The example pool holds 520 generated clips of 49 frames, which are processed whole, and 60 BrokenVideos clips, which use the window rule of the test set.
Gemini Omni 1.1 Flash was called through the \texttt{gemini-omni-1.1-flash} API.
We use EditCtrl instead of the earlier VideoPainter~\cite{bian2025videopainter} because it is more recent and runs on the same VACE-14B model as our pipeline, and VideoPainter's repairs were visibly worse in our initial tests.

\section{Further results referenced in the main paper}

\begin{figure}[t]
\centering
\includegraphics[width=0.72\linewidth]{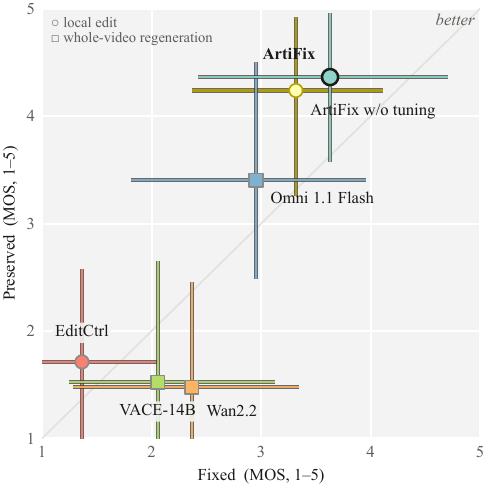}
\caption{\textbf{Only \method and the same pipeline without preference tuning are rated well on both questions} (SUREAL MOS, eleven observers, 95\% confidence intervals). Circles edit a region and squares regenerate the video.}
\label{fig:fixpreserve}
\end{figure}

\begin{figure}[t]
\centering
\includegraphics[width=\linewidth]{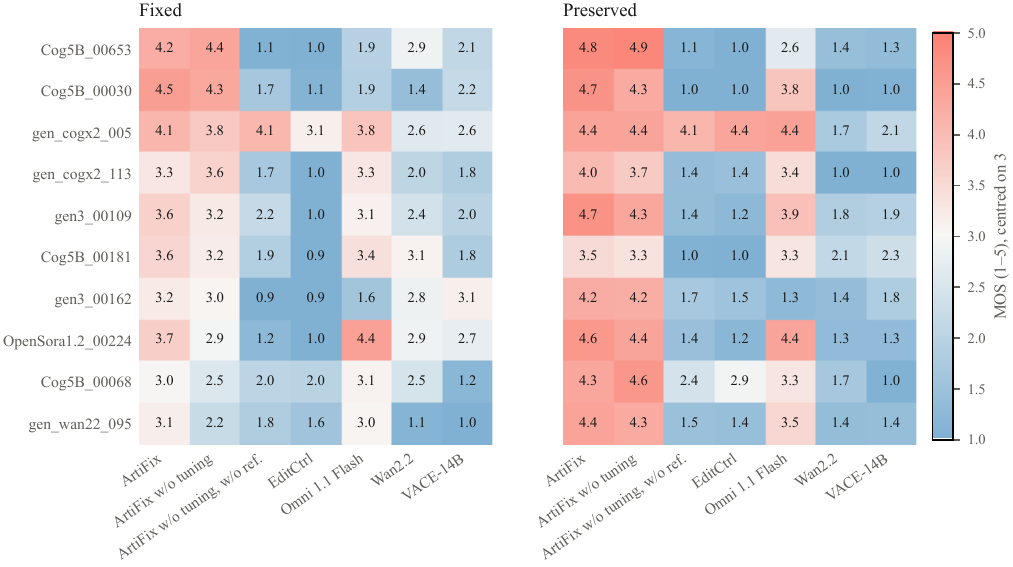}
\caption{\textbf{Ratings of every clip} (SUREAL MOS per clip and inpainting condition, fixing left and preservation right).}
\label{fig:supp_heatmaps}
\end{figure}

\paragraph{Ratings per method and clip.}
\Cref{fig:fixpreserve} plots each method of the perceptual study by its fixing and preservation ratings, and \cref{fig:supp_heatmaps} gives the ratings of every clip.

\paragraph{Region size.}
A region that is too small leaves part of the artifact for the inpainter to continue, and SAMWISE, whose regions cover 72\% of the annotated artifact, gives the worst repairs among the localizers with high coverage (judge $\Delta$ $+0.55$ against $+0.30$ for ours, \cref{tab:supp_localization}).
A region that is too large costs little, because the reference cut-out keeps the objects inside it and every pixel outside it is copied, as the false regions on videos without artifacts show (DINOv2 $0.925$).

\paragraph{Object-level regions.}
\Cref{fig:supp_tight} shows a study clip where the annotation marks only the knob, so the hand that turns it is not redrawn and the repaired knob no longer matches it, while our region covers the hand and the door, and observers rated this repair higher on fixing and preservation ($2.47$ and $4.32$ against $2.12$ and $3.42$).

\begin{figure}[t]
\centering
\input{figures/grid_tight}
\caption{\textbf{A repair needs the object and what it interacts with, not only the broken pixels} (study clip, both regions repaired by the same inpainter without preference tuning, as shown to observers). The dataset annotation marks only the knob, so it misses the hand that turns it, and the repaired knob no longer matches the hand and stays wrong once the hand stops covering it (frame 44). Our region covers the knob together with the hand and the door, so the whole interaction is redrawn consistently.}
\label{fig:supp_tight}
\end{figure}

\input{sec/X_prompts}

%% file: figures/supp_block_pairapp.tex
\begin{figure*}[t]
\centering
\newcommand{\pairshot}[1]{\IfFileExists{figures/assets/#1}{\includegraphics[width=0.32\textwidth]{figures/assets/#1}}{\fbox{\parbox[c][0.18\textwidth][c]{0.3\textwidth}{\centering\scriptsize #1}}}}
\pairshot{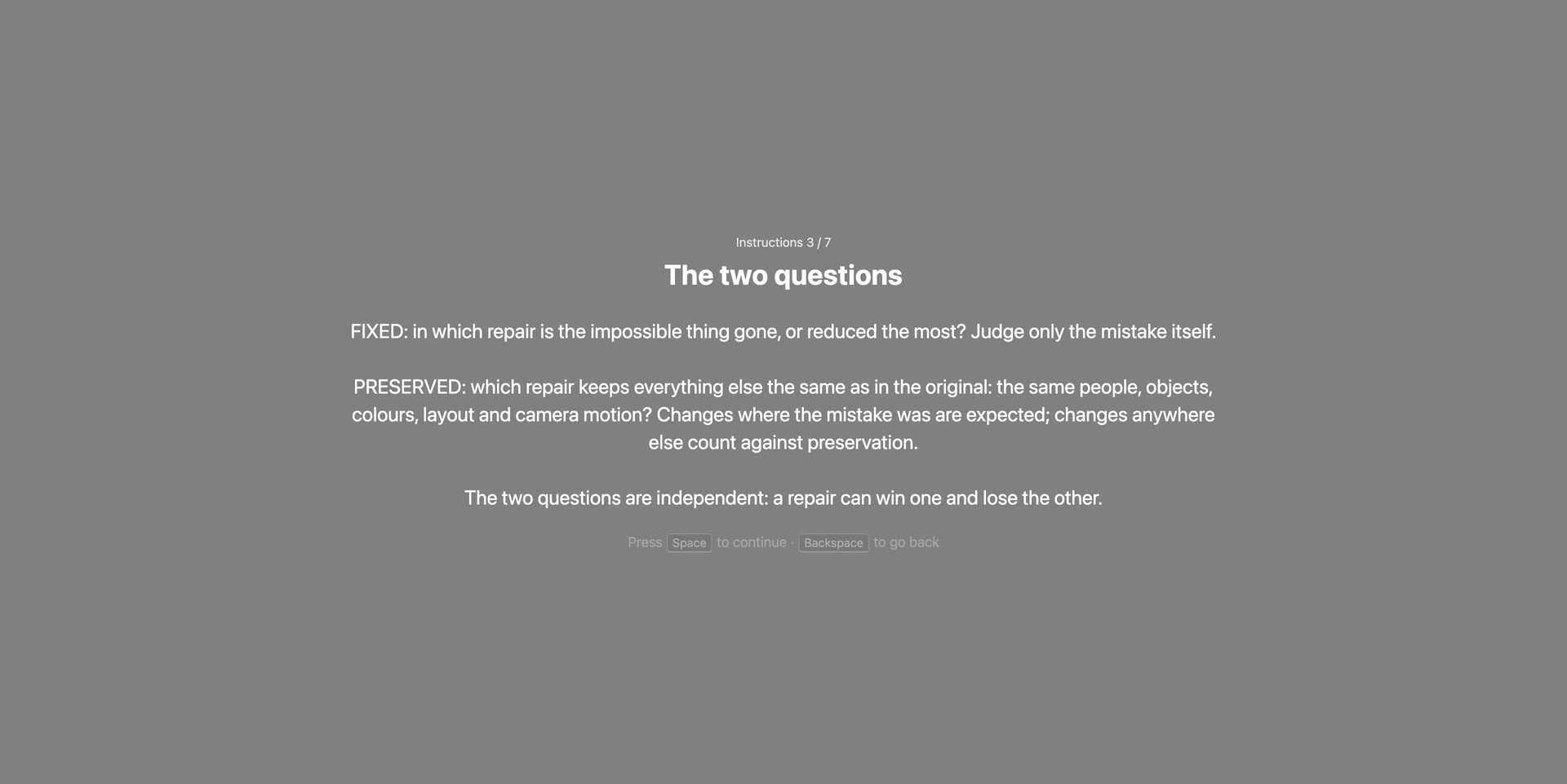}\hfill
\pairshot{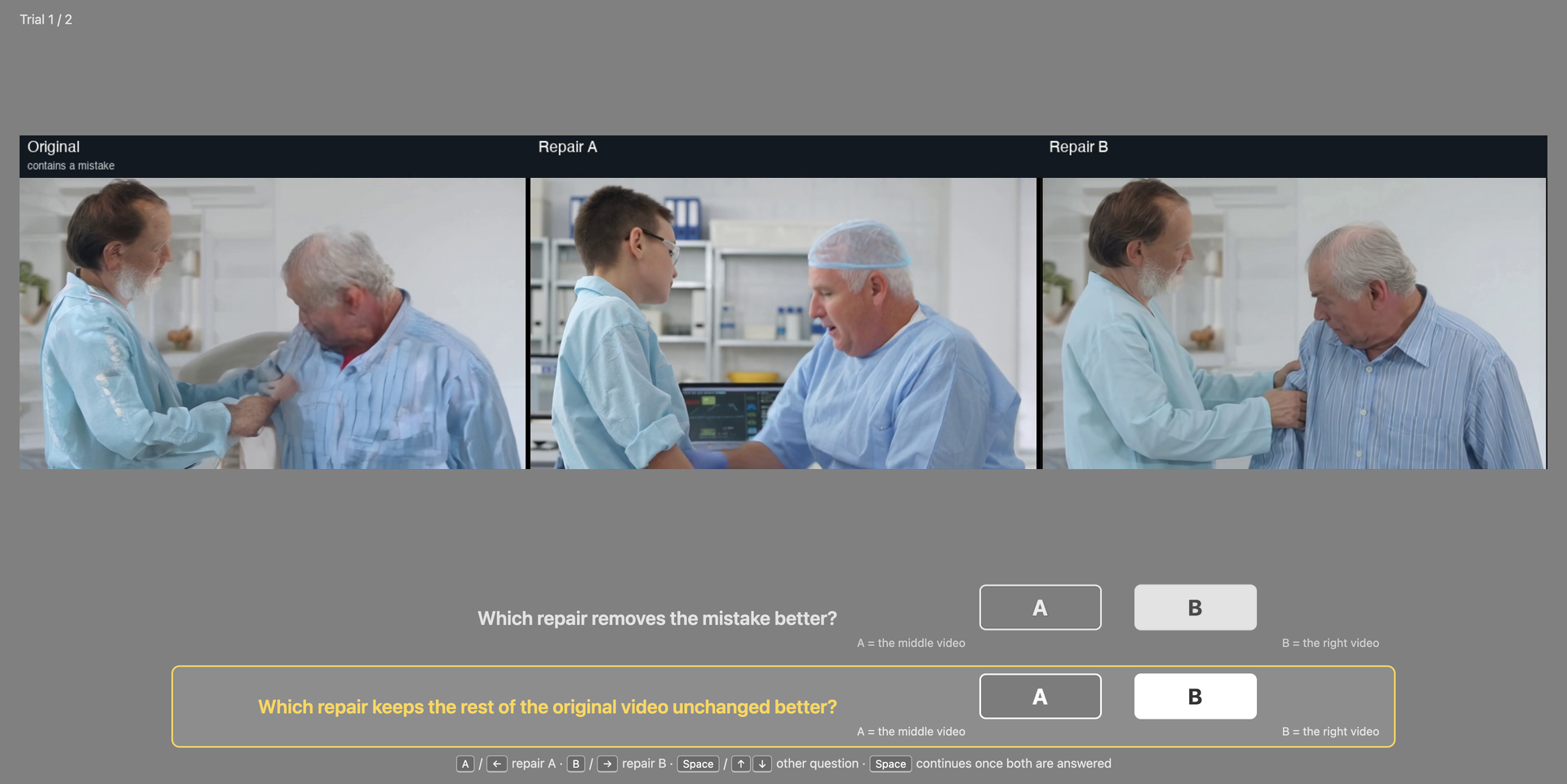}\hfill
\pairshot{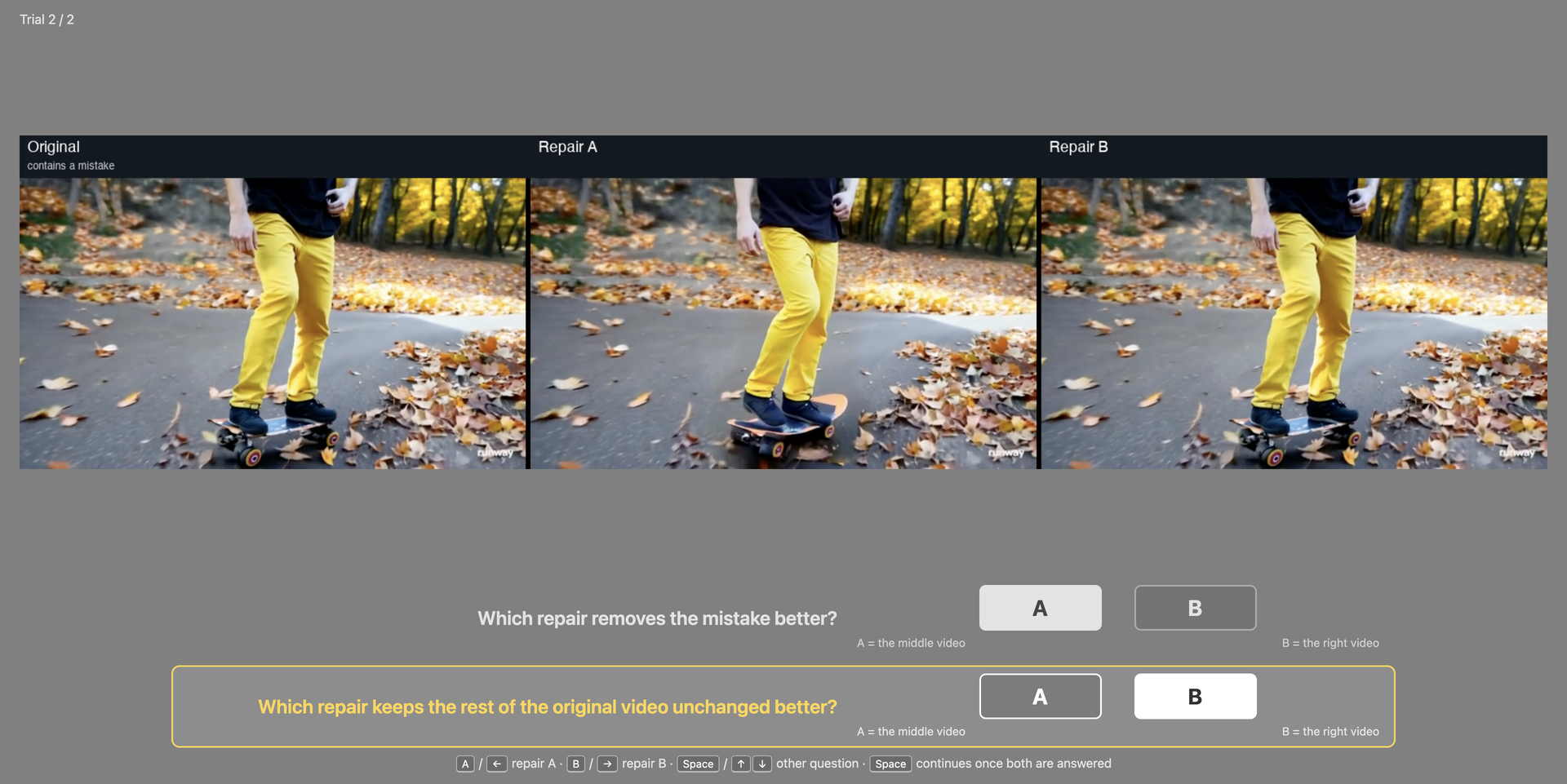}
\caption{\textbf{The blind follow-up shows no region overlay.} The blind pairwise follow-up in the rating application. The instruction page that defines the two questions, a trial with the original and two repairs labeled only A and B, and a trial in which one of the two options is the unrepaired original. Observers answer each question with the left and right keys.}
\label{fig:supp_pairapp}
\end{figure*}

\Cref{fig:supp_pairapp} shows the follow-up in the same application, without any region overlay.

%% file: tables/t1_main.tex
\begin{table*}[t]
\centering\footnotesize
\setlength{\tabcolsep}{3pt}
\resizebox{\linewidth}{!}{%
\begin{tabular}{llrrccccccccc}
\toprule
 & & & & & & & & & \multicolumn{2}{c}{MOS (study)} & \multicolumn{2}{c}{JOD (follow-up)} \\
\cmidrule(lr){10-11}\cmidrule(lr){12-13}
Method & Edits & Params & s/clip & LPIPS$\downarrow$ & CLIP$\uparrow$ & DINOv2$\uparrow$ & FID$\downarrow$ & Judge $\Delta\downarrow$ & Fixed$\uparrow$ & Preserved$\uparrow$ & Fixed$\uparrow$ & Preserved$\uparrow$ \\
\midrule
ProPainter~\cite{zhou2023propainter} & inpaint & 50M & 36 & 0.252 & 0.798 & 0.581 & 141 & +2.33 & -- & -- & -- & -- \\
SDXL inpaint (per frame)~\cite{podell2024sdxl} & inpaint & 2.6B & 33 & 0.207 & 0.840 & 0.710 & 73 & +2.44 & -- & -- & -- & -- \\
EditCtrl~\cite{litman2026editctrl} & inpaint & 17B + 3.3B & 52 & 0.210 & 0.845 & 0.741 & 85 & +1.93 & 1.36$^\dagger$ & 1.71$^\dagger$ & -- & -- \\
VACE-1.3B~\cite{jiang2025vace} & regen & 2.2B & \textbf{27} & 0.381 & 0.814 & 0.641 & 95 & -0.75 & -- & -- & -- & -- \\
VACE-14B~\cite{jiang2025vace} & regen & 17B & 99 & 0.371 & 0.832 & 0.711 & 88 & -0.72 & 2.06$^\dagger$ & 1.52$^\dagger$ & -- & -- \\
Wan2.2-A14B~\cite{wan2025wan} & regen & 27B (14B act.) & 81 & 0.391 & 0.826 & 0.694 & 92 & \textbf{-0.87} & 2.37$^\dagger$ & 1.48$^\dagger$ & $-$0.07 & $-$2.34 \\
Gemini Omni 1.1 Flash (API) & regen & -- & API & 0.158 & 0.937 & \textbf{0.906} & 80$^*$ & -0.59 & 2.95 & 3.40$^\dagger$ & +0.15 & $-$0.33 \\
\midrule
\method w/o preference tuning & local & 17B + 45M & 77 & 0.122 & \textbf{0.938} & \textbf{0.906} & 38 & +0.30 & 3.32$^\dagger$ & 4.24 & -- & -- \\
\textbf{\method} & local & 17B + 229M & 83 & \textbf{0.119} & \textbf{0.938} & \textbf{0.906} & \textbf{37} & +0.13 & \textbf{3.63} & \textbf{4.36} & \textbf{+0.85} & \textbf{0} \\
\bottomrule
\end{tabular}}
\caption{\textbf{\method changes the video least of all repair methods, and observers prefer it in both studies.} Main comparison on the 103 held-out BrokenVideos test clips (49-frame windows). LPIPS / CLIP / DINOv2 / FID are measured against the original window and measure preservation. The judge $\Delta$ is the change in artifact severity rated by Qwen2.5-VL-72B, where negative means cleaner, and serves only as a screen. \emph{Edits} gives the kind of method. Inpaint fills a given region, regen regenerates the whole video at strength 0.85, and local is ours, which detects and then redraws the region. $^*$Omni on the 63 clips it accepted (FID over fewer frames is not comparable). s/clip on one H200. \method uses the preference-tuned inpainter (Sec.~3.4 of the main paper), and the row above it is the same pipeline without the tuning. Params counts the denoiser as loaded, without text encoders (VACE-14B and -1.3B have 17.3B and 2.2B parameters with their VACE blocks), plus EditCtrl's added modules and LoRA (3.3B) or our localizer (45M) and LoRA (184M). MOS (1--5) comes from the perceptual study (Sec.~4.3 of the main paper), where $^\dagger$ differs from \method (Wilcoxon over clips, Holm-corrected, $p<.05$). JOD comes from the blind pairwise follow-up on 30 other clips (\cref{sec:supp_pairwise}), with 0 at the unrepaired original for fixing and at \method for preservation.}
\label{tab:main}
\end{table*}

%% file: figures/supp_block_examples.tex
\Cref{fig:supp_examples,fig:supp_examples_inpaint} show one frame of each repair example of the supplementary videos for every method of \cref{tab:main}. \method changes only the localized region, the regeneration methods redraw the whole scene, and the region-inpainting methods replace or erase the object inside our region.

\begin{figure*}[t]
\centering
\input{figures/grid_examples_regen}
\caption{\textbf{\method redraws only the localized region, while the regeneration methods redraw the whole video} (supplementary videos, one frame each). Example $n$ is supplementary video $n$. For example 5, which was not among the clips sent to Omni for the evaluation, Omni received the instruction ``Fix the artifacts.''}
\label{fig:supp_examples}
\end{figure*}

\begin{figure*}[t]
\centering
\input{figures/grid_examples_inpaint}
\caption{\textbf{On the same region, the region-inpainting methods do not keep the original object, while \method does} (the examples of \cref{fig:supp_examples}). EditCtrl, ProPainter and SDXL per frame redraw the region that our localizer predicts, without a reference cut-out.}
\label{fig:supp_examples_inpaint}
\end{figure*}

%% file: figures/grid_examples_regen.tex
\providecolor{artifixteal}{HTML}{3E9C8F}%
{\setlength{\tabcolsep}{0.6pt}\renewcommand{\arraystretch}{0.6}\setlength{\fboxsep}{0pt}%
\begin{tabular}{@{}rccccccc@{}}
 & \scriptsize Original & \scriptsize Localized region & \scriptsize\textbf{\textcolor{artifixteal}{ArtiFix}} & \scriptsize Omni 1.1 Flash & \scriptsize VACE-14B & \scriptsize VACE-1.3B & \scriptsize Wan2.2 \\[1pt]
\scriptsize Example 1\hspace{2pt} & \setlength{\fboxrule}{0.3pt}\fcolorbox{gray!60}{white}{\includegraphics[width=\dimexpr0.1257\textwidth-2\fboxrule\relax]{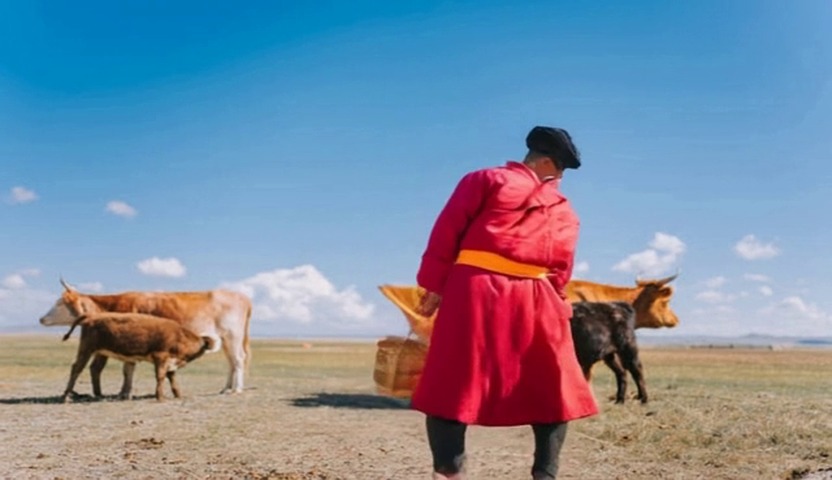}} & \setlength{\fboxrule}{0.3pt}\fcolorbox{gray!60}{white}{\includegraphics[width=\dimexpr0.1257\textwidth-2\fboxrule\relax]{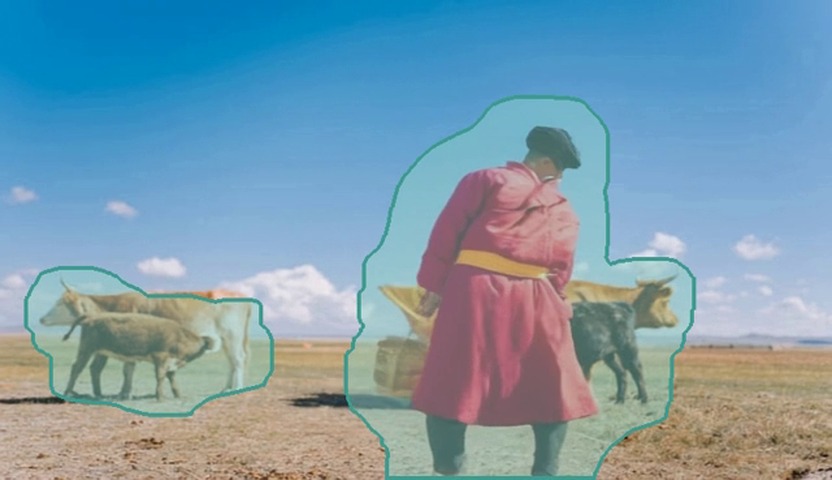}} & \setlength{\fboxrule}{0.8pt}\fcolorbox{artifixteal}{white}{\includegraphics[width=\dimexpr0.1257\textwidth-2\fboxrule\relax]{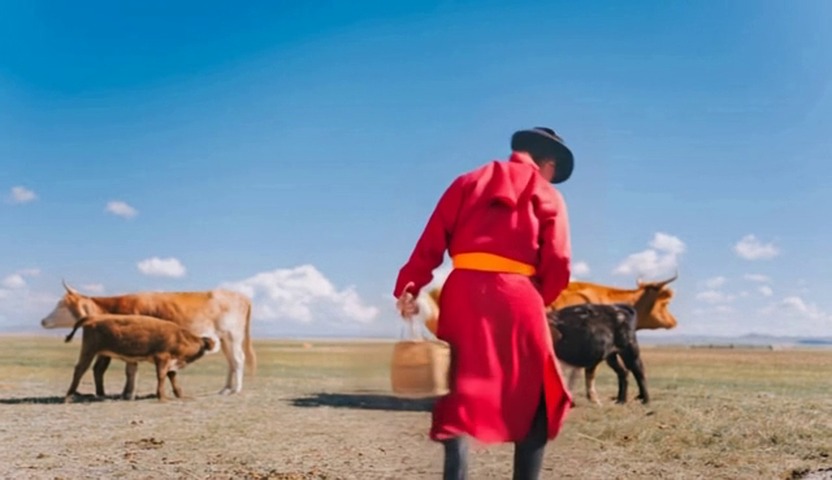}} & \setlength{\fboxrule}{0.3pt}\fcolorbox{gray!60}{white}{\includegraphics[width=\dimexpr0.1257\textwidth-2\fboxrule\relax]{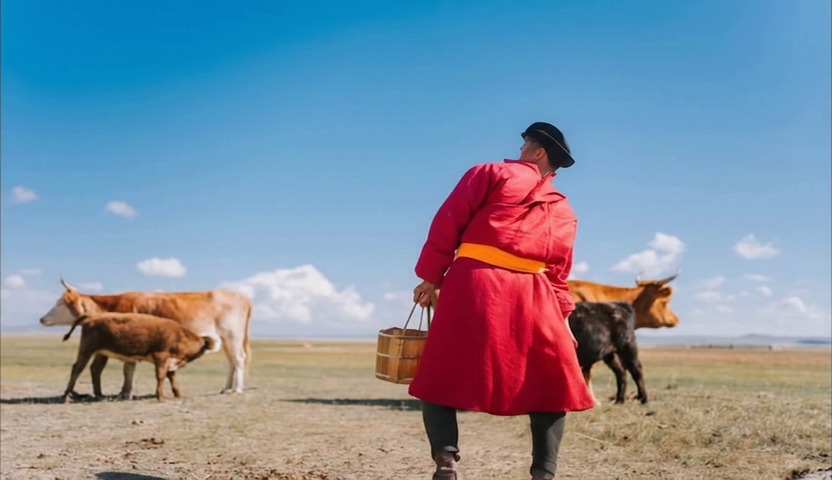}} & \setlength{\fboxrule}{0.3pt}\fcolorbox{gray!60}{white}{\includegraphics[width=\dimexpr0.1257\textwidth-2\fboxrule\relax]{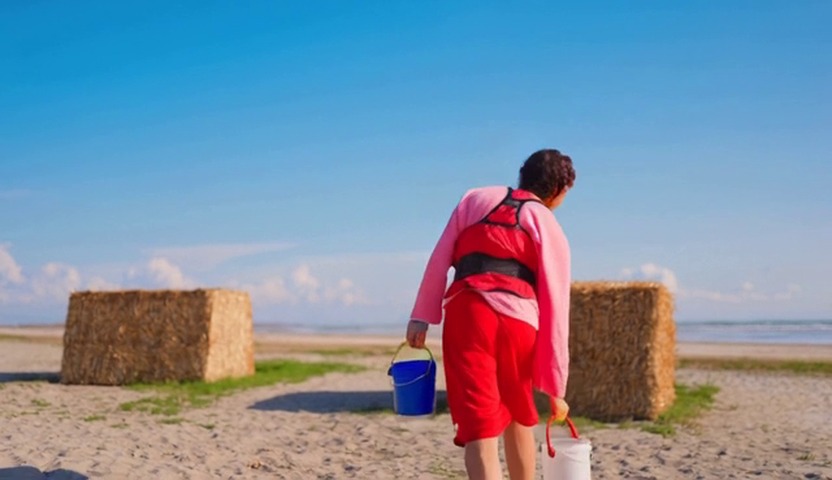}} & \setlength{\fboxrule}{0.3pt}\fcolorbox{gray!60}{white}{\includegraphics[width=\dimexpr0.1257\textwidth-2\fboxrule\relax]{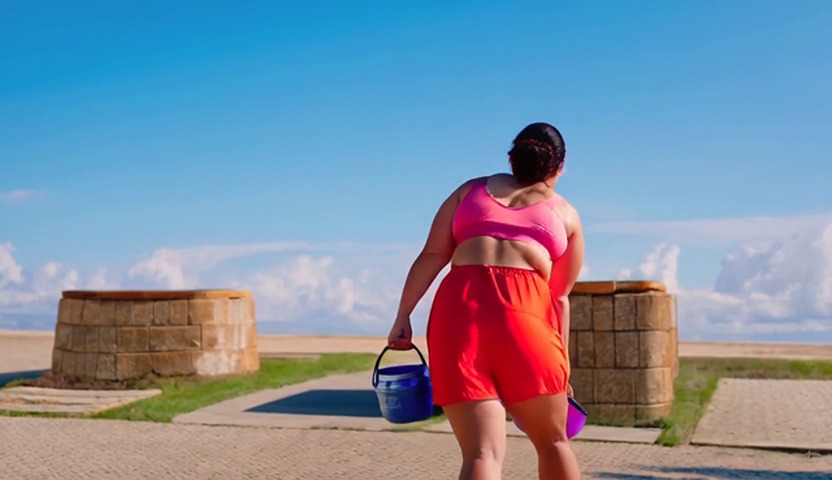}} & \setlength{\fboxrule}{0.3pt}\fcolorbox{gray!60}{white}{\includegraphics[width=\dimexpr0.1257\textwidth-2\fboxrule\relax]{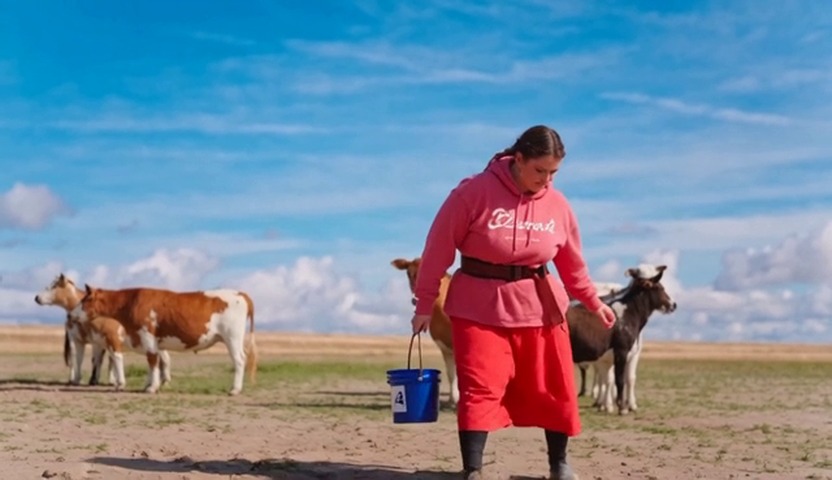}} \\
\scriptsize Example 2\hspace{2pt} & \setlength{\fboxrule}{0.3pt}\fcolorbox{gray!60}{white}{\includegraphics[width=\dimexpr0.1257\textwidth-2\fboxrule\relax]{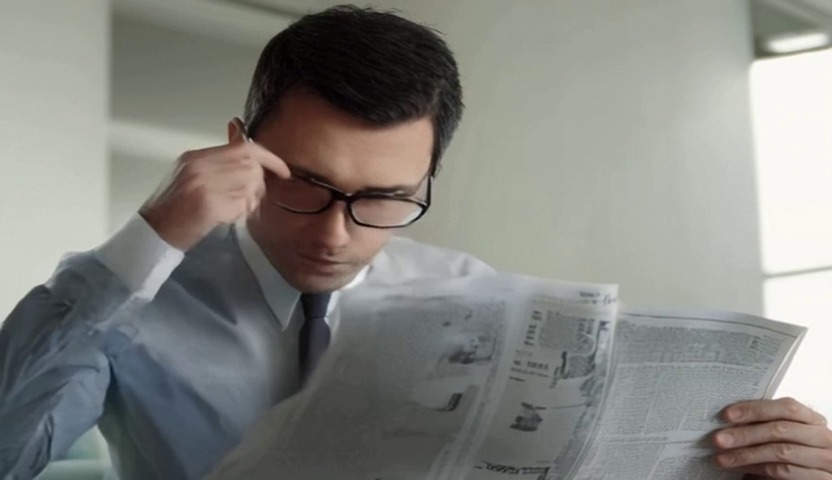}} & \setlength{\fboxrule}{0.3pt}\fcolorbox{gray!60}{white}{\includegraphics[width=\dimexpr0.1257\textwidth-2\fboxrule\relax]{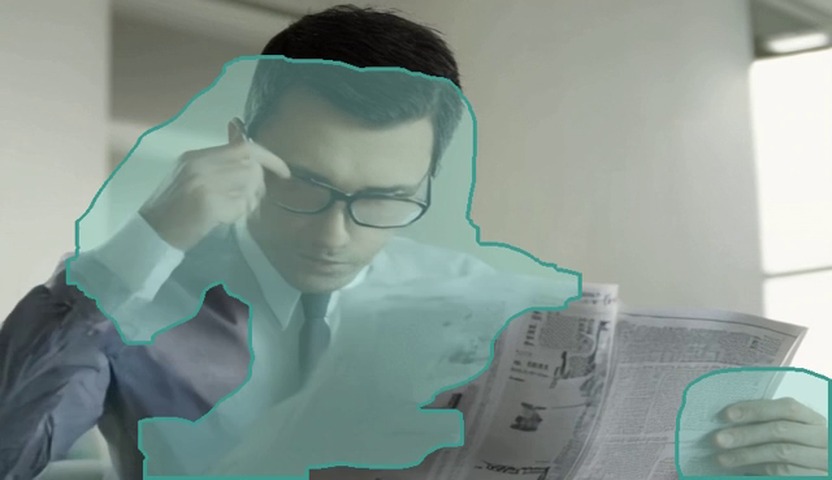}} & \setlength{\fboxrule}{0.8pt}\fcolorbox{artifixteal}{white}{\includegraphics[width=\dimexpr0.1257\textwidth-2\fboxrule\relax]{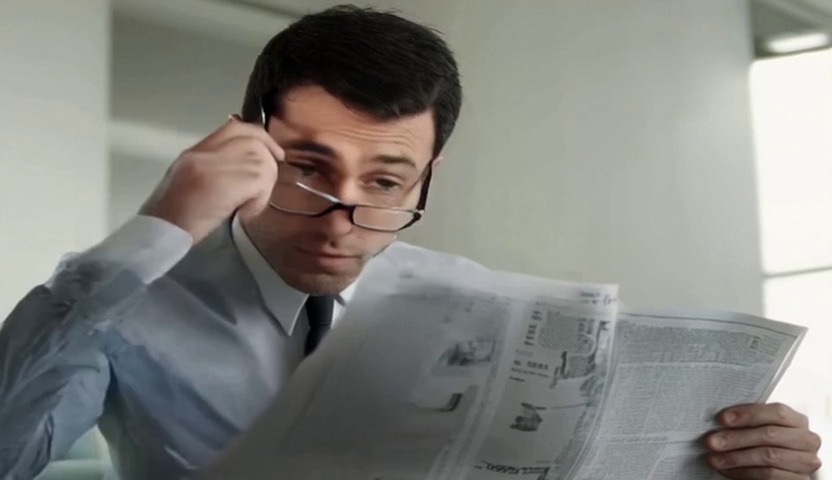}} & \setlength{\fboxrule}{0.3pt}\fcolorbox{gray!60}{white}{\includegraphics[width=\dimexpr0.1257\textwidth-2\fboxrule\relax]{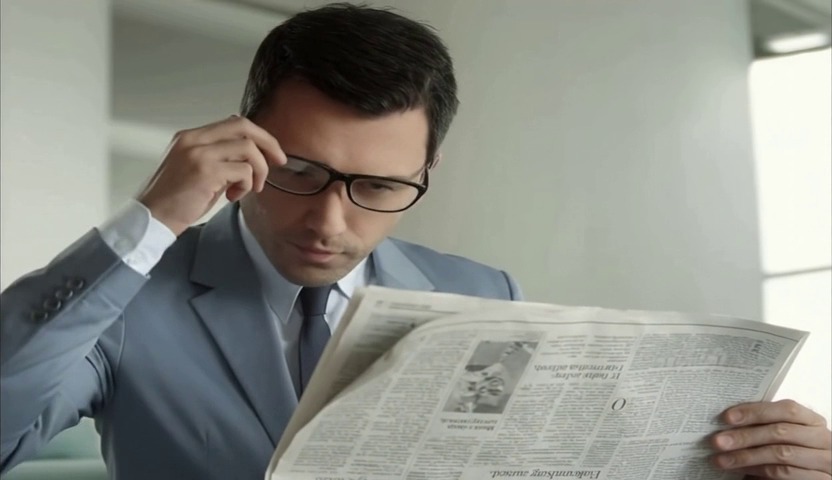}} & \setlength{\fboxrule}{0.3pt}\fcolorbox{gray!60}{white}{\includegraphics[width=\dimexpr0.1257\textwidth-2\fboxrule\relax]{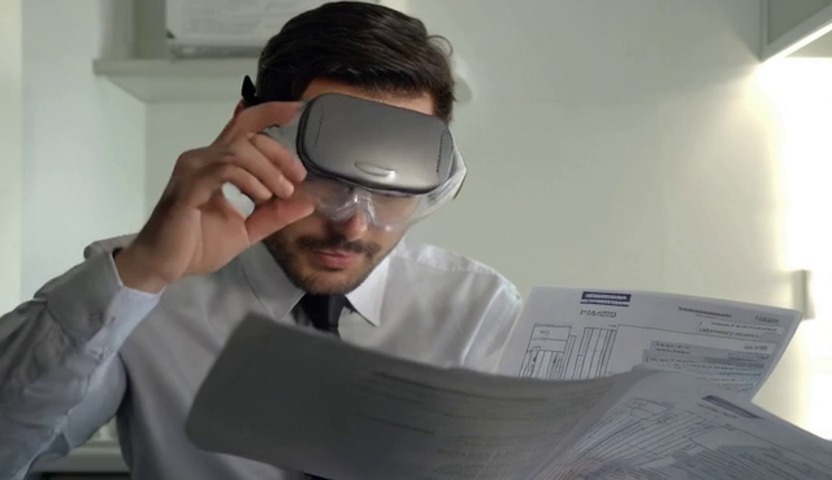}} & \setlength{\fboxrule}{0.3pt}\fcolorbox{gray!60}{white}{\includegraphics[width=\dimexpr0.1257\textwidth-2\fboxrule\relax]{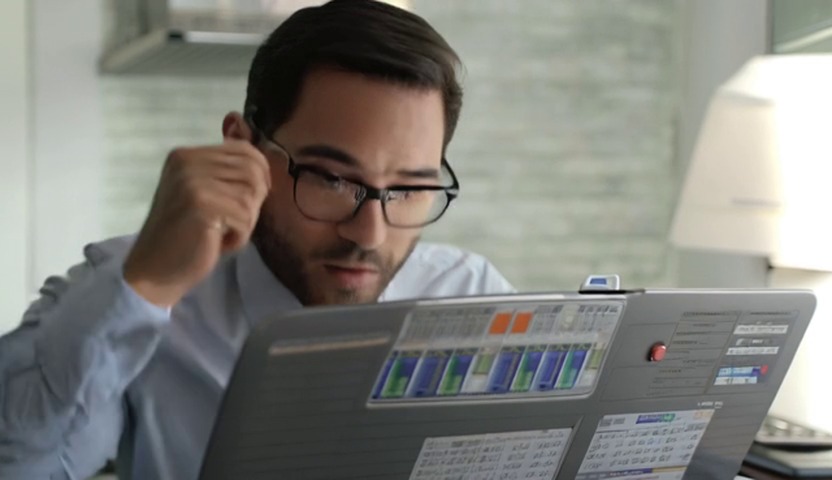}} & \setlength{\fboxrule}{0.3pt}\fcolorbox{gray!60}{white}{\includegraphics[width=\dimexpr0.1257\textwidth-2\fboxrule\relax]{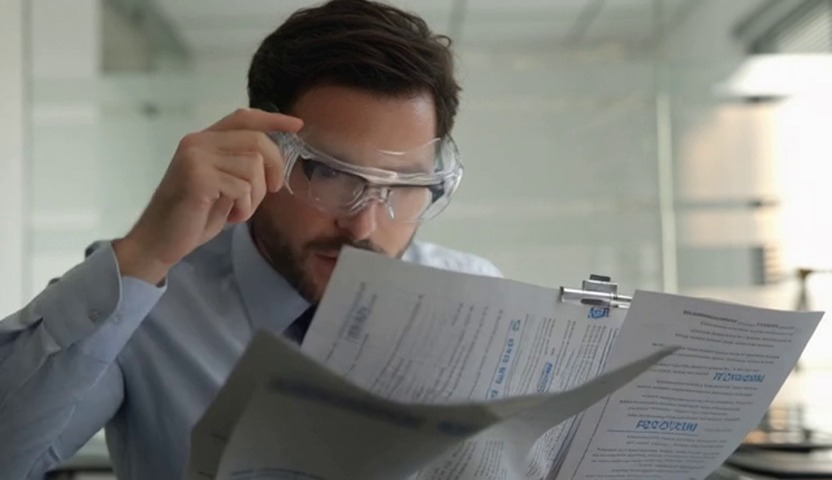}} \\
\scriptsize Example 3\hspace{2pt} & \setlength{\fboxrule}{0.3pt}\fcolorbox{gray!60}{white}{\includegraphics[width=\dimexpr0.1257\textwidth-2\fboxrule\relax]{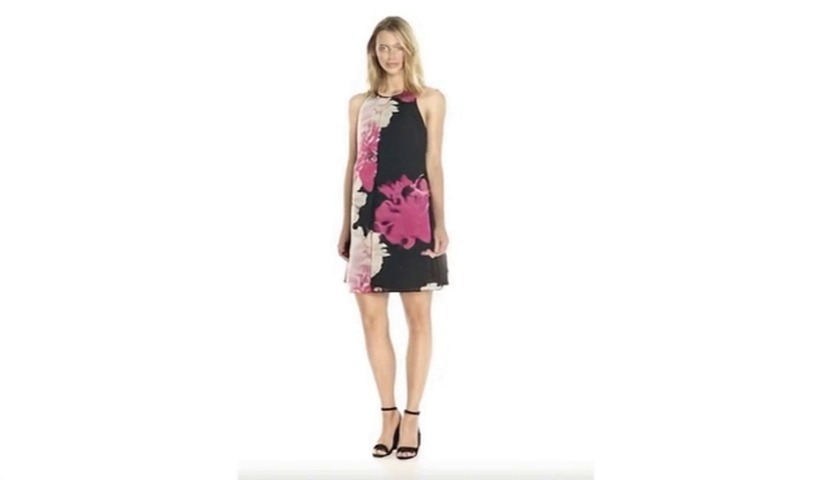}} & \setlength{\fboxrule}{0.3pt}\fcolorbox{gray!60}{white}{\includegraphics[width=\dimexpr0.1257\textwidth-2\fboxrule\relax]{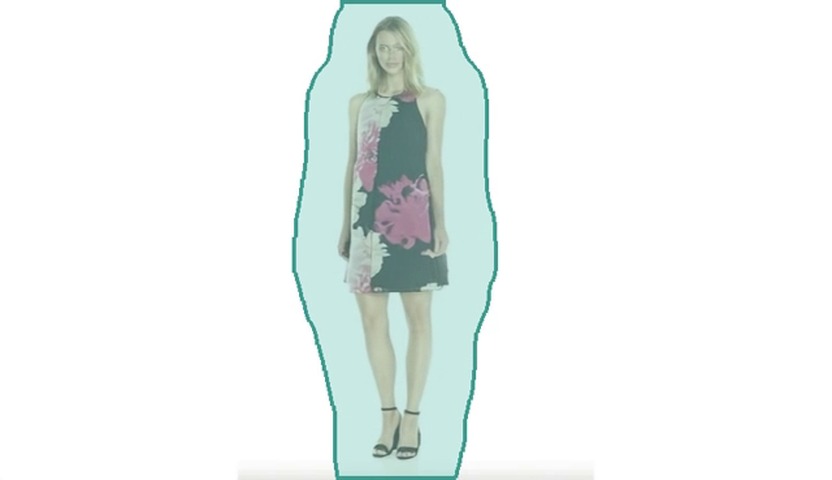}} & \setlength{\fboxrule}{0.8pt}\fcolorbox{artifixteal}{white}{\includegraphics[width=\dimexpr0.1257\textwidth-2\fboxrule\relax]{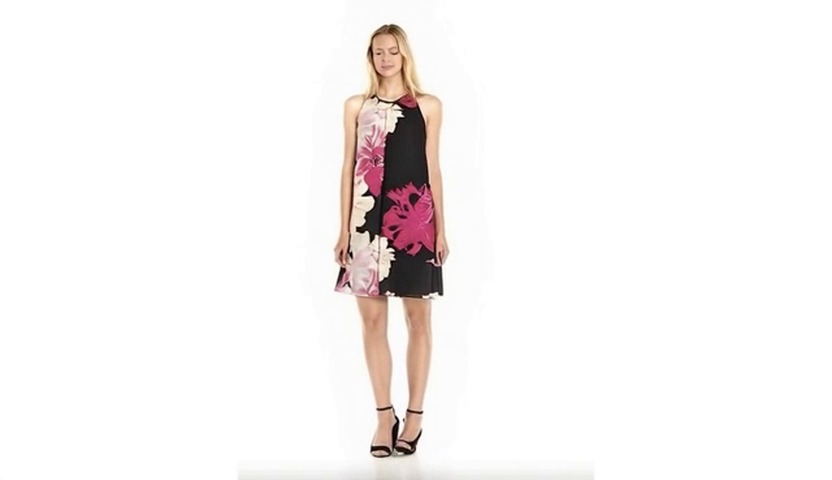}} & \setlength{\fboxrule}{0.3pt}\fcolorbox{gray!60}{white}{\includegraphics[width=\dimexpr0.1257\textwidth-2\fboxrule\relax]{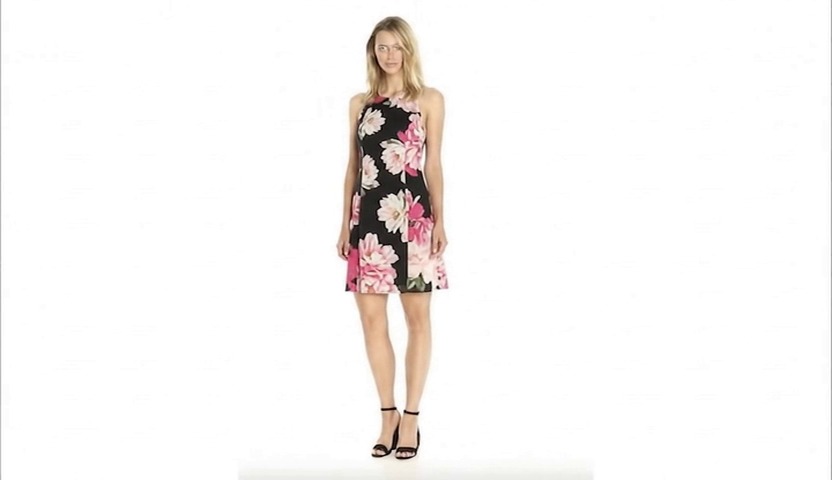}} & \setlength{\fboxrule}{0.3pt}\fcolorbox{gray!60}{white}{\includegraphics[width=\dimexpr0.1257\textwidth-2\fboxrule\relax]{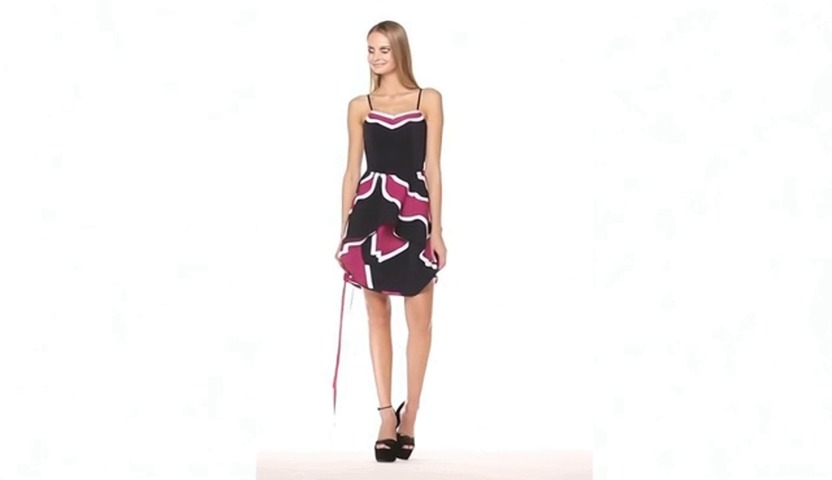}} & \setlength{\fboxrule}{0.3pt}\fcolorbox{gray!60}{white}{\includegraphics[width=\dimexpr0.1257\textwidth-2\fboxrule\relax]{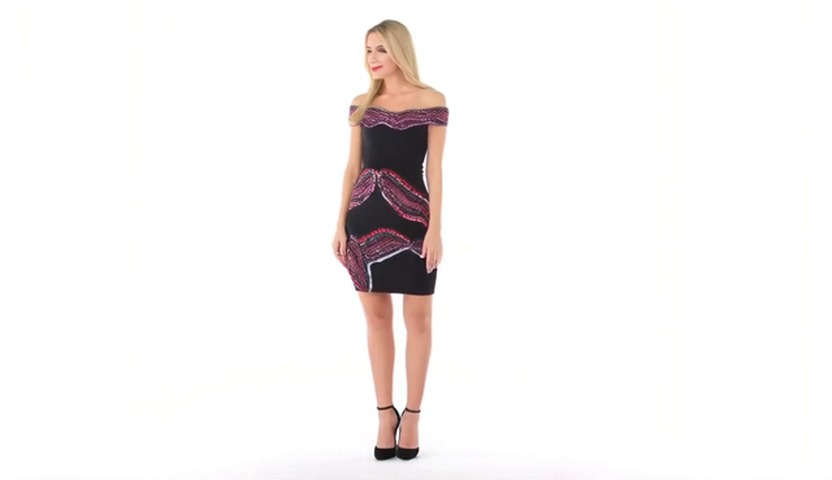}} & \setlength{\fboxrule}{0.3pt}\fcolorbox{gray!60}{white}{\includegraphics[width=\dimexpr0.1257\textwidth-2\fboxrule\relax]{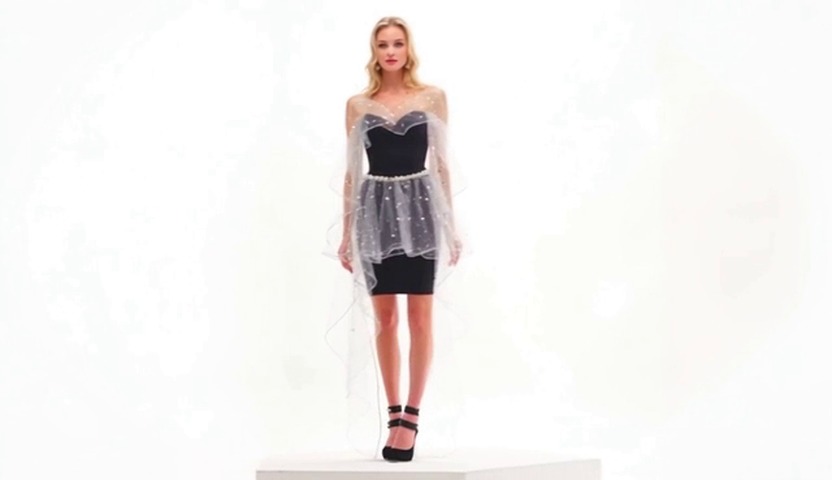}} \\
\scriptsize Example 4\hspace{2pt} & \setlength{\fboxrule}{0.3pt}\fcolorbox{gray!60}{white}{\includegraphics[width=\dimexpr0.1257\textwidth-2\fboxrule\relax]{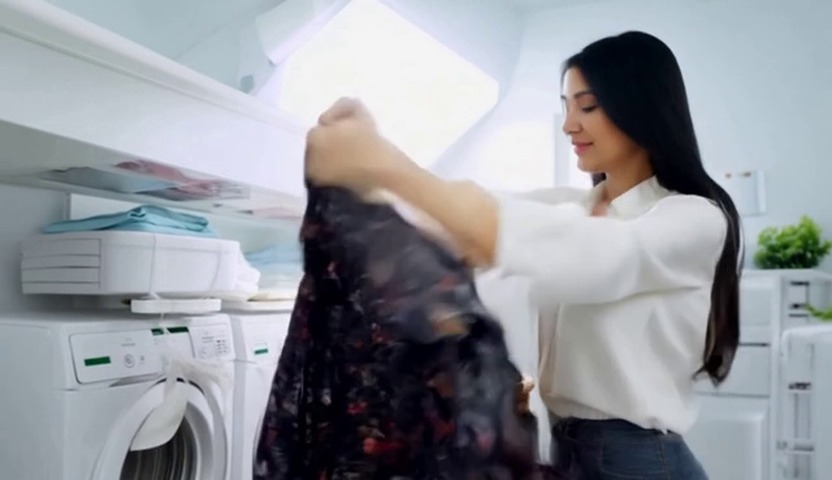}} & \setlength{\fboxrule}{0.3pt}\fcolorbox{gray!60}{white}{\includegraphics[width=\dimexpr0.1257\textwidth-2\fboxrule\relax]{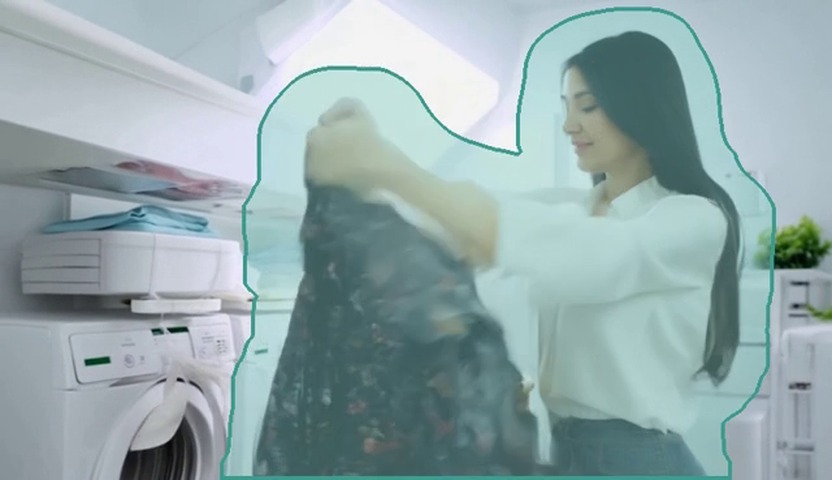}} & \setlength{\fboxrule}{0.8pt}\fcolorbox{artifixteal}{white}{\includegraphics[width=\dimexpr0.1257\textwidth-2\fboxrule\relax]{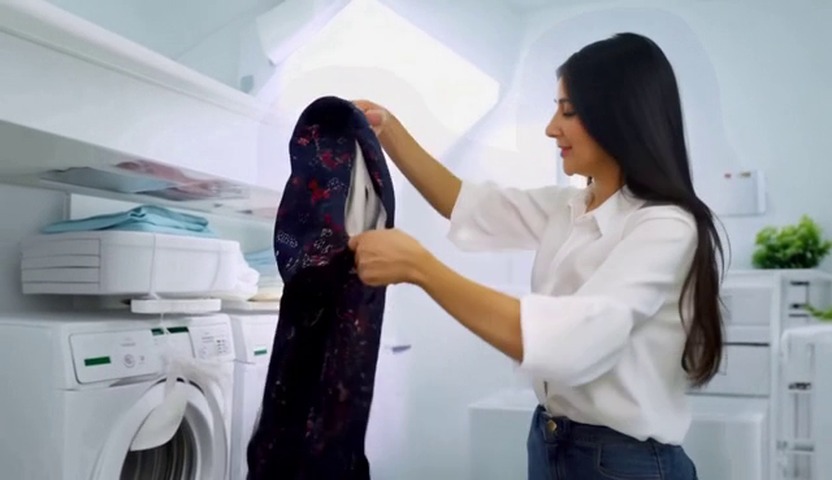}} & \setlength{\fboxrule}{0.3pt}\fcolorbox{gray!60}{white}{\includegraphics[width=\dimexpr0.1257\textwidth-2\fboxrule\relax]{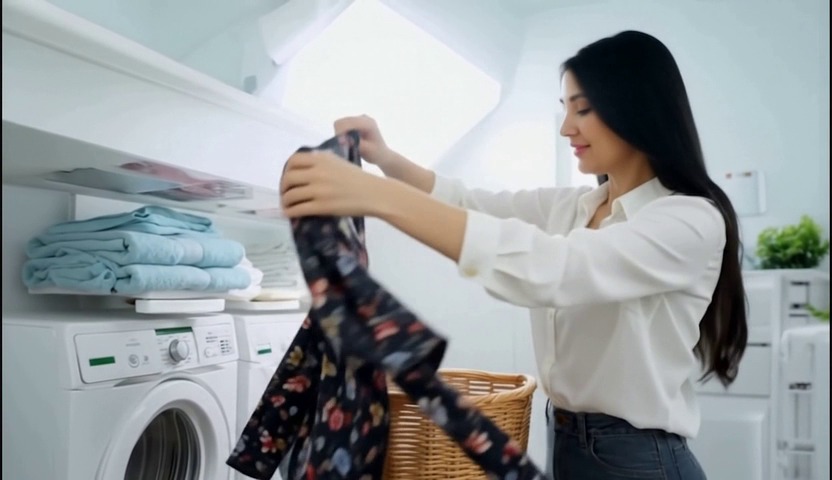}} & \setlength{\fboxrule}{0.3pt}\fcolorbox{gray!60}{white}{\includegraphics[width=\dimexpr0.1257\textwidth-2\fboxrule\relax]{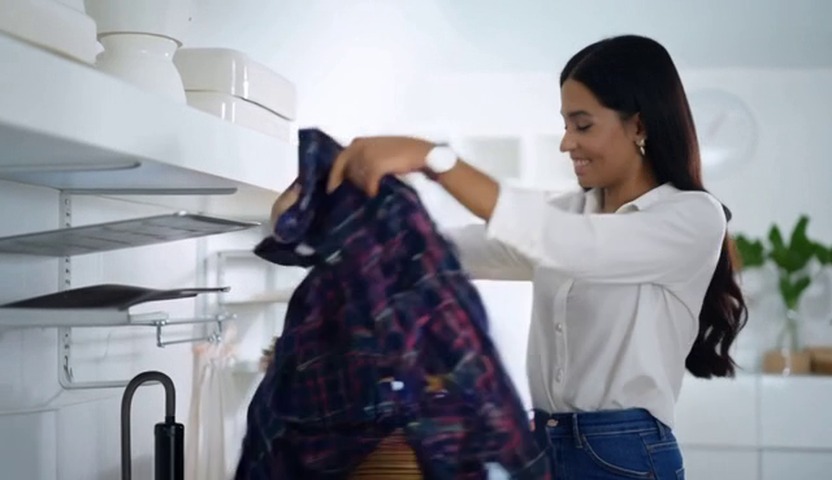}} & \setlength{\fboxrule}{0.3pt}\fcolorbox{gray!60}{white}{\includegraphics[width=\dimexpr0.1257\textwidth-2\fboxrule\relax]{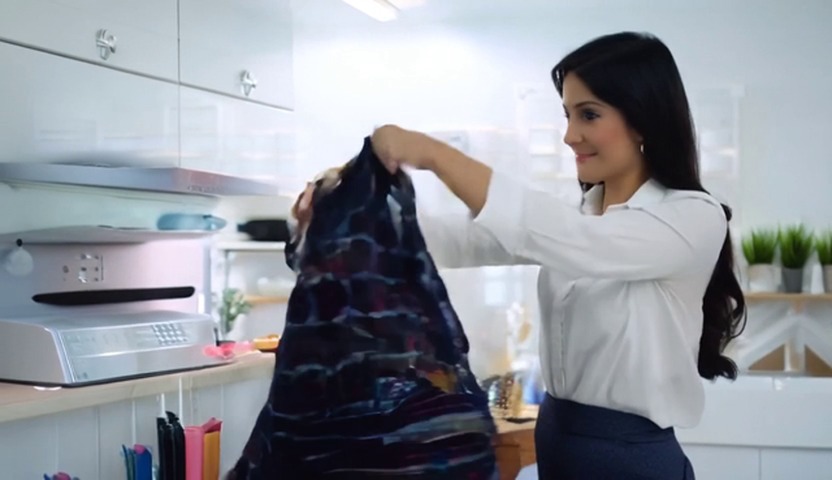}} & \setlength{\fboxrule}{0.3pt}\fcolorbox{gray!60}{white}{\includegraphics[width=\dimexpr0.1257\textwidth-2\fboxrule\relax]{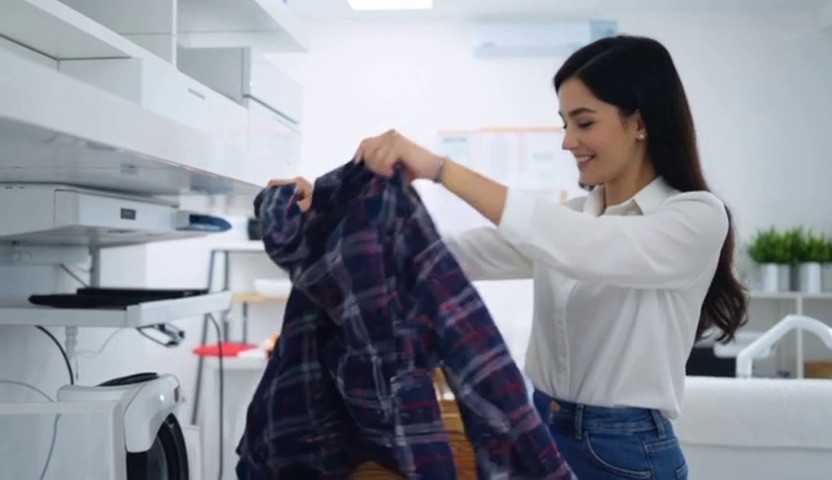}} \\
\scriptsize Example 5\hspace{2pt} & \setlength{\fboxrule}{0.3pt}\fcolorbox{gray!60}{white}{\includegraphics[width=\dimexpr0.1257\textwidth-2\fboxrule\relax]{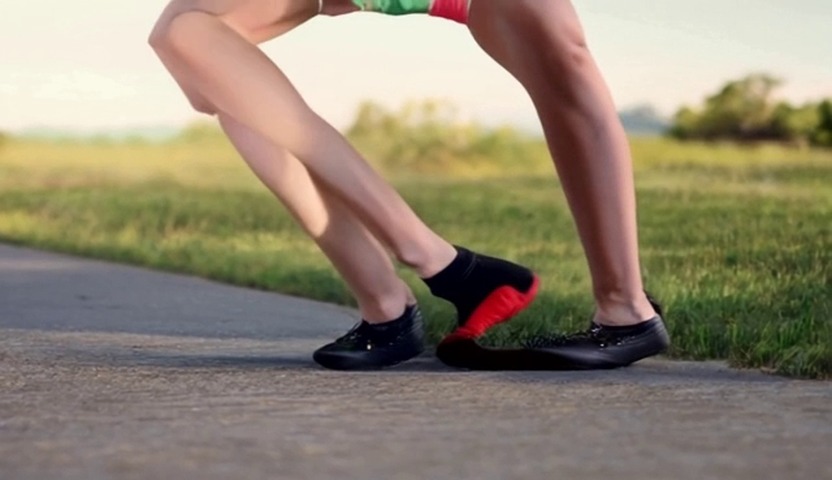}} & \setlength{\fboxrule}{0.3pt}\fcolorbox{gray!60}{white}{\includegraphics[width=\dimexpr0.1257\textwidth-2\fboxrule\relax]{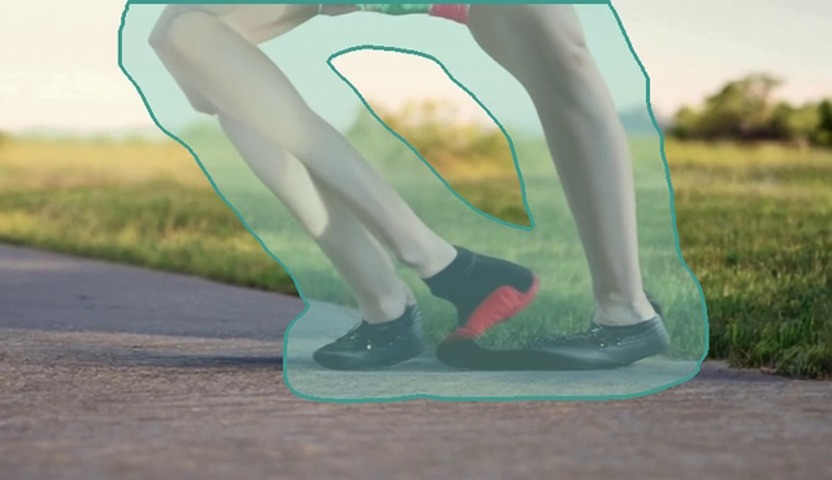}} & \setlength{\fboxrule}{0.8pt}\fcolorbox{artifixteal}{white}{\includegraphics[width=\dimexpr0.1257\textwidth-2\fboxrule\relax]{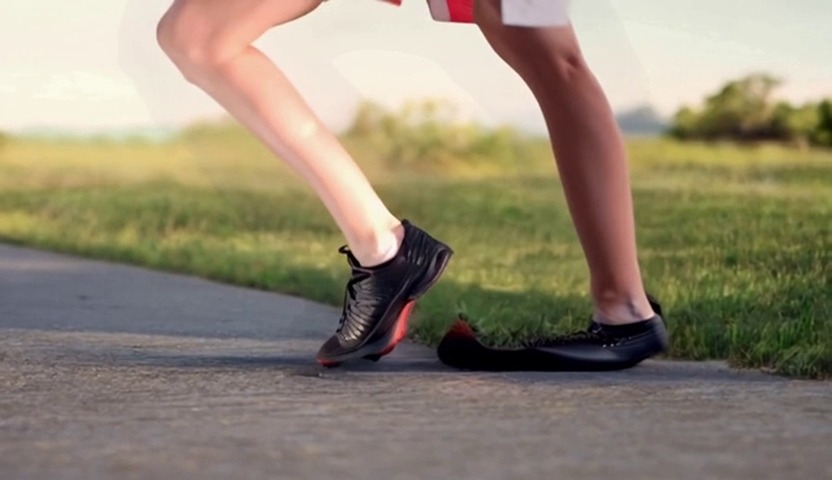}} & \setlength{\fboxrule}{0.3pt}\fcolorbox{gray!60}{white}{\includegraphics[width=\dimexpr0.1257\textwidth-2\fboxrule\relax]{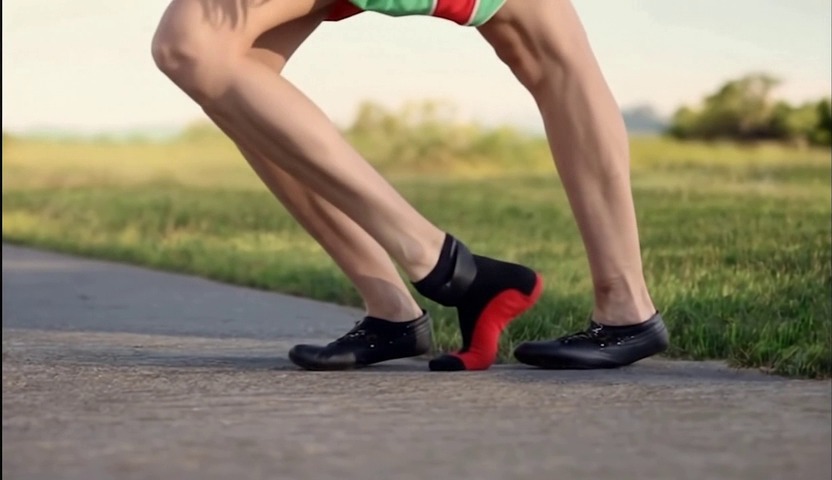}} & \setlength{\fboxrule}{0.3pt}\fcolorbox{gray!60}{white}{\includegraphics[width=\dimexpr0.1257\textwidth-2\fboxrule\relax]{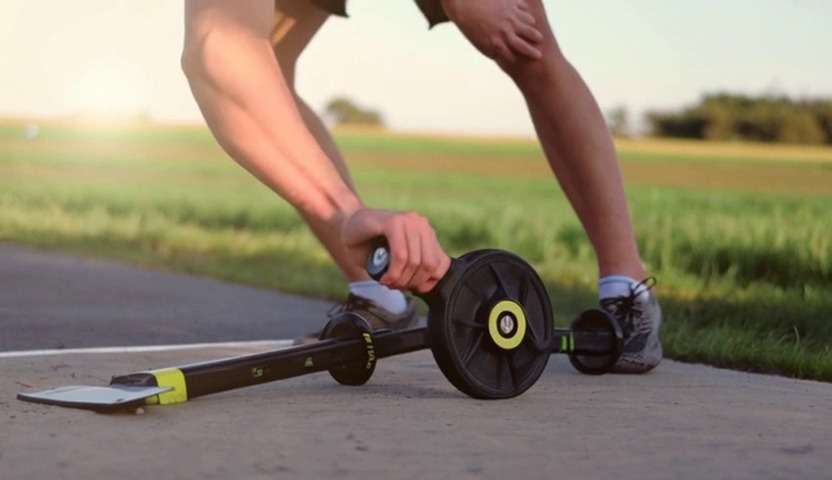}} & \setlength{\fboxrule}{0.3pt}\fcolorbox{gray!60}{white}{\includegraphics[width=\dimexpr0.1257\textwidth-2\fboxrule\relax]{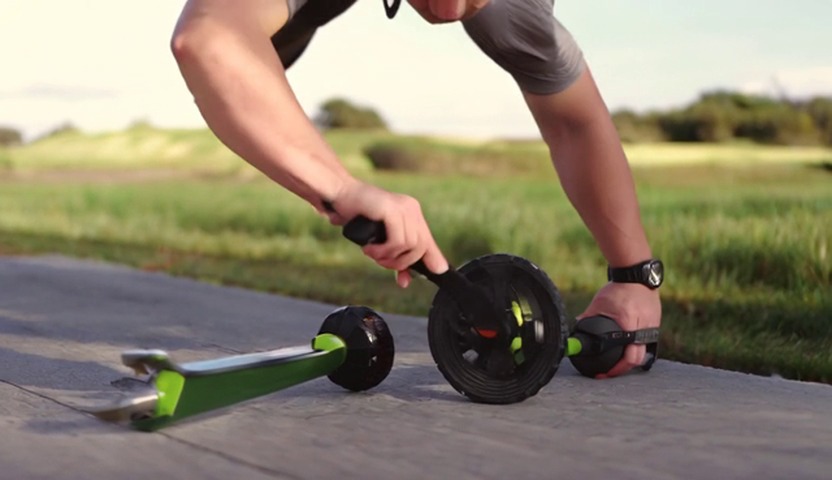}} & \setlength{\fboxrule}{0.3pt}\fcolorbox{gray!60}{white}{\includegraphics[width=\dimexpr0.1257\textwidth-2\fboxrule\relax]{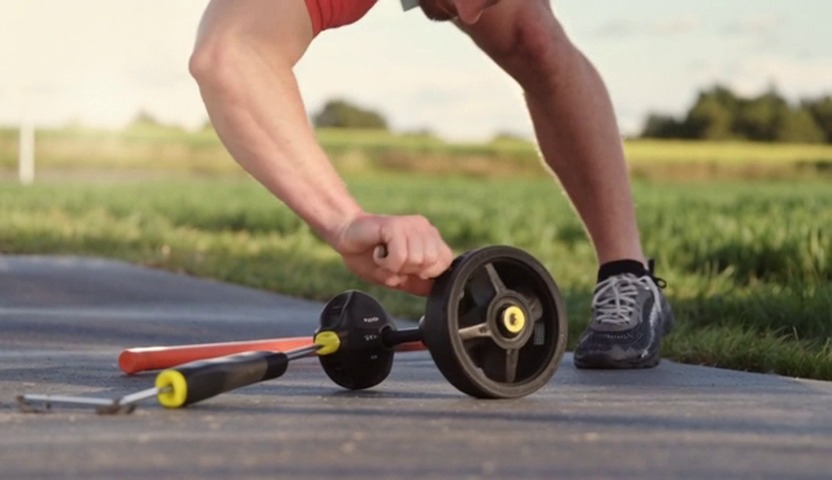}} \\
\end{tabular}}

%% file: figures/grid_examples_inpaint.tex
\providecolor{artifixteal}{HTML}{3E9C8F}%
{\setlength{\tabcolsep}{0.6pt}\renewcommand{\arraystretch}{0.6}\setlength{\fboxsep}{0pt}%
\begin{tabular}{@{}rcccccc@{}}
 & \scriptsize Original & \scriptsize Localized region & \scriptsize\textbf{\textcolor{artifixteal}{ArtiFix}} & \scriptsize EditCtrl & \scriptsize ProPainter & \scriptsize SDXL per frame \\[1pt]
\scriptsize Example 1\hspace{2pt} & \setlength{\fboxrule}{0.3pt}\fcolorbox{gray!60}{white}{\includegraphics[width=\dimexpr0.1475\textwidth-2\fboxrule\relax]{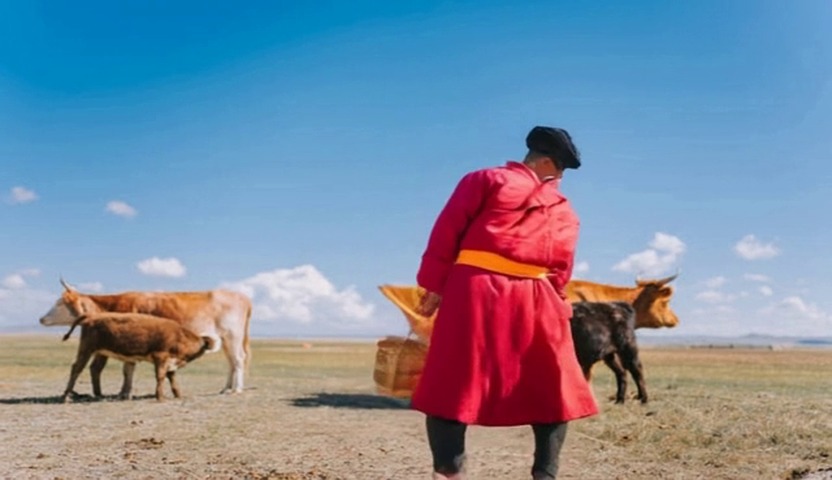}} & \setlength{\fboxrule}{0.3pt}\fcolorbox{gray!60}{white}{\includegraphics[width=\dimexpr0.1475\textwidth-2\fboxrule\relax]{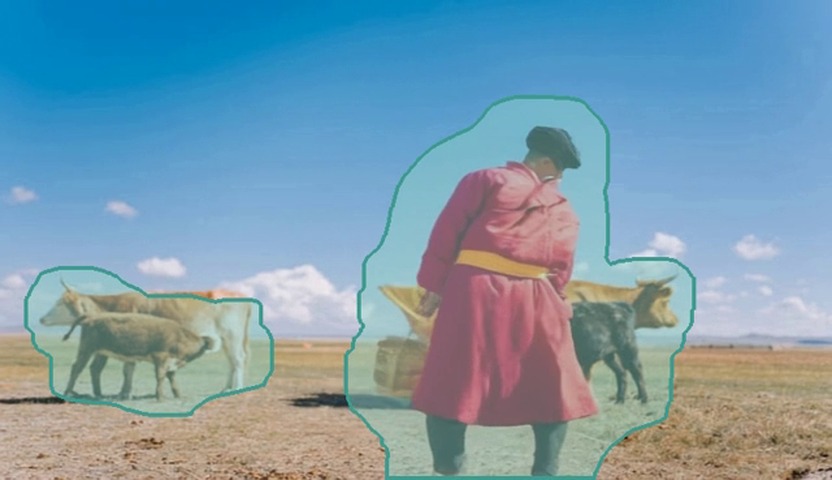}} & \setlength{\fboxrule}{0.8pt}\fcolorbox{artifixteal}{white}{\includegraphics[width=\dimexpr0.1475\textwidth-2\fboxrule\relax]{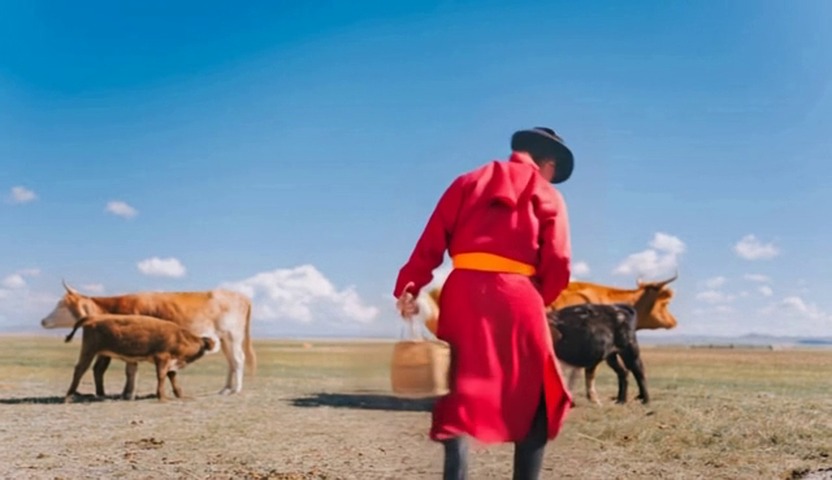}} & \setlength{\fboxrule}{0.3pt}\fcolorbox{gray!60}{white}{\includegraphics[width=\dimexpr0.1475\textwidth-2\fboxrule\relax]{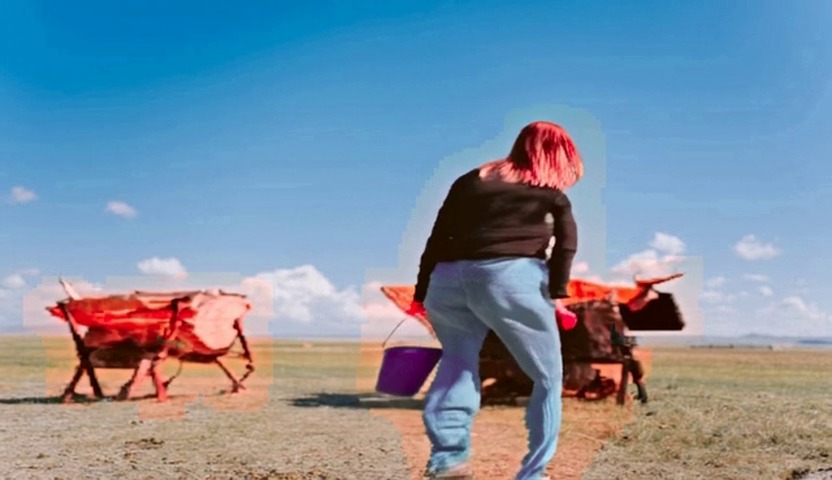}} & \setlength{\fboxrule}{0.3pt}\fcolorbox{gray!60}{white}{\includegraphics[width=\dimexpr0.1475\textwidth-2\fboxrule\relax]{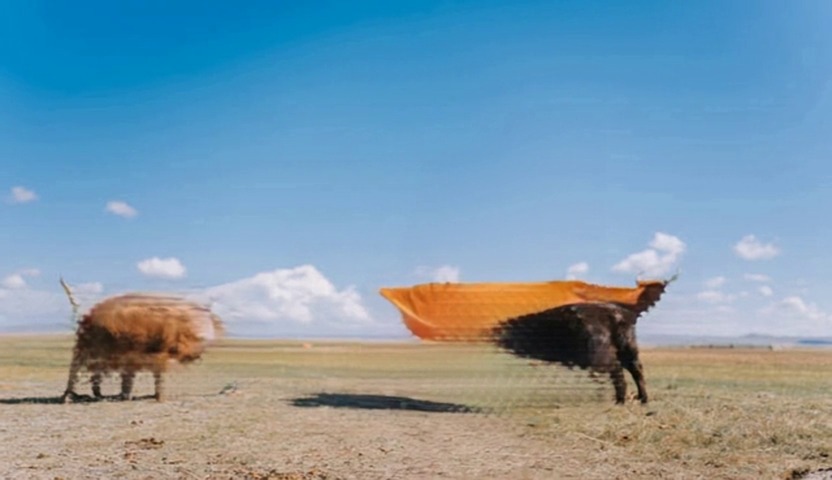}} & \setlength{\fboxrule}{0.3pt}\fcolorbox{gray!60}{white}{\includegraphics[width=\dimexpr0.1475\textwidth-2\fboxrule\relax]{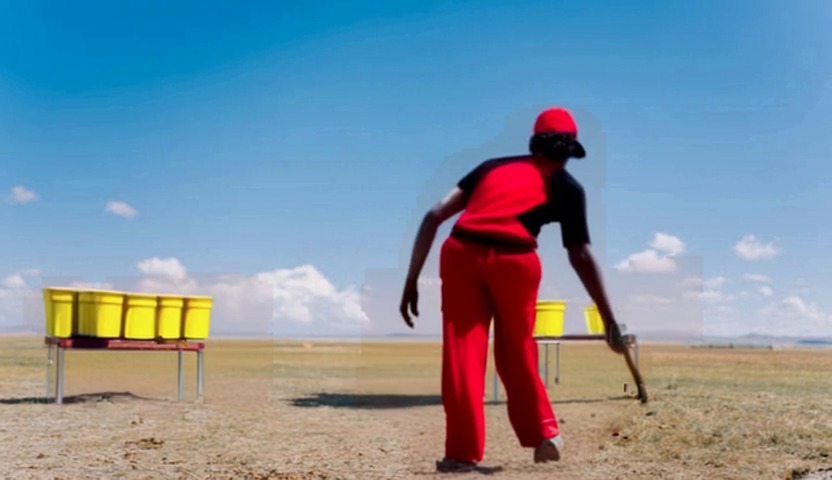}} \\
\scriptsize Example 2\hspace{2pt} & \setlength{\fboxrule}{0.3pt}\fcolorbox{gray!60}{white}{\includegraphics[width=\dimexpr0.1475\textwidth-2\fboxrule\relax]{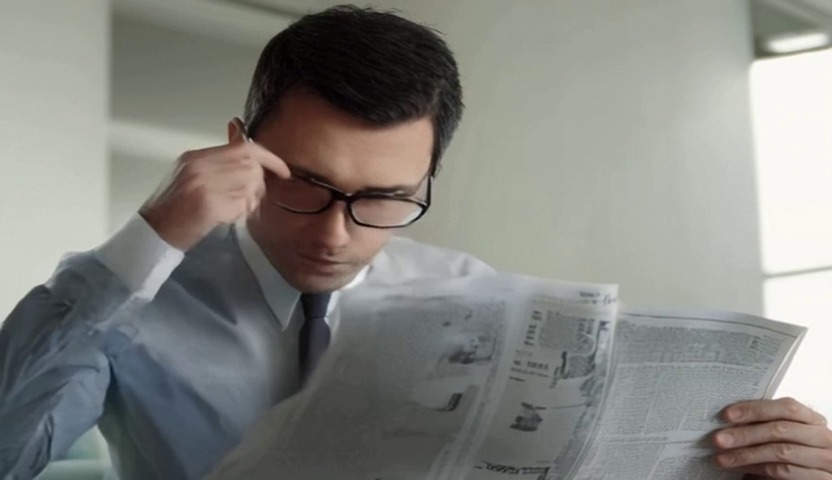}} & \setlength{\fboxrule}{0.3pt}\fcolorbox{gray!60}{white}{\includegraphics[width=\dimexpr0.1475\textwidth-2\fboxrule\relax]{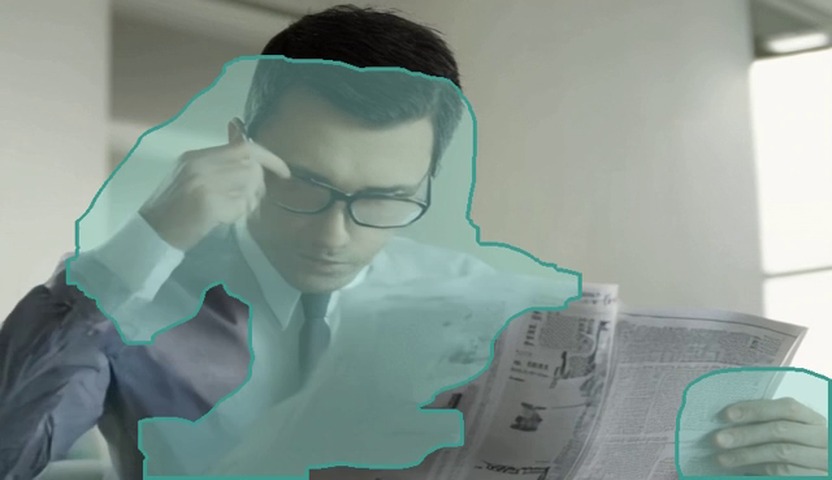}} & \setlength{\fboxrule}{0.8pt}\fcolorbox{artifixteal}{white}{\includegraphics[width=\dimexpr0.1475\textwidth-2\fboxrule\relax]{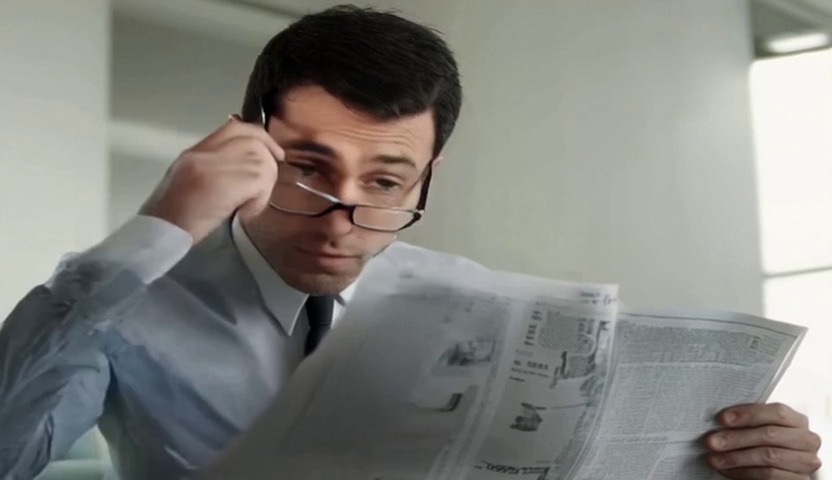}} & \setlength{\fboxrule}{0.3pt}\fcolorbox{gray!60}{white}{\includegraphics[width=\dimexpr0.1475\textwidth-2\fboxrule\relax]{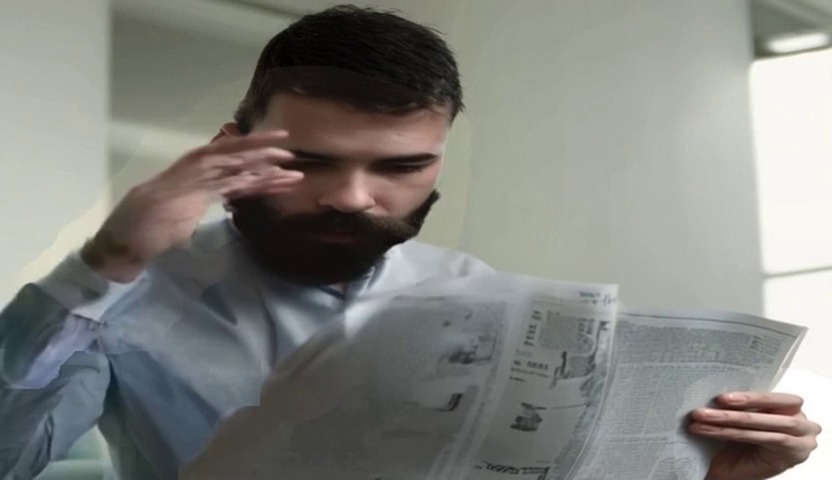}} & \setlength{\fboxrule}{0.3pt}\fcolorbox{gray!60}{white}{\includegraphics[width=\dimexpr0.1475\textwidth-2\fboxrule\relax]{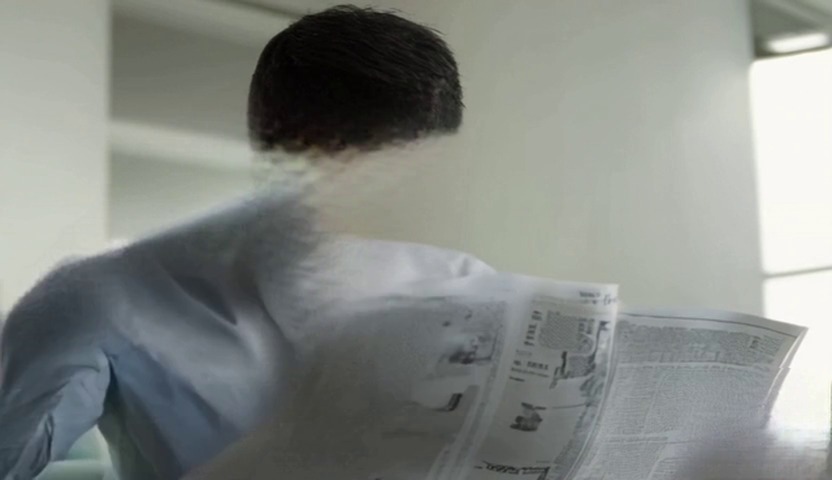}} & \setlength{\fboxrule}{0.3pt}\fcolorbox{gray!60}{white}{\includegraphics[width=\dimexpr0.1475\textwidth-2\fboxrule\relax]{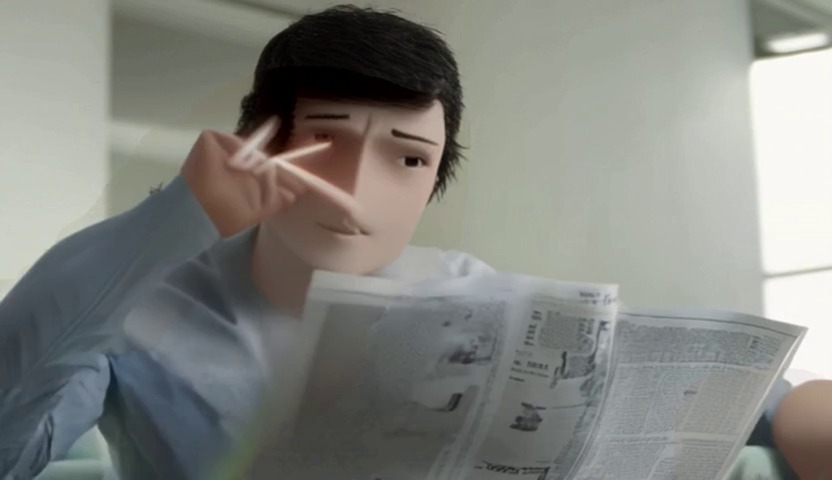}} \\
\scriptsize Example 3\hspace{2pt} & \setlength{\fboxrule}{0.3pt}\fcolorbox{gray!60}{white}{\includegraphics[width=\dimexpr0.1475\textwidth-2\fboxrule\relax]{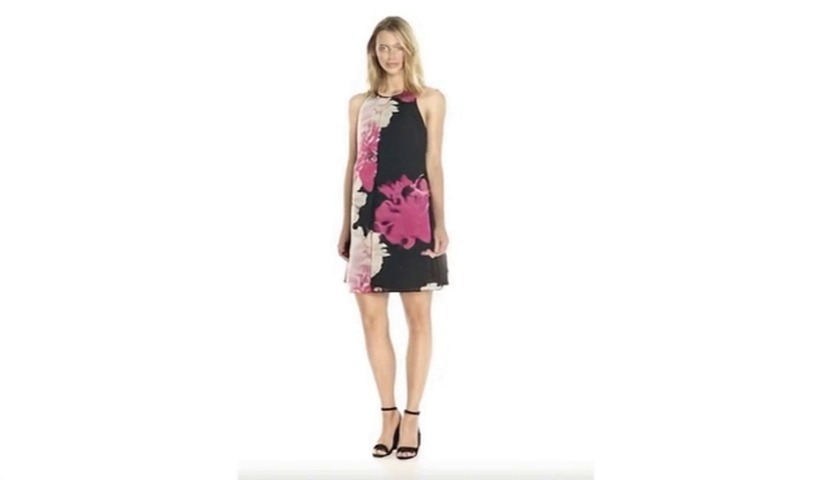}} & \setlength{\fboxrule}{0.3pt}\fcolorbox{gray!60}{white}{\includegraphics[width=\dimexpr0.1475\textwidth-2\fboxrule\relax]{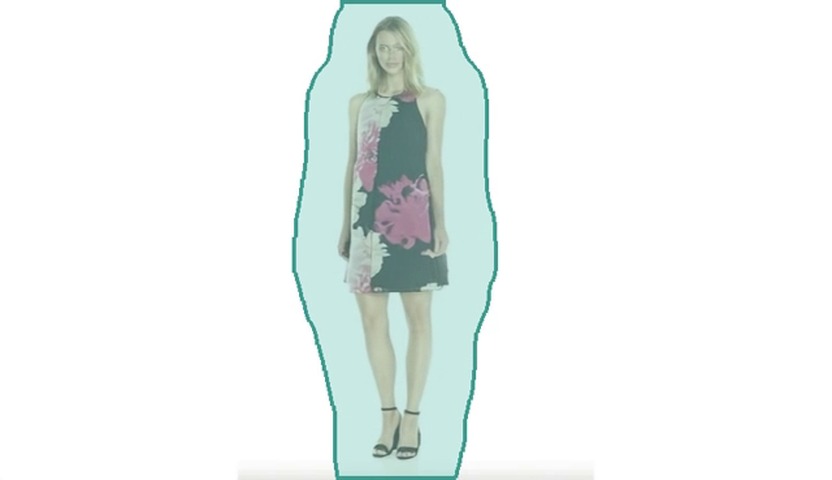}} & \setlength{\fboxrule}{0.8pt}\fcolorbox{artifixteal}{white}{\includegraphics[width=\dimexpr0.1475\textwidth-2\fboxrule\relax]{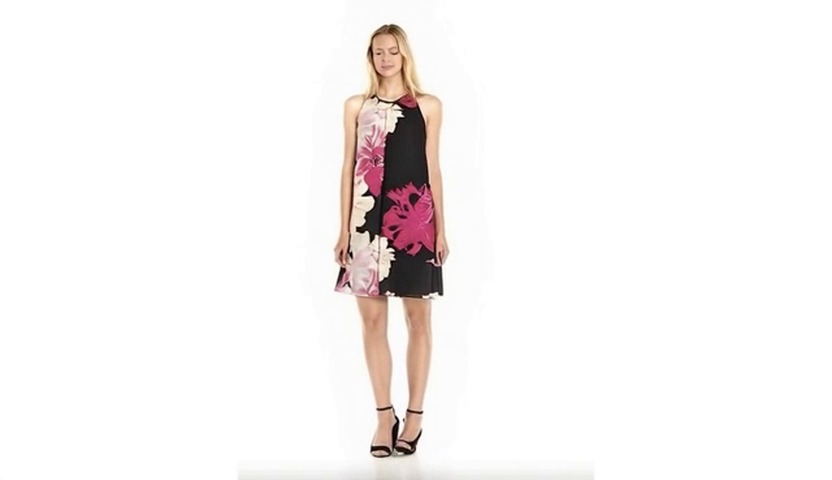}} & \setlength{\fboxrule}{0.3pt}\fcolorbox{gray!60}{white}{\includegraphics[width=\dimexpr0.1475\textwidth-2\fboxrule\relax]{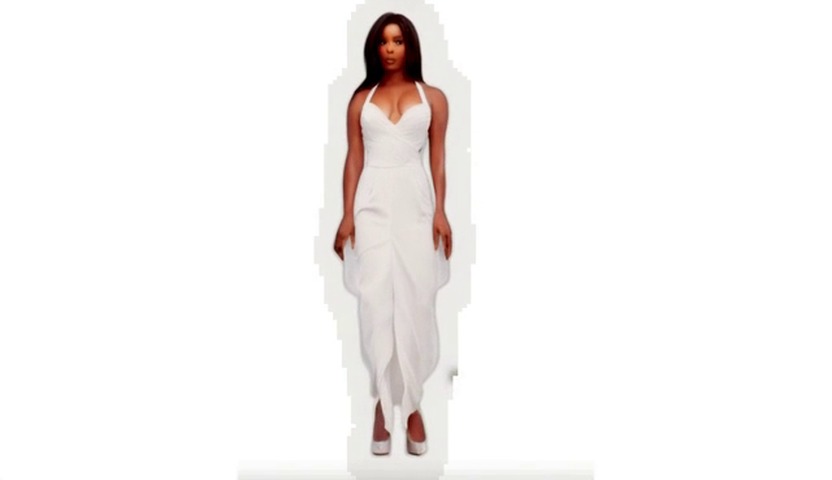}} & \setlength{\fboxrule}{0.3pt}\fcolorbox{gray!60}{white}{\includegraphics[width=\dimexpr0.1475\textwidth-2\fboxrule\relax]{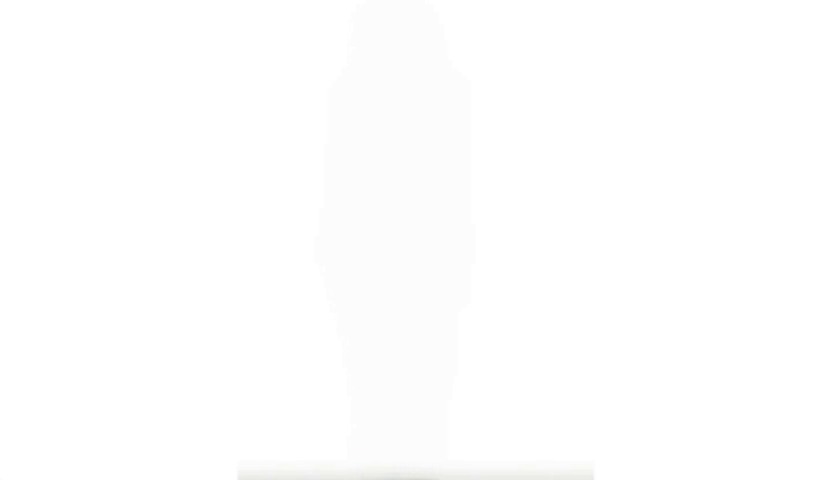}} & \setlength{\fboxrule}{0.3pt}\fcolorbox{gray!60}{white}{\includegraphics[width=\dimexpr0.1475\textwidth-2\fboxrule\relax]{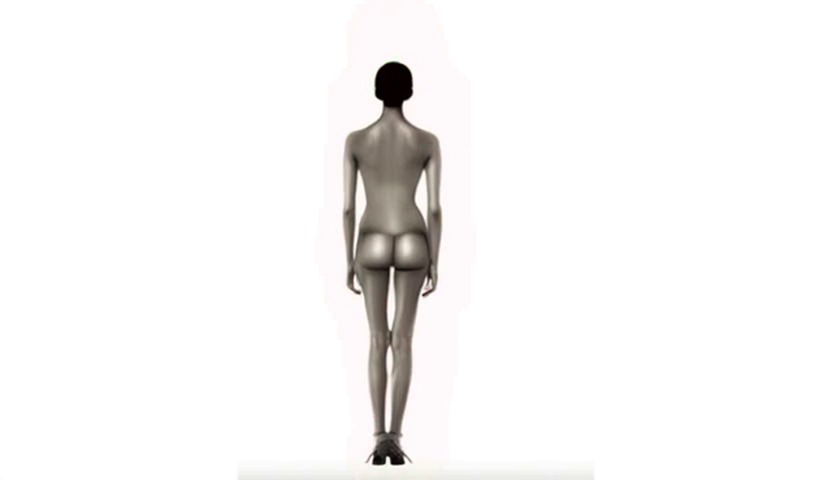}} \\
\scriptsize Example 4\hspace{2pt} & \setlength{\fboxrule}{0.3pt}\fcolorbox{gray!60}{white}{\includegraphics[width=\dimexpr0.1475\textwidth-2\fboxrule\relax]{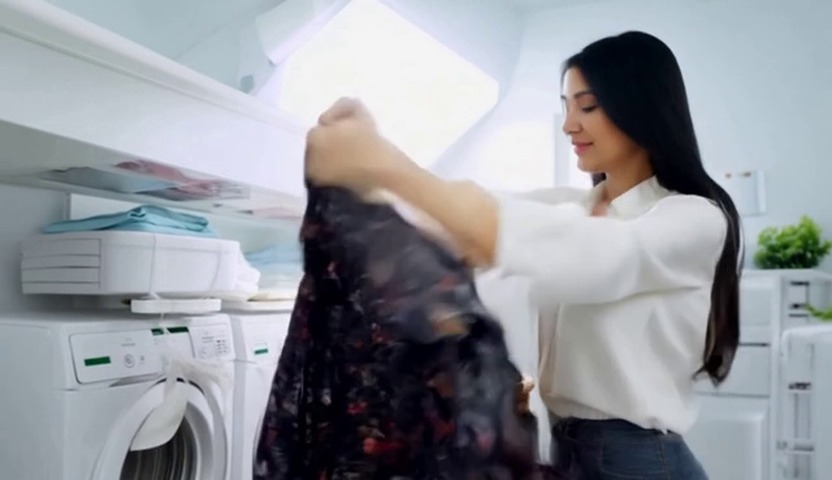}} & \setlength{\fboxrule}{0.3pt}\fcolorbox{gray!60}{white}{\includegraphics[width=\dimexpr0.1475\textwidth-2\fboxrule\relax]{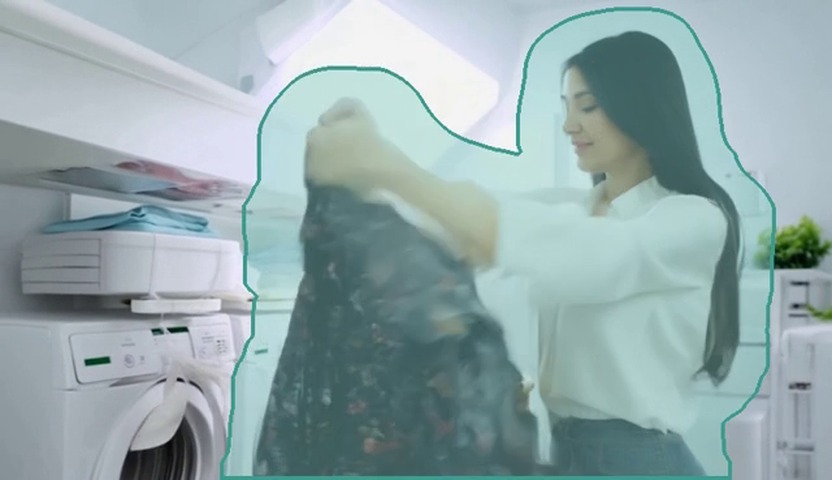}} & \setlength{\fboxrule}{0.8pt}\fcolorbox{artifixteal}{white}{\includegraphics[width=\dimexpr0.1475\textwidth-2\fboxrule\relax]{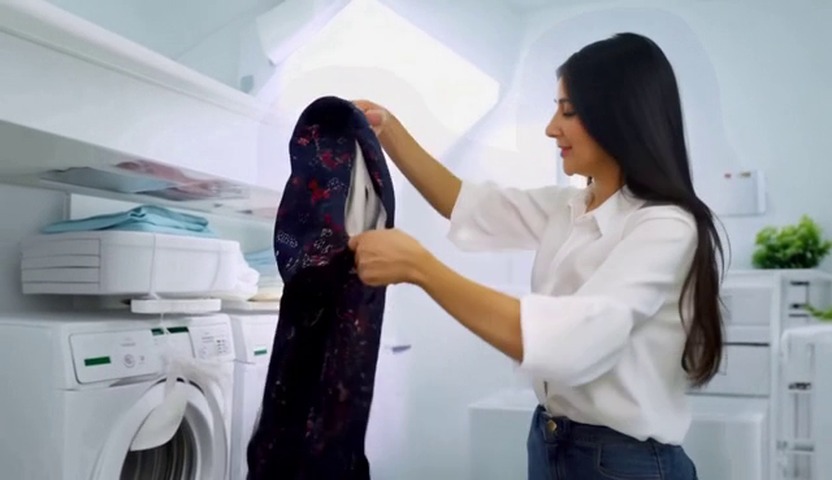}} & \setlength{\fboxrule}{0.3pt}\fcolorbox{gray!60}{white}{\includegraphics[width=\dimexpr0.1475\textwidth-2\fboxrule\relax]{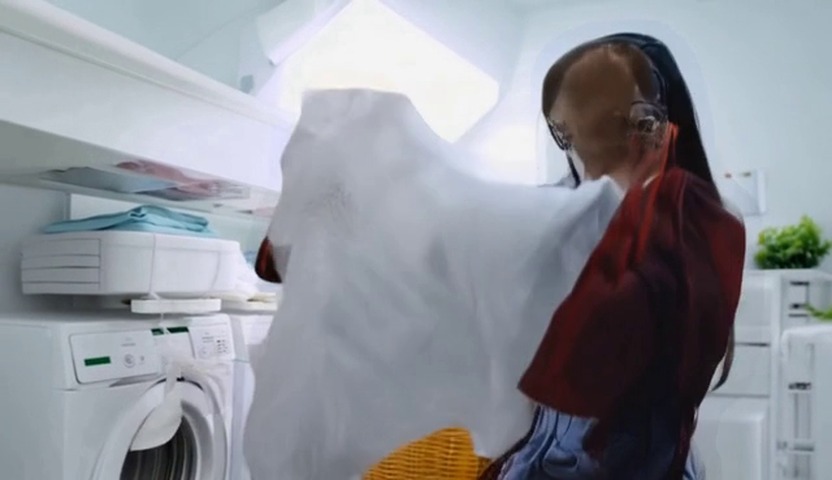}} & \setlength{\fboxrule}{0.3pt}\fcolorbox{gray!60}{white}{\includegraphics[width=\dimexpr0.1475\textwidth-2\fboxrule\relax]{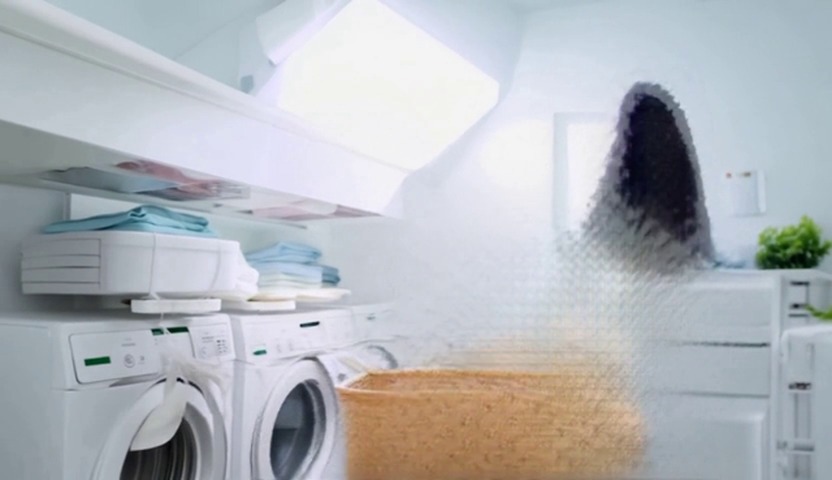}} & \setlength{\fboxrule}{0.3pt}\fcolorbox{gray!60}{white}{\includegraphics[width=\dimexpr0.1475\textwidth-2\fboxrule\relax]{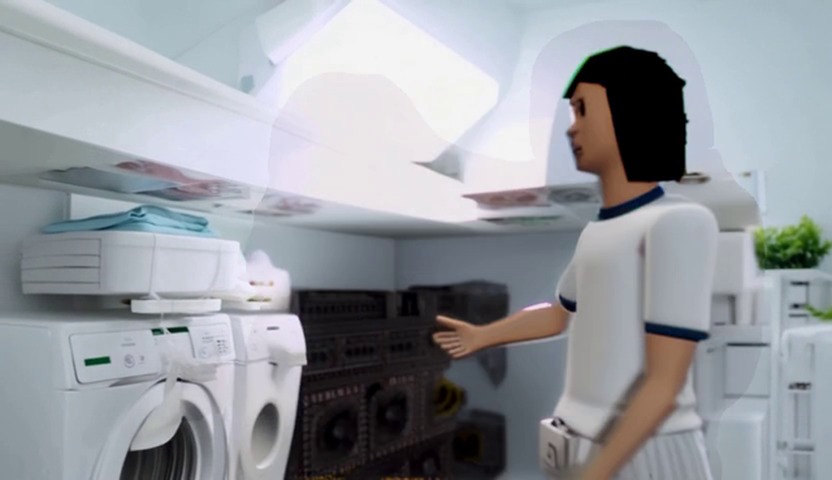}} \\
\scriptsize Example 5\hspace{2pt} & \setlength{\fboxrule}{0.3pt}\fcolorbox{gray!60}{white}{\includegraphics[width=\dimexpr0.1475\textwidth-2\fboxrule\relax]{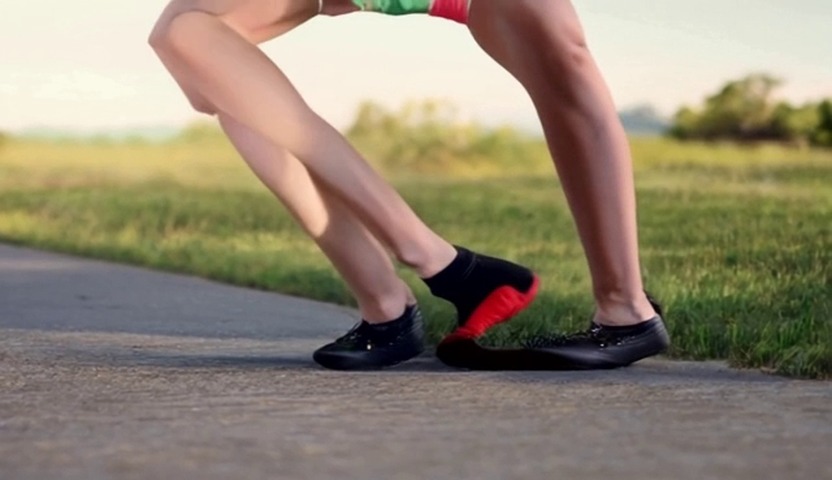}} & \setlength{\fboxrule}{0.3pt}\fcolorbox{gray!60}{white}{\includegraphics[width=\dimexpr0.1475\textwidth-2\fboxrule\relax]{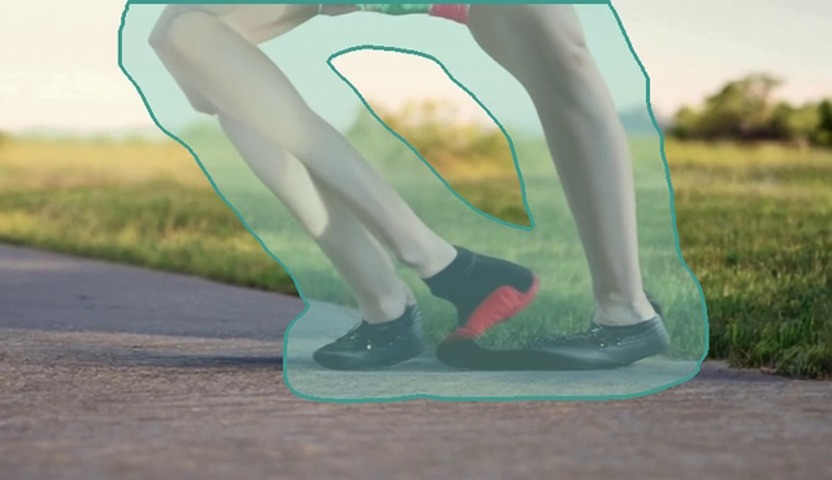}} & \setlength{\fboxrule}{0.8pt}\fcolorbox{artifixteal}{white}{\includegraphics[width=\dimexpr0.1475\textwidth-2\fboxrule\relax]{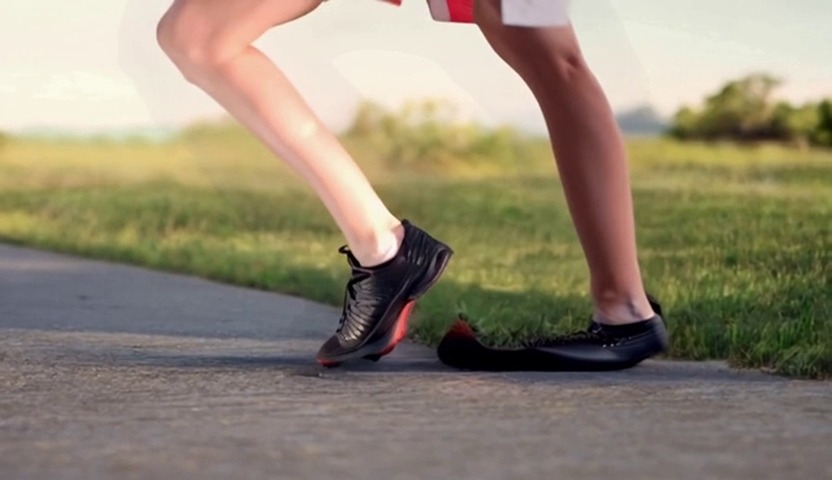}} & \setlength{\fboxrule}{0.3pt}\fcolorbox{gray!60}{white}{\includegraphics[width=\dimexpr0.1475\textwidth-2\fboxrule\relax]{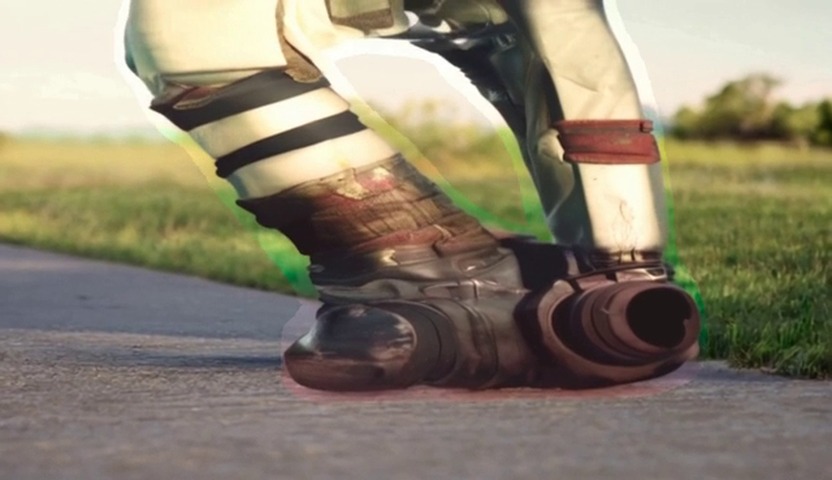}} & \setlength{\fboxrule}{0.3pt}\fcolorbox{gray!60}{white}{\includegraphics[width=\dimexpr0.1475\textwidth-2\fboxrule\relax]{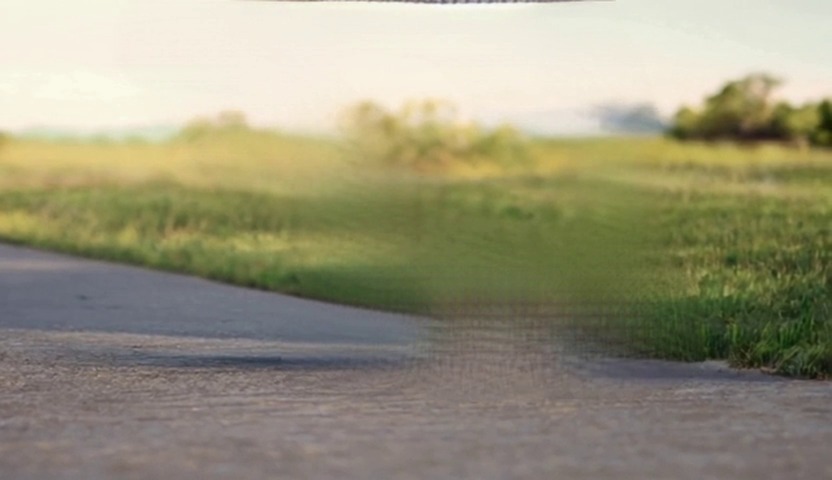}} & \setlength{\fboxrule}{0.3pt}\fcolorbox{gray!60}{white}{\includegraphics[width=\dimexpr0.1475\textwidth-2\fboxrule\relax]{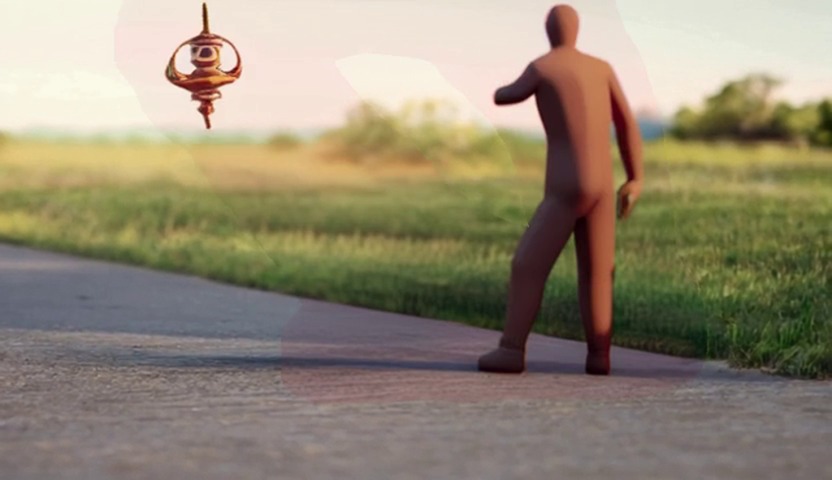}} \\
\end{tabular}}

%% file: tables/t3_localization_full.tex
\begin{table*}[t]
\centering\small
\setlength{\tabcolsep}{3.5pt}
\resizebox{\linewidth}{!}{%
\begin{tabular}{lrrcccccccccc}
\toprule
Localizer & Params & s/clip & J$\uparrow$ & F$\uparrow$ & J\&F$\uparrow$ & Coverage$\uparrow$ & Area & Detection F1$\uparrow$ & Repair judge $\Delta\downarrow$ & Human localize$\uparrow$ & Human fixed$\uparrow$ & Human preserved$\uparrow$ \\
\midrule
BrokenVideos (oracle) & -- & -- & -- & -- & -- & -- & 12\% & -- & +0.38 & 3.94{\scriptsize$\pm$1.16} & 3.18{\scriptsize$\pm$1.00} & 4.40{\scriptsize$\pm$0.78} \\
\midrule
Qwen2.5-VL-72B + SAM3 & 72B & $\sim$140 & 0.24 & 0.06 & 0.15 & 0.97 & 31\% & -- & +0.48 & 3.26{\scriptsize$\pm$0.97} & 2.60{\scriptsize$\pm$0.90} & 3.56{\scriptsize$\pm$0.97} \\
Qwen2.5-VL-32B + SAM3 & 32B & $\sim$140 & 0.23 & 0.05 & 0.14 & \textbf{0.98} & 33\% & -- & +0.40 & -- & -- & -- \\
SAM3 + localizer head~\cite{carion2026sam3} & 0.8B & 52.3 & 0.28 & 0.07 & 0.17 & 0.90 & 19\% & 0.60 & +0.49 & -- & -- & -- \\
DiffDoctor~\cite{wang2025diffdoctor} & 85M & 5.2 & 0.02 & 0.04 & 0.03 & 0.03 & 0\% & 0.50 & \textbf{-0.02} & -- & -- & -- \\
Skyra-RL~\cite{li2026skyra} & 7B & 57.6 & 0.05 & 0.03 & 0.04 & 0.29 & 17\% & 0.62 & +0.01 & -- & -- & -- \\
SAMWISE (fine-tuned)~\cite{cuttano2025samwise} & 202M & 4.7 & \textbf{0.34} & \textbf{0.34} & \textbf{0.34} & 0.72 & 11\% & 0.60 & +0.55 & -- & -- & -- \\
\textbf{\method} & 45M & 1.3 & 0.21 & 0.05 & 0.13 & 0.97 & 33\% & \textbf{0.92} & +0.30 & \textbf{3.63}{\scriptsize$\pm$0.80} & \textbf{3.10}{\scriptsize$\pm$1.22} & \textbf{4.31}{\scriptsize$\pm$0.95} \\
\bottomrule
\end{tabular}}
\caption{\textbf{Among the localizers tested on clean videos, only ours both covers the artifact and stays empty on most clean videos.} Timings from a separate run. Localization. Every region source feeds the same inpainter (ours, generic prompt). J (region similarity) and F (boundary measure, DAVIS protocol) are measured against the BrokenVideos artifact masks, which are hand-drawn around the visibly broken pixels only, so they reward tight masks. Coverage is the share of labeled artifact pixels inside the region. Detection F1 counts a frame as detected when the region is non-empty, with the annotated artifact frames as positives and every frame of 300 held-out real videos without artifacts as negatives (not run for the VLM pipelines). The VLM rows plan the repair without the human annotation and segment the named objects with SAM3, and their time includes the VLM ($\sim$95\,s on two GPUs). Human columns are MOS (1--5) recovered with SUREAL on the 7 rated clips that have dataset labels, from the localization trials of the perceptual study, all with our inpainter without preference tuning, $\pm$ half the 95\% CI. $^\dagger$ differs from our localizer in the same trials (Wilcoxon over clips, Holm-corrected, $p<.05$). Best value per column in bold.}
\label{tab:supp_localization}
\end{table*}

%% file: figures/grid_fp.tex
\providecolor{artifixteal}{HTML}{3E9C8F}%
{\setlength{\tabcolsep}{0.6pt}\renewcommand{\arraystretch}{0.6}\setlength{\fboxsep}{0pt}%
\begin{tabular}{@{}cc@{\hspace{3pt}}cc@{\hspace{3pt}}cc@{\hspace{3pt}}cc@{}}
\scriptsize Input and false region & \scriptsize\textbf{\textcolor{artifixteal}{ArtiFix}} & \scriptsize Input and false region & \scriptsize\textbf{\textcolor{artifixteal}{ArtiFix}} & \scriptsize Input and false region & \scriptsize\textbf{\textcolor{artifixteal}{ArtiFix}} & \scriptsize Input and false region & \scriptsize\textbf{\textcolor{artifixteal}{ArtiFix}} \\[1pt]
\setlength{\fboxrule}{0.3pt}\fcolorbox{gray!60}{white}{\includegraphics[width=\dimexpr0.1187\textwidth-2\fboxrule\relax]{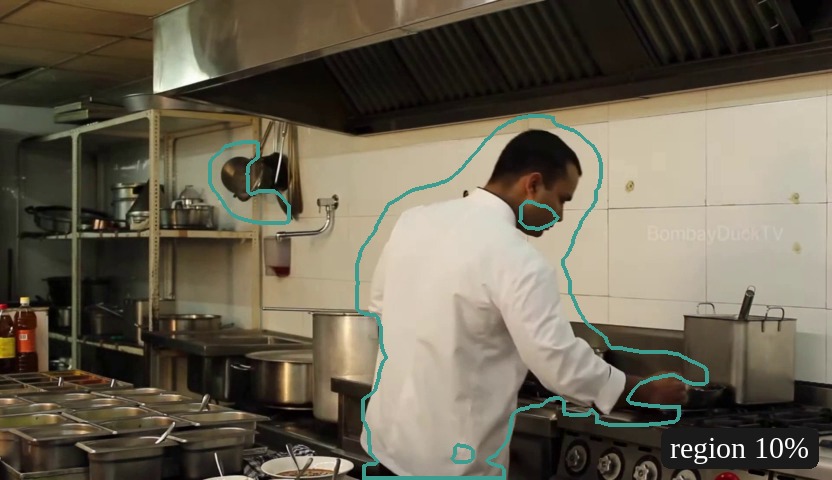}} & \setlength{\fboxrule}{0.8pt}\fcolorbox{artifixteal}{white}{\includegraphics[width=\dimexpr0.1187\textwidth-2\fboxrule\relax]{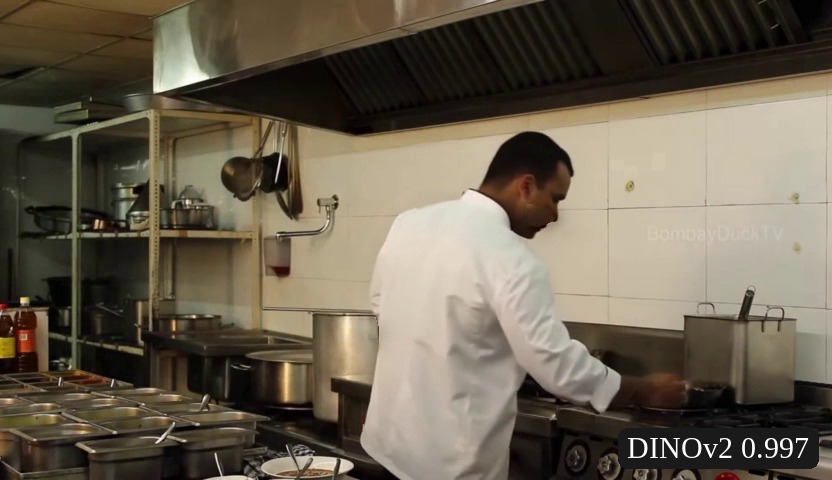}} & \setlength{\fboxrule}{0.3pt}\fcolorbox{gray!60}{white}{\includegraphics[width=\dimexpr0.1187\textwidth-2\fboxrule\relax]{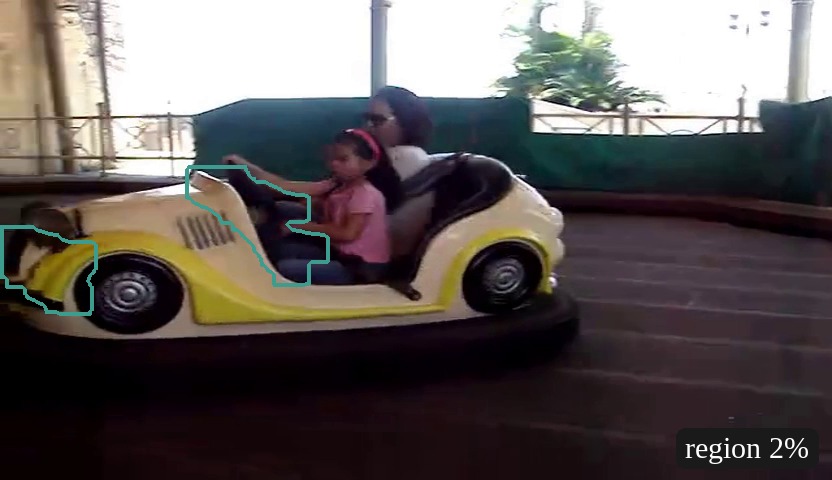}} & \setlength{\fboxrule}{0.8pt}\fcolorbox{artifixteal}{white}{\includegraphics[width=\dimexpr0.1187\textwidth-2\fboxrule\relax]{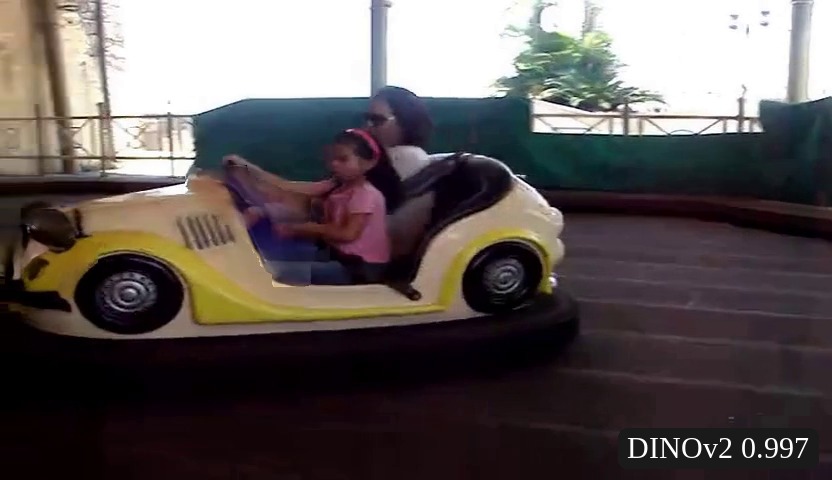}} & \setlength{\fboxrule}{0.3pt}\fcolorbox{gray!60}{white}{\includegraphics[width=\dimexpr0.1187\textwidth-2\fboxrule\relax]{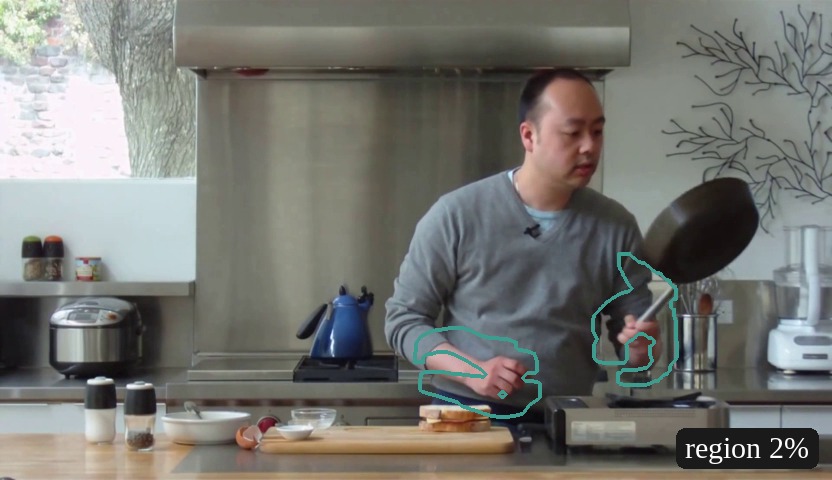}} & \setlength{\fboxrule}{0.8pt}\fcolorbox{artifixteal}{white}{\includegraphics[width=\dimexpr0.1187\textwidth-2\fboxrule\relax]{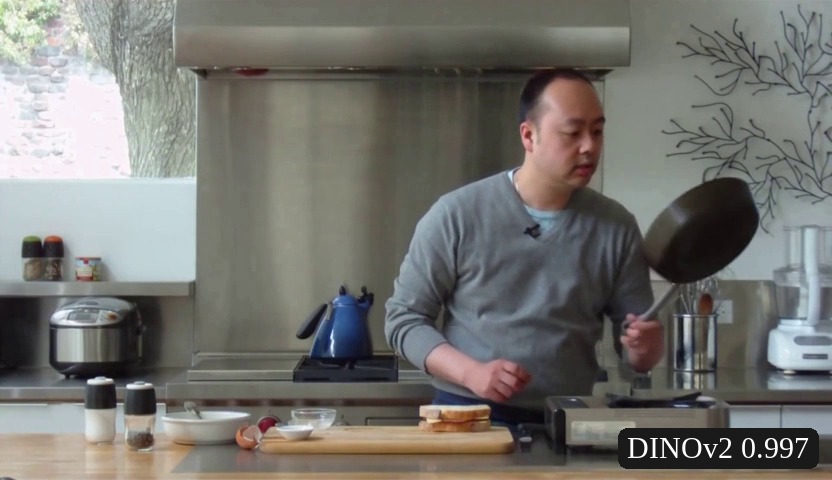}} & \setlength{\fboxrule}{0.3pt}\fcolorbox{gray!60}{white}{\includegraphics[width=\dimexpr0.1187\textwidth-2\fboxrule\relax]{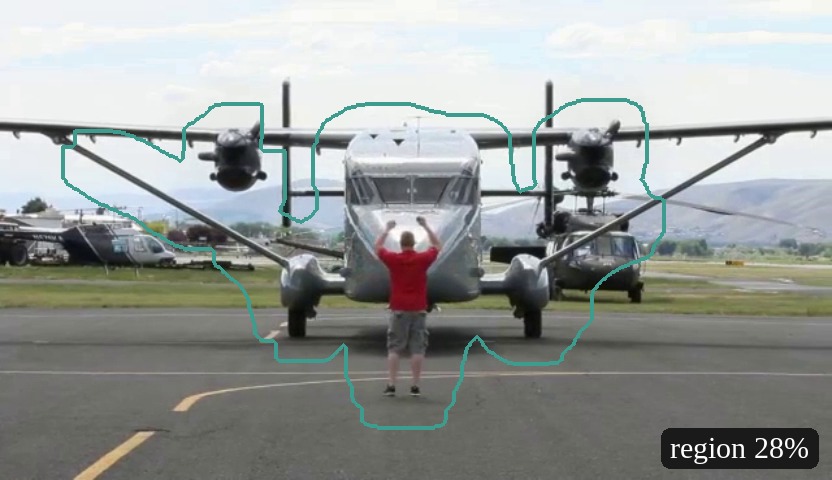}} & \setlength{\fboxrule}{0.8pt}\fcolorbox{artifixteal}{white}{\includegraphics[width=\dimexpr0.1187\textwidth-2\fboxrule\relax]{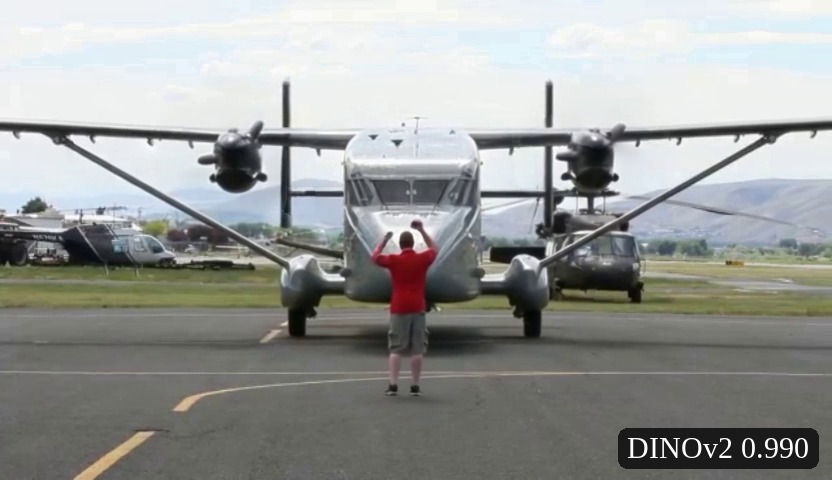}} \\
\setlength{\fboxrule}{0.3pt}\fcolorbox{gray!60}{white}{\includegraphics[width=\dimexpr0.1187\textwidth-2\fboxrule\relax]{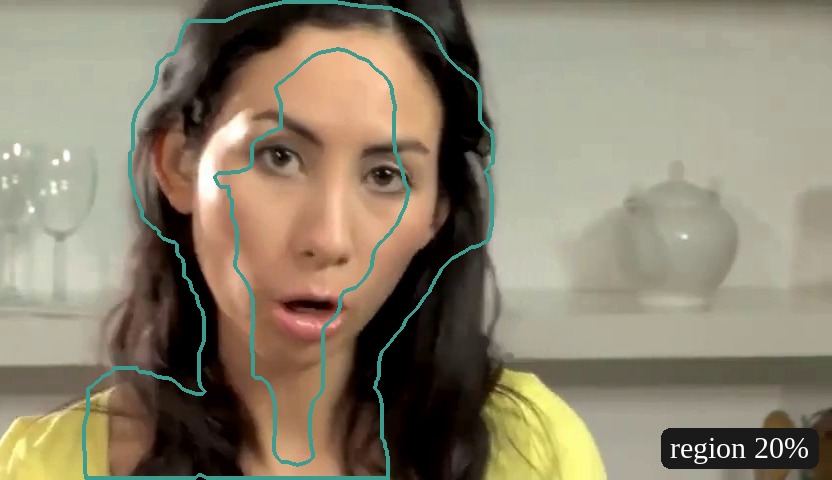}} & \setlength{\fboxrule}{0.8pt}\fcolorbox{artifixteal}{white}{\includegraphics[width=\dimexpr0.1187\textwidth-2\fboxrule\relax]{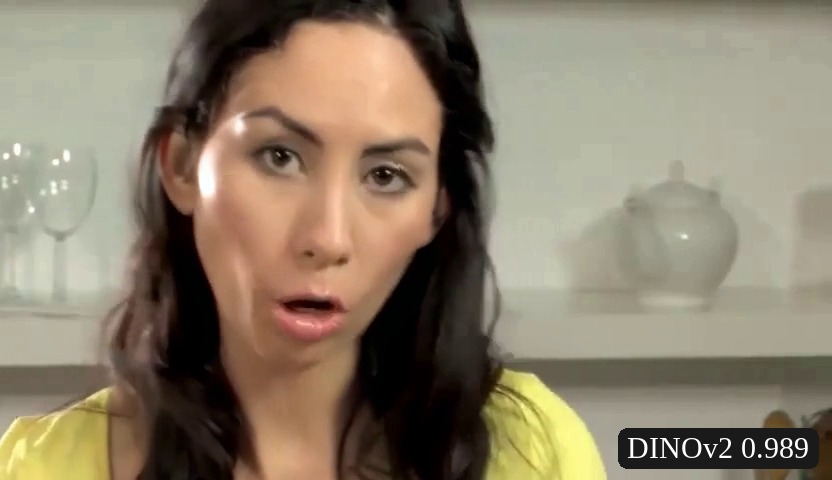}} & \setlength{\fboxrule}{0.3pt}\fcolorbox{gray!60}{white}{\includegraphics[width=\dimexpr0.1187\textwidth-2\fboxrule\relax]{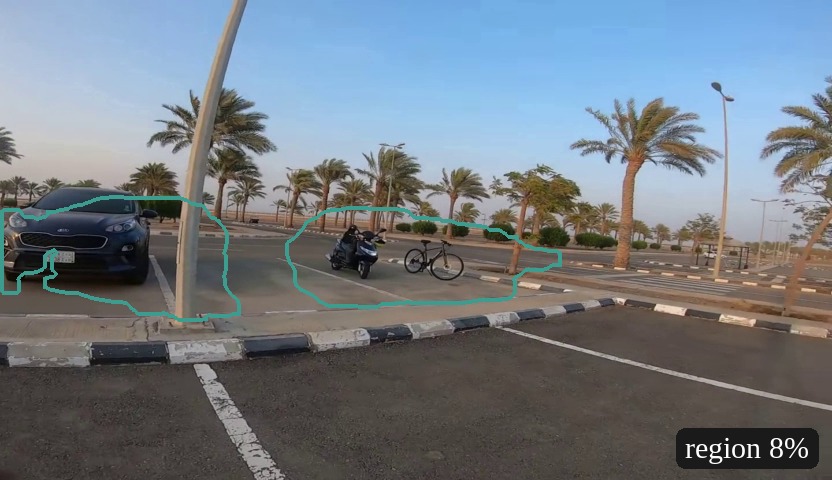}} & \setlength{\fboxrule}{0.8pt}\fcolorbox{artifixteal}{white}{\includegraphics[width=\dimexpr0.1187\textwidth-2\fboxrule\relax]{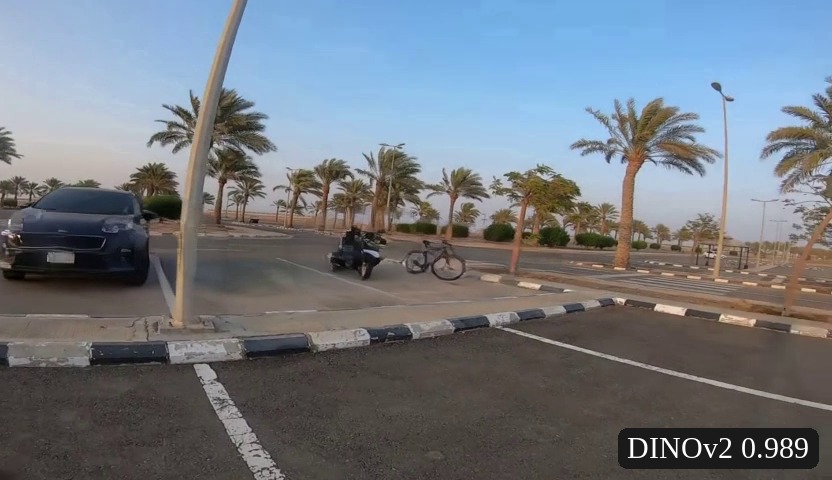}} & \setlength{\fboxrule}{0.3pt}\fcolorbox{gray!60}{white}{\includegraphics[width=\dimexpr0.1187\textwidth-2\fboxrule\relax]{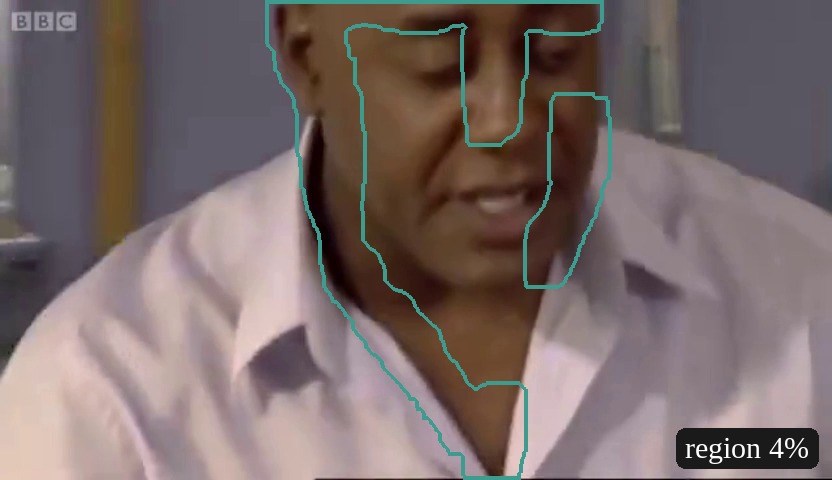}} & \setlength{\fboxrule}{0.8pt}\fcolorbox{artifixteal}{white}{\includegraphics[width=\dimexpr0.1187\textwidth-2\fboxrule\relax]{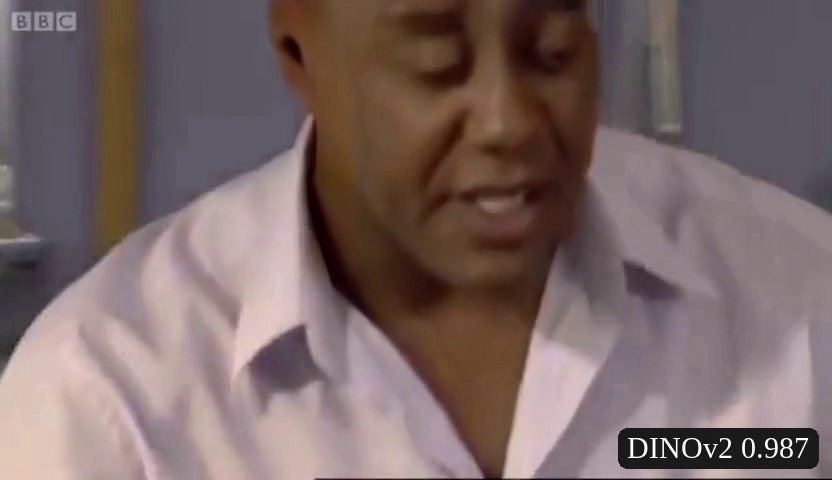}} & \setlength{\fboxrule}{0.3pt}\fcolorbox{gray!60}{white}{\includegraphics[width=\dimexpr0.1187\textwidth-2\fboxrule\relax]{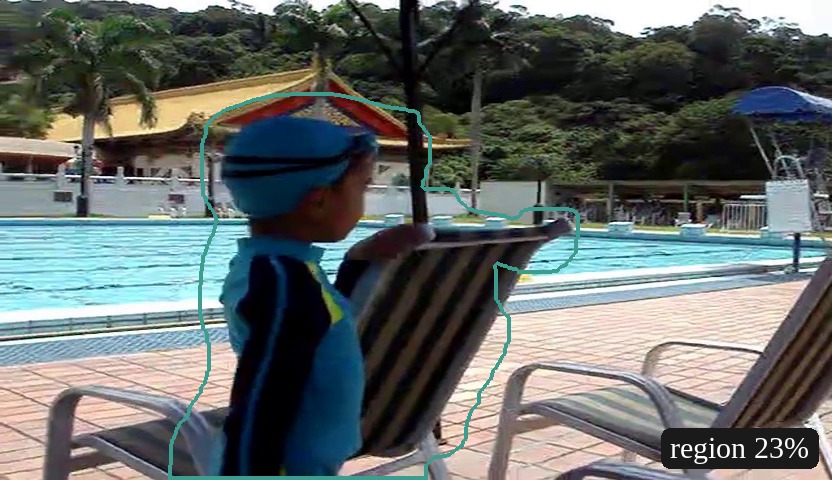}} & \setlength{\fboxrule}{0.8pt}\fcolorbox{artifixteal}{white}{\includegraphics[width=\dimexpr0.1187\textwidth-2\fboxrule\relax]{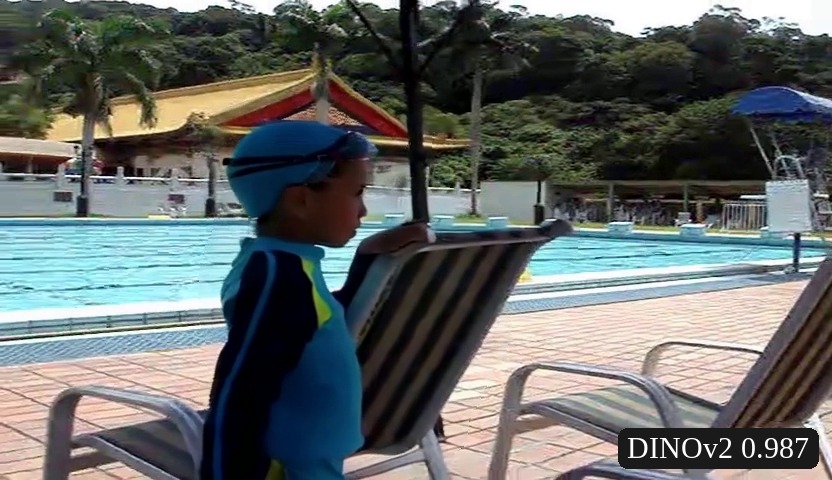}} \\
\end{tabular}}

%% file: tables/s6_ladder_full.tex
\begin{table}[t]
\centering\footnotesize
\setlength{\tabcolsep}{2.2pt}
\begin{tabular}{lrcccc}
\toprule
Added & s/clip & DiT & LPIPS$\downarrow$ & DINOv2$\uparrow$ & Judge $\Delta\downarrow$ \\
\midrule
0. Regenerate (0.85) & 99 & 84\% & 0.371 & 0.711 & \textbf{-0.72} \\
1. Region, hidden & 115 & 100\% & 0.222 & 0.591 & +1.69 \\
\quad \emph{visible instead} & 115 & 100\% & 0.048 & 0.966 & +0.59 \\
2. + reference cut-out & 125 & 100\% & 0.121 & \textbf{0.907} & +0.33 \\
3. + sparse compute & \textbf{77} & \textbf{60}\% & 0.123 & 0.905 & +0.33 \\
4. + edge blending & \textbf{77} & \textbf{60}\% & 0.122 & 0.906 & +0.30 \\
5. + tuning (\textbf{\method}) & 83 & \textbf{60}\% & \textbf{0.119} & 0.906 & +0.13 \\
\bottomrule
\end{tabular}
\caption{\textbf{The reference cut-out brings the gain in preservation, and region-sparse denoising the saving in compute.} Full component ladder on the 103 test clips, on our regions. Each row adds one component. DiT is the share of tokens computed, and LPIPS and DINOv2 measure preservation.}
\label{tab:supp_ladder}
\end{table}

%% file: tables/t5_ablations.tex
\begin{table}[t]
\centering\footnotesize
\begin{tabular}{lccc}
\toprule
Variant & Judge $\Delta\downarrow$ & DINOv2$\uparrow$ & CLIP$\uparrow$ \\
\midrule
\multicolumn{4}{l}{\emph{Prompt (teacher regions)}} \\
\quad 72B physics description & \textbf{+0.21} & \textbf{0.907} & \textbf{0.938} \\
\quad generic (\method) & \textbf{+0.21} & 0.904 & 0.937 \\
\midrule
\multicolumn{4}{l}{\emph{Preference-tuning reward (our regions)}} \\
\quad none & +0.30 & \textbf{0.906} & \textbf{0.938} \\
\quad CLIP + localizer & +0.68 & 0.904 & 0.933 \\
\quad Video Jev (\method) & \textbf{+0.13} & \textbf{0.906} & \textbf{0.938} \\
\bottomrule
\end{tabular}
\caption{\textbf{The generic prompt is enough, and the ranking signal of preference tuning matters.} Ablations on the 103 test clips. The prompt rows use the teacher's regions so that they isolate the inpainter, and the reward rows use our localizer's regions. Best value per group in bold.}
\label{tab:ablations}
\end{table}

%% file: tables/s5_jev.tex

\begin{table}[t]
\centering\footnotesize
\begin{tabular}{lccc}
\toprule
Seed picked by & Judge $\Delta\downarrow$ & LPIPS$\downarrow$ & DINOv2$\uparrow$ \\
\midrule
Random & +0.41 & 0.124 & 0.897 \\
Highest Video Jev (used for tuning) & \textbf{+0.00} & \textbf{0.120} & 0.905 \\
Lowest Video Jev & +0.74 & 0.128 & 0.884 \\
CLIP + localizer reward & +0.35 & \textbf{0.120} & \textbf{0.906} \\
\bottomrule
\end{tabular}
\caption{\textbf{Within one repair, Video Jev picks the cleanest of several seeds.} Picking one of four seeds of the pipeline without preference tuning on the 103 test clips. The seed with the highest Video Jev score gives the cleanest repairs at no cost in preservation.}
\label{tab:supp_jev}
\end{table}

%% file: figures/supp_block_jev.tex
\Cref{fig:supp_jev} shows one of the preference pairs, where the seed with the lower Video Jev score adds a headset and a laptop that are not in the original.

\begin{figure}[t]
\centering
\input{figures/grid_jev}
\caption{\textbf{Video Jev scores the repair without added objects higher.} A preference pair from the training data. Two seeds of the pipeline without preference tuning repair the same clip, and Video Jev scores the middle one higher. The lower-scored repair adds objects that are not in the original.}
\label{fig:supp_jev}
\end{figure}

%% file: figures/grid_jev.tex
\providecolor{artifixteal}{HTML}{3E9C8F}%
{\setlength{\tabcolsep}{0.6pt}\renewcommand{\arraystretch}{0.6}\setlength{\fboxsep}{0pt}%
\begin{tabular}{@{}rccc@{}}
 & \scriptsize Original & \scriptsize\textbf{\textcolor{artifixteal}{Higher Video Jev}} & \scriptsize Lower Video Jev \\[1pt]
\scriptsize Frame 6\hspace{2pt} & \setlength{\fboxrule}{0.3pt}\fcolorbox{gray!60}{white}{\includegraphics[width=\dimexpr0.2650\linewidth-2\fboxrule\relax]{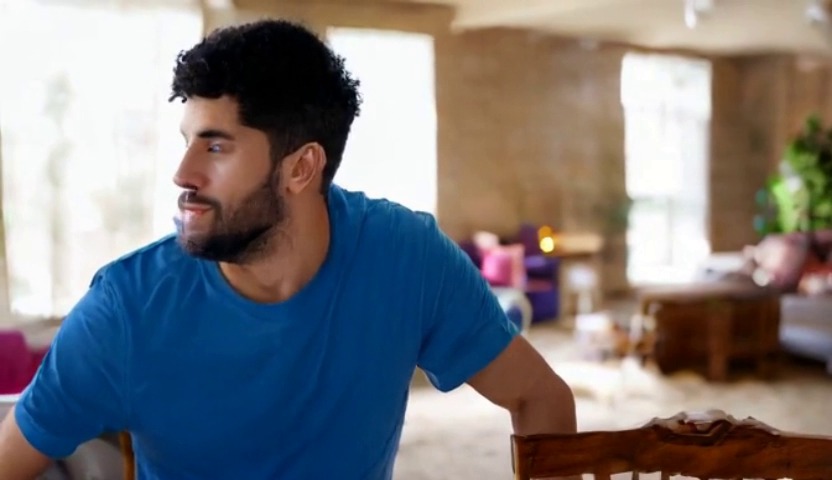}} & \setlength{\fboxrule}{0.8pt}\fcolorbox{artifixteal}{white}{\includegraphics[width=\dimexpr0.2650\linewidth-2\fboxrule\relax]{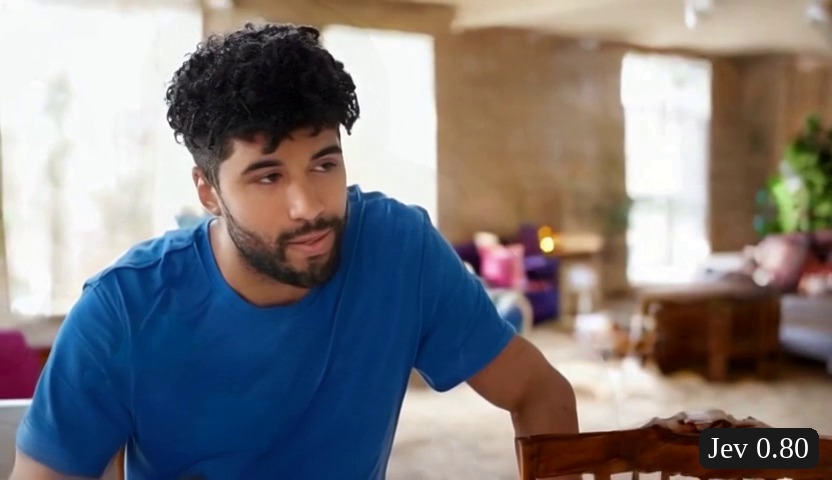}} & \setlength{\fboxrule}{0.3pt}\fcolorbox{gray!60}{white}{\includegraphics[width=\dimexpr0.2650\linewidth-2\fboxrule\relax]{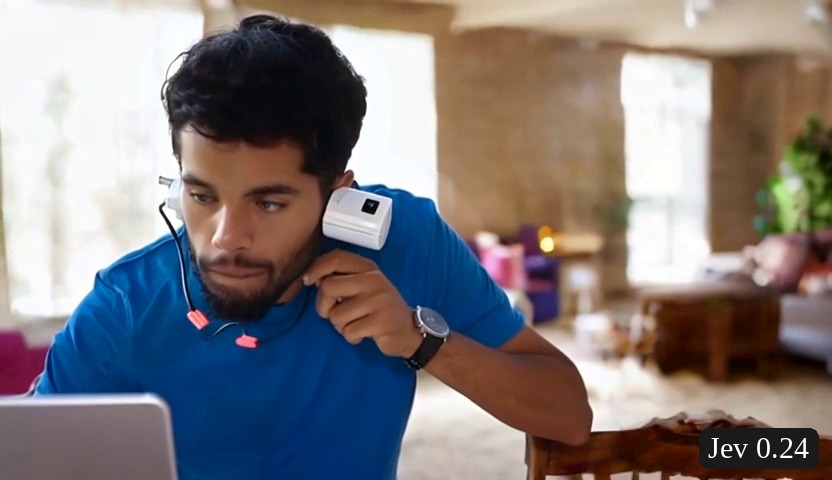}} \\
\scriptsize Frame 24\hspace{2pt} & \setlength{\fboxrule}{0.3pt}\fcolorbox{gray!60}{white}{\includegraphics[width=\dimexpr0.2650\linewidth-2\fboxrule\relax]{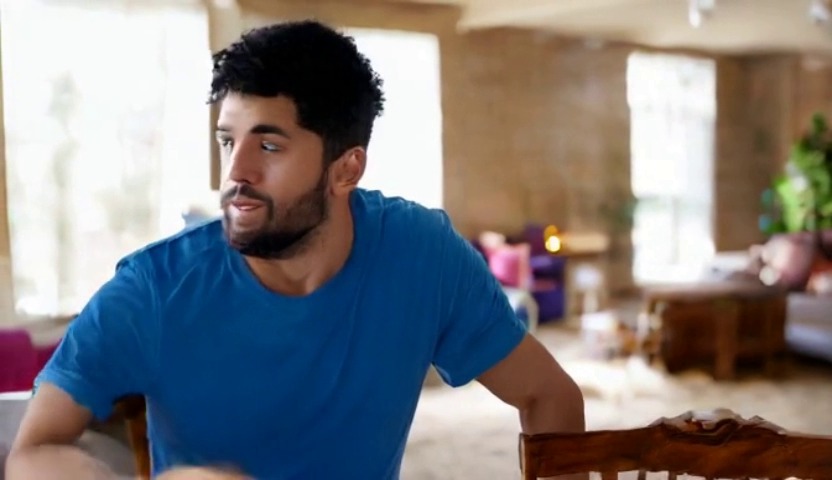}} & \setlength{\fboxrule}{0.8pt}\fcolorbox{artifixteal}{white}{\includegraphics[width=\dimexpr0.2650\linewidth-2\fboxrule\relax]{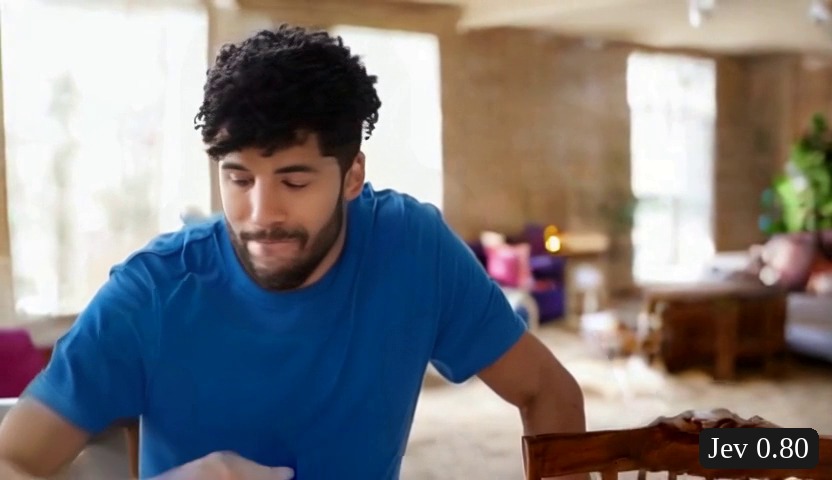}} & \setlength{\fboxrule}{0.3pt}\fcolorbox{gray!60}{white}{\includegraphics[width=\dimexpr0.2650\linewidth-2\fboxrule\relax]{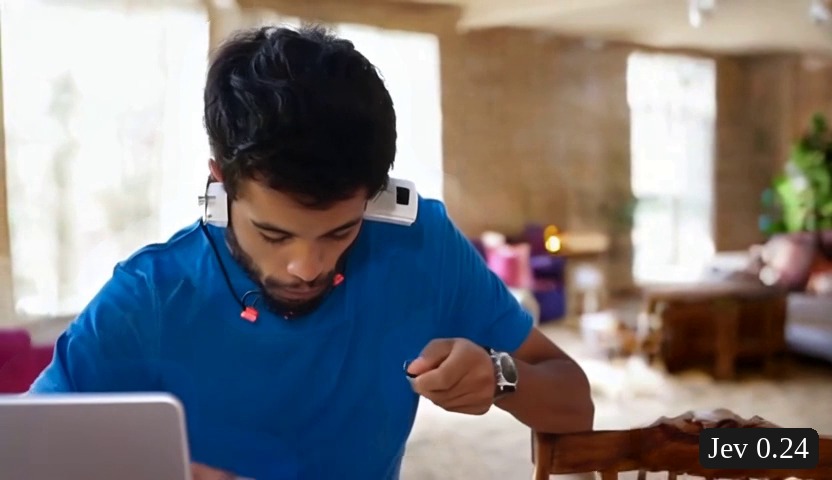}} \\
\scriptsize Frame 42\hspace{2pt} & \setlength{\fboxrule}{0.3pt}\fcolorbox{gray!60}{white}{\includegraphics[width=\dimexpr0.2650\linewidth-2\fboxrule\relax]{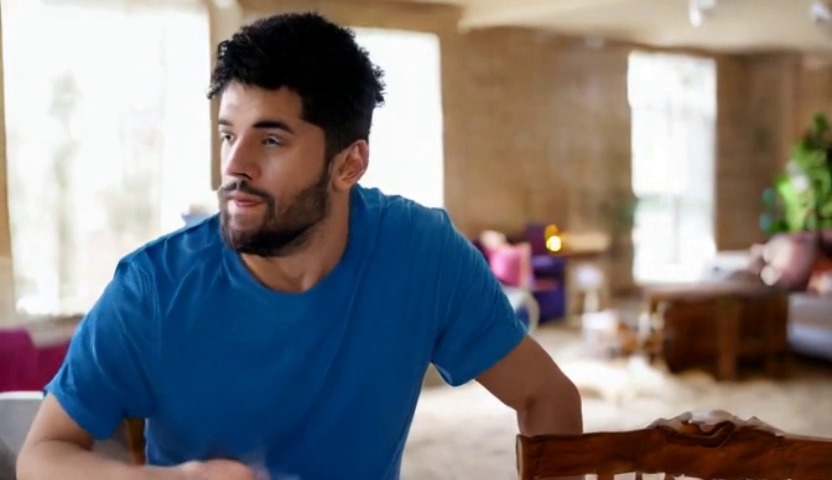}} & \setlength{\fboxrule}{0.8pt}\fcolorbox{artifixteal}{white}{\includegraphics[width=\dimexpr0.2650\linewidth-2\fboxrule\relax]{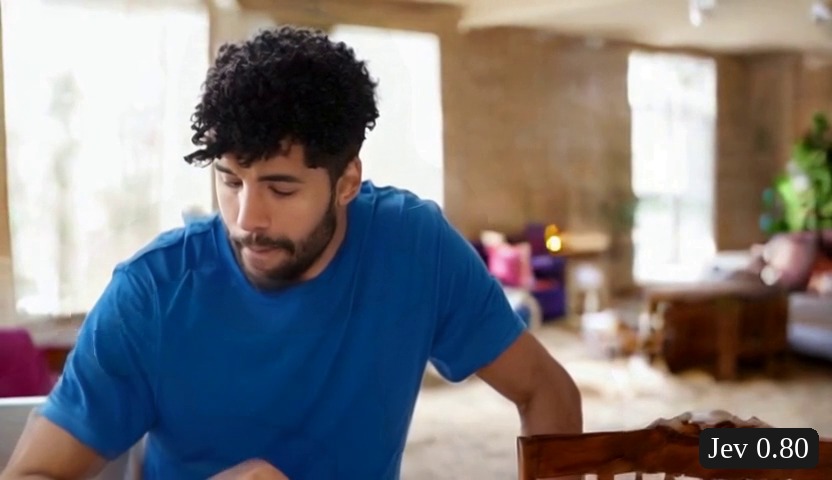}} & \setlength{\fboxrule}{0.3pt}\fcolorbox{gray!60}{white}{\includegraphics[width=\dimexpr0.2650\linewidth-2\fboxrule\relax]{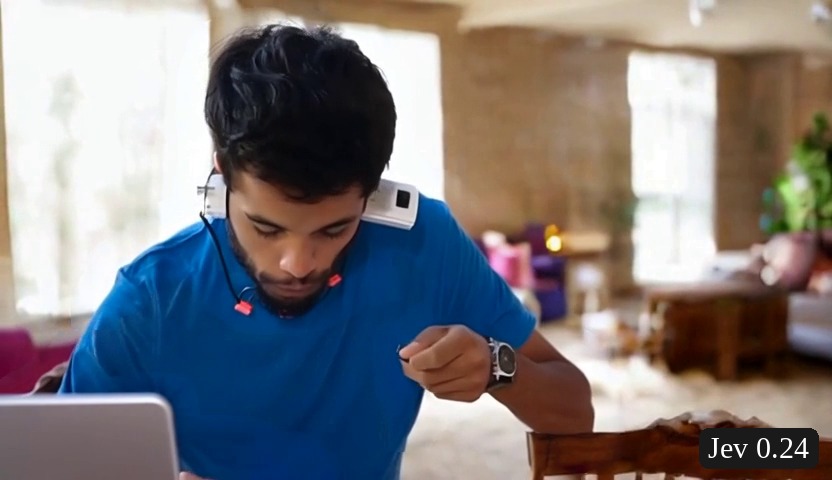}} \\
\end{tabular}}

%% file: tables/t6_dpo_sweep.tex
\begin{table}[t]
\centering\small
\setlength{\tabcolsep}{3pt}
\resizebox{\linewidth}{!}{%
\begin{tabular}{llccccc}
\toprule
Rank & Learning rate & Judge $\Delta\downarrow$ & Cleaner & LPIPS$\downarrow$ & DINOv2$\uparrow$ \\
\midrule
\multicolumn{2}{l}{no tuning} & +0.30 & 41 & 0.122 & 0.906 \\
\midrule
8 & $4\times10^{-6}$ & +0.33 & 40 & 0.121 & 0.905 \\
8 & $2\times10^{-5}$ & +0.18 & \textbf{53} & \textbf{0.118} & 0.907 \\
8 & $1\times10^{-4}$ & +0.17 & 44 & 0.120 & 0.904 \\
16 & $4\times10^{-6}$ & +0.31 & 42 & 0.121 & 0.906 \\
16 & $2\times10^{-5}$ & \textbf{+0.10} & \textbf{53} & \textbf{0.118} & 0.907 \\
16 & $1\times10^{-4}$ & +0.98 & 22 & 0.143 & 0.872 \\
32 & $2\times10^{-5}$ (\method) & +0.13 & 49 & 0.119 & 0.906 \\
64 & $4\times10^{-6}$ & +0.23 & 47 & 0.120 & 0.903 \\
64 & $2\times10^{-5}$ & +0.21 & 42 & \textbf{0.118} & \textbf{0.909} \\
64 & $1\times10^{-4}$ & +3.84 & 8 & 0.249 & 0.584 \\
\bottomrule
\end{tabular}}
\caption{\textbf{At a learning rate of $2\times10^{-5}$, preference tuning helps at every rank.} Preference tuning with different LoRA ranks and learning rates (200 steps, the same 178 pairs), on the 103 test clips with our localizer's regions. Cleaner counts the clips whose repair the judge rates cleaner than the input. Best values in bold.}
\label{tab:dposweep}
\end{table}

%% file: tables/t2_efficiency.tex
\begin{table}[t]
\centering\footnotesize
\begin{tabular}{lccc}
\toprule
Region & Clips & Dense & Region-sparse \\
\midrule
$<$15\% & 19 & 100\% / 125\,s & \textbf{42\% / 56\,s} \\
15--30\% & 39 & 100\% / 125\,s & \textbf{53\% / 69\,s} \\
30--50\% & 33 & 100\% / 125\,s & \textbf{67\% / 84\,s} \\
$>$50\% & 12 & 100\% / 125\,s & \textbf{98\% / 122\,s} \\ \midrule
All & 103 & 100\% / 125\,s & \textbf{61\% / 78\,s} \\
\bottomrule
\end{tabular}
\caption{\textbf{Compute follows the region.} Timings from a separate run. Share of tokens computed and seconds per clip on one H200, by the share of the frame covered by the region.}
\label{tab:efficiency}
\end{table}

%% file: figures/fig_patterns.tex
\begin{figure}[t]
\centering
\includegraphics[width=0.92\linewidth]{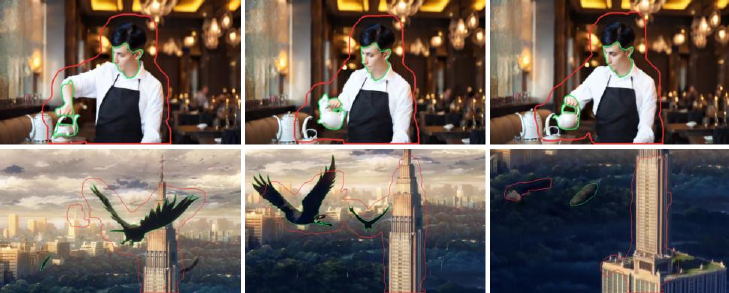}
\caption{\textbf{Our regions cover whole objects, while the annotations mark only broken pixels.} Dataset annotations (green) mark only the visibly broken pixels, such as the teapot spout or parts of the birds. Our regions (red) cover the objects involved and the places they move through, and leave out correctly rendered parts.}
\label{fig:patterns}
\end{figure}

%% file: tables/s3_study_mos.tex
\begin{table}[t]
\centering\footnotesize
\setlength{\tabcolsep}{4pt}
\begin{tabular}{lccc}
\toprule
Condition & Localize & Fixed & Preserved \\
\midrule
\method & 3.72 & 3.63 & 4.36 \\
\method w/o tuning & 3.67 & 3.32 & 4.24 \\
\method w/o tuning and ref. & 3.52 & 1.84 & 1.70 \\
EditCtrl & 3.62 & 1.36 & 1.71 \\
Gemini Omni 1.1 Flash & 1.02 & 2.95 & 3.40 \\
Wan2.2-A14B & 1.05 & 2.37 & 1.48 \\
VACE-14B & 1.05 & 2.06 & 1.52 \\
\midrule
Dataset label (+24\,px) & 3.91 & 3.19 & 4.28 \\
Qwen2.5-VL-72B + SAM3 & 3.33 & 2.80 & 3.73 \\
\method localizer (retest) & 3.67 & 3.14 & 4.28 \\
\bottomrule
\end{tabular}
\caption{\textbf{\method is rated highest on fixing and preservation, and our regions above those of the 72B pipeline.} All conditions of the perceptual study, eleven observers and ten clips. MOS on a 5-point scale, recovered with SUREAL. The upper rows repair our region and the lower rows vary the region with the inpainter held fixed.}
\label{tab:supp_mos}
\end{table}

%% file: tables/s4_auto_human.tex
\begin{table}[t]
\centering\small
\setlength{\tabcolsep}{3pt}
\resizebox{\linewidth}{!}{%
\begin{tabular}{lccccc}
\toprule
 & \multicolumn{3}{c}{Spearman $\rho$} & \multicolumn{2}{c}{Pairs} \\
Measure & Loc. & Fix. & Pres. & Fix. & Pres. \\
\midrule
VLM judge, change in severity & -0.12 & +0.43 & +0.31 & 68\% & 64\% \\
VLM judge, severity of the result & -0.19 & +0.31 & +0.31 & 68\% & 64\% \\
LPIPS & +0.39 & +0.28 & +0.58 & 72\% & 86\% \\
FID per clip & +0.20 & +0.64 & +0.77 & 77\% & 80\% \\
CLIP similarity & +0.20 & +0.65 & +0.83 & 76\% & 88\% \\
DINOv2 similarity & +0.16 & +0.66 & +0.81 & 75\% & 86\% \\
\bottomrule
\end{tabular}}
\caption{\textbf{The preservation measures track the observers, and the VLM judge tracks fixing only moderately.} Automatic measures against the observers on the 80 study trials without preference tuning. Spearman $\rho$ with the mean rating per trial (signed so that higher is better), and the share of same-clip method pairs ordered as the observers order them.}
\label{tab:supp_auto}
\end{table}

%% file: figures/grid_tight.tex
{\setlength{\tabcolsep}{0.6pt}\renewcommand{\arraystretch}{0.6}\setlength{\fboxsep}{0pt}\setlength{\fboxrule}{0.3pt}%
\begin{tabular}{@{}rccc@{}}
 & \scriptsize Region, frame 12 & \scriptsize Repair, frame 12 & \scriptsize Repair, frame 44 \\[1pt]
\scriptsize Annotation (+24\,px)\hspace{2pt} & \fcolorbox{gray!60}{white}{\includegraphics[width=\dimexpr0.27\linewidth-2\fboxrule\relax]{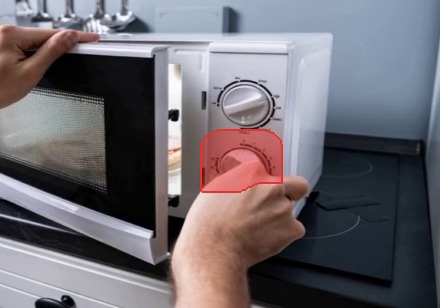}} & \fcolorbox{gray!60}{white}{\includegraphics[width=\dimexpr0.27\linewidth-2\fboxrule\relax]{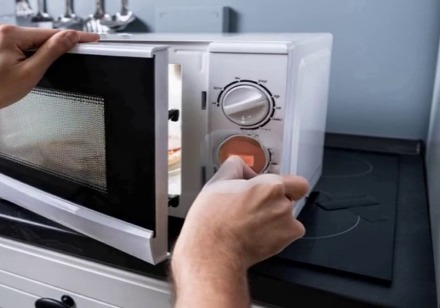}} & \fcolorbox{gray!60}{white}{\includegraphics[width=\dimexpr0.27\linewidth-2\fboxrule\relax]{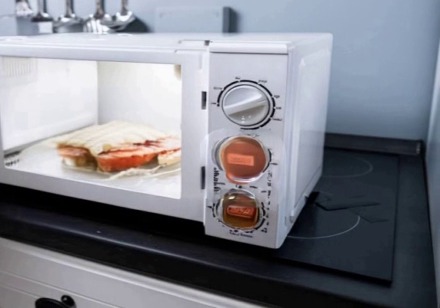}} \\
\scriptsize Our region\hspace{2pt} & \fcolorbox{gray!60}{white}{\includegraphics[width=\dimexpr0.27\linewidth-2\fboxrule\relax]{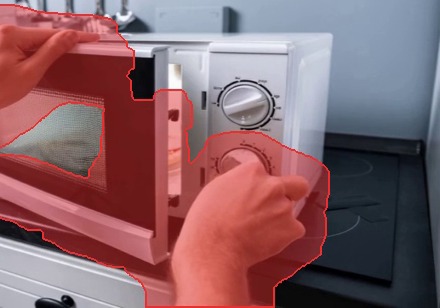}} & \fcolorbox{gray!60}{white}{\includegraphics[width=\dimexpr0.27\linewidth-2\fboxrule\relax]{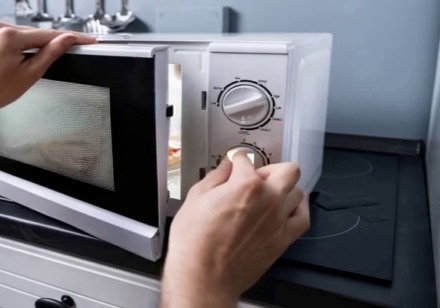}} & \fcolorbox{gray!60}{white}{\includegraphics[width=\dimexpr0.27\linewidth-2\fboxrule\relax]{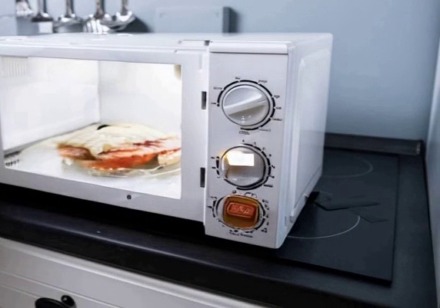}} \\
\end{tabular}}

%% file: sec/X_prompts.tex
\section{Prompts}
\label{sec:supp_prompts}
This section lists every prompt used by the pipeline, its training and the evaluation, verbatim. Text in braces is filled in per clip.

\subsection{Teacher for the region targets}
\paragraph{Physics script.} Qwen2.5-VL-72B sees a strip of low-resolution frames and a few full frames with the artifact annotation outlined in red, and answers with a description of the clip and the objects involved.
\begin{quote}\footnotesize\ttfamily\raggedright
You are shown ONE short AI-generated video. First \{n\_strip\} low-resolution frames in time order (to judge motion), then \{k\} full frames; a RED outline marks where a human annotator saw a generative artifact (the annotation may be incomplete).\\[0.6em]
Your job is to plan a PHYSICALLY PLAUSIBLE repair that keeps the video's meaning. Rules:\\
- Keep the scene's intent. If a person opens a door, the repaired video still shows the person opening that door -- the door must stay and move correctly (rotate on its hinges). NEVER fix an artifact by deleting the object it belongs to, and never add objects that are not already there.\\
- Name the objects that must be REDRAWN (their whole shape and motion, not just the red region) and everything that must stay untouched.\\[0.6em]
Respond with ONLY this JSON:\\
\{"scene": "\textless{}one sentence: what the video shows\textgreater{}",\\
 "intent": "\textless{}the action or event the video is about\textgreater{}",\\
 "involved\_objects": [\{"name": "\textless{}object\textgreater{}", "grounding\_phrase": "\textless{}1-4 words for a segmenter\textgreater{}", "role": "\textless{}what it does in the scene\textgreater{}"\}],\\
 "violations": [\{"what": "\textless{}what is physically wrong\textgreater{}", "when": "\textless{}early|middle|late|throughout\textgreater{}", "why\_impossible": "\textless{}which physical/anatomical rule it breaks\textgreater{}"\}],\\
 "correct\_behavior": "\textless{}what a physically correct version of this video shows, concretely: positions, motion, contacts, counts\textgreater{}",\\
 "must\_keep": ["\textless{}objects/people/background that must remain present and unchanged in identity\textgreater{}"],\\
 "must\_not": ["\textless{}forbidden 'fixes', e.g. removing the door, freezing the person\textgreater{}"],\\
 "edit\_scope": "\textless{}texture|object|motion\textgreater{}"\}
\end{quote}

\paragraph{Redraw or context.} A second call decides which of the named objects must be redrawn and which can stay.
\begin{quote}\footnotesize\ttfamily\raggedright
You are shown \{n\_strip\} frames, in time order, of ONE short AI-generated video, then \{k\} full frames where a RED outline marks a generative artifact. A repair is planned:\\
- what is physically wrong: \{violations\}\\
- what the corrected video must show: \{correct\}\\[0.6em]
For each object below decide whether the repair must REDRAW it (its pixels have to change: it is the broken object, or it must move / change shape so the corrected action is possible -- e.g. the door AND its hinge side, the hand that pushes it) or whether it is CONTEXT (it is touched or nearby but can stay exactly as it is -- e.g. the floor, the table, the wave, the background people).\\
Objects: \{objects\}\\[0.6em]
Respond with ONLY this JSON:\\
\{"redraw": ["\textless{}object names copied exactly from the list\textgreater{}"], "context": ["\textless{}object names copied exactly\textgreater{}"], "missing\_redraw": ["\textless{}1-4 word grounding phrase for any object that must be redrawn but is not in the list\textgreater{}"]\}
\end{quote}

\paragraph{Segmentation.} SAM3 is prompted with the \texttt{grounding\_phrase} of every object marked for redrawing (one to four words, e.g.\ ``kitchen appliance''), and tracks it through the clip.

\subsection{Inpainter}
\paragraph{Prompt.} Used for every clip. Its umT5 embedding is computed once and stored, together with the negative prompt below, so the text encoder is not loaded at test time.
\begin{quote}\footnotesize\ttfamily\raggedright
the video with all artifacts corrected, physically plausible
\end{quote}

\paragraph{Negative prompt.} The standard negative prompt of Wan2.1, used with classifier-free guidance.
\begin{quote}\footnotesize\ttfamily\raggedright
Bright tones, overexposed, static, blurred details, subtitles, style, works, paintings, images, static, overall gray, worst quality, low quality, JPEG compression residue, ugly, incomplete, extra fingers, poorly drawn hands, poorly drawn faces, deformed, disfigured, misshapen limbs, fused fingers, still picture, messy background, three legs, many people in the background, walking backwards
\end{quote}

\subsection{Baselines}
\paragraph{Regeneration.} VACE-1.3B, VACE-14B and Wan2.2-A14B use the same prompt and negative prompt as our inpainter.

\paragraph{Gemini Omni.} The instruction sent with each video.
\begin{quote}\footnotesize\ttfamily\raggedright
can you fix the AI generated artifacts in this video to make it look more natural? fix this most salient ones while preserving the rest of the scene content to make it seem like the user is viewing the same video, just a fixed version.
\end{quote}

\subsection{Preference ranking with Video Jev}
\paragraph{Questions.} Each is a yes-or-no question answered with a probability, and the score of a sample is P(no) of the first times P(no) of the second.
\begin{quote}\footnotesize\ttfamily\raggedright
Is this video AI-generated?  (yes, it is AI-generated / no, it is real footage)\\
Does this video contain visual artifacts, such as objects or body parts that morph, melt, or appear from nowhere?  (yes, there are visual artifacts / no visual artifacts)
\end{quote}

\subsection{VLM judge}
\paragraph{Severity.} Qwen2.5-VL-72B sees the full middle frame and three crops around the region (at 25, 50 and 75\% of the clip). The severity is the expected digit under its next-token probabilities, and the judge $\Delta$ is the severity of the result minus that of the original.
\begin{quote}\footnotesize\ttfamily\raggedright
You are an expert inspector of AI-GENERATED video. A generative artifact is content that is semantically WRONG for the real world, not merely low resolution. Inspect the region for each of these categories:\\
  1. Deformation / shape distortion: limbs bending like rubber, malformed hands or faces (wrong finger count, extra eyes), bodies melting or warping during movement.\\
  2. Merging / splitting: two objects fusing into one, one object splitting in two, objects overlapping impossibly.\\
  3. Unnatural motion: physically impossible movement paths, people walking into walls, still things moving (a trail moves), unsmooth or repeating motion, wrong liquid physics.\\
  4. Temporal inconsistency: things appearing or disappearing without cause, content changing identity/colour/texture between frames, a hole opening and closing.\\
  5. Texture collapse / blurring: sudden abnormal blur on one object while the rest is sharp, smeared or dissolving texture, background distortion around a moving object.\\
  6. Lighting / shadow / colour errors: shadows that do not match, a colour cast or hue shift on part of the scene.\\
  7. Compositing seams: a visible border where a region does not match its surroundings in sharpness, colour or motion.\\
Artifacts often have weak or no physical boundary -- judge by semantic wrongness, not object edges. Some artifacts are TEMPORAL and only visible across the frames shown. Ignore low resolution and compression.\\[0.6em]
Rate how BROKEN the highlighted region looks, 0-9, counting artifacts from the categories above. 0 = indistinguishable from real camera footage. 2 = a trained eye notices something slightly off. 5 = clearly wrong content a casual viewer would notice. 7 = grossly broken or dissolving. 9 = unrecognizable. Answer with EXACTLY ONE digit 0-9 and nothing else.
\end{quote}